\documentclass{article}

\usepackage[utf8]{inputenc}
\usepackage[T1]{fontenc}
\usepackage{microtype}
\usepackage{graphicx}
\usepackage{subcaption}
\usepackage{booktabs}

\usepackage{hyperref}

\usepackage[preprint]{neurips_2026}

\usepackage{amsmath}
\usepackage{amssymb}
\usepackage{mathtools}
\usepackage{amsthm}

\usepackage{amsmath,amsfonts,bm}

\def\eqref#1{equation~\ref{#1}}

\def\1{\bm{1}}

\DeclareMathAlphabet{\mathsfit}{\encodingdefault}{\sfdefault}{m}{sl}
\SetMathAlphabet{\mathsfit}{bold}{\encodingdefault}{\sfdefault}{bx}{n}

\usepackage{url}
\usepackage{placeins}
\usepackage{multirow}
\usepackage{tikz}
\usetikzlibrary{arrows.meta, positioning, calc, shapes.geometric, fit, backgrounds, decorations.pathreplacing}
\usepackage{shapesymbols}
\usepackage{stfloats}
\usepackage{algorithm}
\usepackage{algorithmic}

\usepackage[capitalize,noabbrev]{cleveref}

\theoremstyle{plain}

\theoremstyle{definition}

\theoremstyle{remark}

\usepackage[disable,textsize=tiny]{todonotes}

\title{Understanding Goal Generalisation in Sequential Reinforcement Learning}

\author{%
  Jason Ross Brown \\
  University of Cambridge \\
  Cambridge, United Kingdom \\
  \texttt{jrb239@cam.ac.uk} \\
  \And
  Edward James Young \\
  University of Cambridge \& Geodesic Research \\
  Cambridge, United Kingdom \\
}

\begin{document}

\maketitle

\begin{abstract}
Reinforcement learning agents often exhibit unintended goal-directed behaviour outside their training distribution, but we currently lack a principled understanding of how such agents will generalise to novel environments based on their training history.
We address this gap for agents trained sequentially on one or more tasks.
We study over 100 sequential training pipelines, evaluating behaviour across over 250 out-of-distribution environments. We find that salient features drive generalisation, and that goals learnt early in training can persist and influence those acquired later.
To explain these phenomena, we introduce \emph{latent policy gradients}, a method that predicts what out-of-distribution behaviour a training pipeline will likely induce.
Our method simulates the evolution of low-dimensional latent variables during training according to what would achieve high reward on the training objective with respect to a simple model of how the latent variables map to behaviour.
It achieves strong predictive accuracy, generalises to unseen types of training pipeline, and is interpretable.
Our findings demonstrate that while out-of-distribution RL agent behaviour is dependent on the whole training pipeline, this dependence has an underlying structure we can capture, laying groundwork for understanding goal generalisation from a developmental perspective.
\end{abstract}

\section{Introduction}\label{sec:introduction}

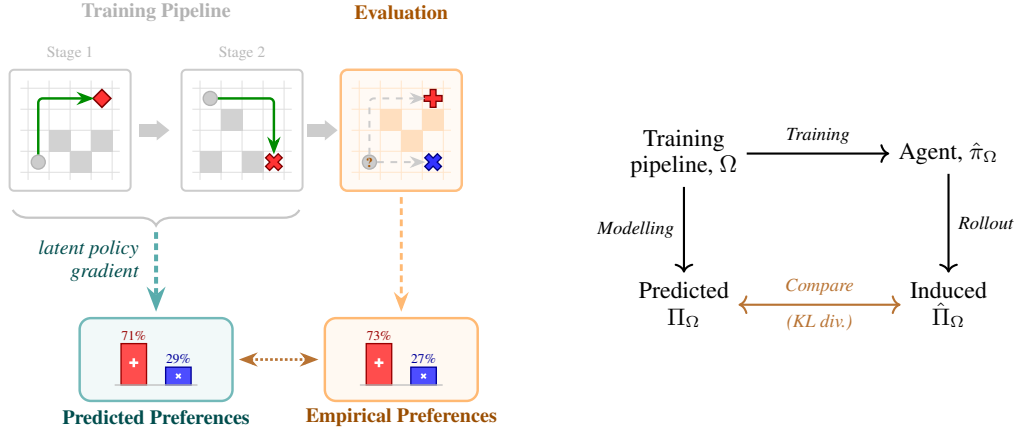
\begin{figure}[t]
    \centering
    \begin{subfigure}[c]{0.48\linewidth}
        \centering
        \resizebox{\linewidth}{!}{
    \begin{tikzpicture}[
        >=Stealth,
        scale=1.0,
        transform shape,
        mazebox/.style={rounded corners=2pt, draw=gray!40, fill=white,
                       minimum height=1.9cm, minimum width=1.9cm, line width=0.9pt},
        evalmazebox/.style={rounded corners=3pt, draw=orange!50, fill=orange!4,
                           minimum height=1.9cm, minimum width=1.9cm, line width=1.1pt},
        prefbox/.style={rounded corners=5pt, draw=teal!55, fill=teal!5, line width=1.1pt,
                       minimum height=1.3cm, minimum width=2.4cm},
    ]

    \newcommand{\agent}[3]{
        \fill[gray!40] (#1, #2) circle (#3);
        \draw[gray!60, line width=0.6pt] (#1, #2) circle (#3);
    }

    \newcommand{\plusshape}[4]{
        \fill[#1]
            (-#4, -#3) -- (-#4, -#4) -- (-#3, -#4) -- (-#3, #4) -- (-#4, #4) --
            (-#4, #3) -- (#4, #3) -- (#4, #4) -- (#3, #4) -- (#3, -#4) --
            (#4, -#4) -- (#4, -#3) -- cycle;
        \draw[#2, line width=0.6pt]
            (-#4, -#3) -- (-#4, -#4) -- (-#3, -#4) -- (-#3, #4) -- (-#4, #4) --
            (-#4, #3) -- (#4, #3) -- (#4, #4) -- (#3, #4) -- (#3, -#4) --
            (#4, -#4) -- (#4, -#3) -- cycle;
    }

    \newcommand{\blockarrow}[3]{
        \fill[#3] (#1, #2+0.1) rectangle (#1+0.28, #2-0.1);
        \fill[#3] (#1+0.28, #2+0.16) -- (#1+0.48, #2) -- (#1+0.28, #2-0.16) -- cycle;
    }

    \node[font=\small\bfseries, gray!60] at (1.35, 1.85) {Training Pipeline};

    \node[font=\scriptsize, gray!55] at (0, 1.2) {Stage 1};
    \node[mazebox] (m1) at (0, 0) {};

    \foreach \x in {-0.66, -0.33, 0, 0.33, 0.66} {
        \draw[gray!25, line width=0.4pt] (\x, -0.78) -- (\x, 0.78);
        \draw[gray!25, line width=0.4pt] (-0.78, \x) -- (0.78, \x);
    }
    \fill[gray!35] (0, -0.66) rectangle (0.33, -0.33);       %
    \fill[gray!35] (0.33, -0.33) rectangle (0.66, 0);        %
    \fill[gray!35] (-0.33, -0.33) rectangle (0, 0);          %

    \agent{-0.50}{-0.50}{0.11}

    \node[fill=red!80, draw=red!50!black, regular polygon, regular polygon sides=4,
          minimum size=8pt, rotate=45, line width=0.7pt, inner sep=0pt] at (0.50, 0.50) {};

    \draw[->, green!55!black, line width=1.1pt, rounded corners=2pt]
        (-0.50, -0.38) -- (-0.50, 0.50) -- (0.38, 0.50);

    \blockarrow{1.08}{0}{gray!40}

    \node[font=\scriptsize, gray!55] at (2.7, 1.2) {Stage 2};
    \node[mazebox] (m2) at (2.7, 0) {};

    \foreach \x in {-0.66, -0.33, 0, 0.33, 0.66} {
        \draw[gray!25, line width=0.4pt] (2.7+\x, -0.78) -- (2.7+\x, 0.78);
        \draw[gray!25, line width=0.4pt] (2.7-0.78, \x) -- (2.7+0.78, \x);
    }
    \fill[gray!35] (2.7-0.66, -0.66) rectangle (2.7-0.33, -0.33);  %
    \fill[gray!35] (2.7-0.33, 0) rectangle (2.7, 0.33);            %
    \fill[gray!35] (2.7, -0.66) rectangle (2.7+0.33, -0.33);       %

    \agent{2.7-0.50}{0.50}{0.11}

    \begin{scope}[shift={(2.7+0.50, -0.50)}, rotate=45]
        \plusshape{red!80}{red!50!black}{0.14}{0.045}
    \end{scope}

    \draw[->, green!55!black, line width=1.1pt, rounded corners=2pt]
        (2.7-0.38, 0.50) -- (2.7+0.50, 0.50) -- (2.7+0.50, -0.38);

    \blockarrow{3.72}{0}{gray!40}

    \node[font=\small\bfseries, orange!65!black] at (5.2, 1.85) {Evaluation};
    \node[evalmazebox] (meval) at (5.2, 0) {};

    \foreach \x in {-0.66, -0.33, 0, 0.33, 0.66} {
        \draw[orange!20, line width=0.35pt] (5.2+\x, -0.78) -- (5.2+\x, 0.78);
        \draw[orange!20, line width=0.35pt] (5.2-0.78, \x) -- (5.2+0.78, \x);
    }
    \fill[orange!25] (5.2-0.33, 0) rectangle (5.2, 0.33);
    \fill[orange!25] (5.2, -0.33) rectangle (5.2+0.33, 0);
    \fill[orange!25] (5.2+0.33, 0) rectangle (5.2+0.66, 0.33);

    \agent{5.2-0.50}{-0.50}{0.11}
    \node[font=\tiny\bfseries, orange!65!black] at (5.2-0.50, -0.50) {?};

    \begin{scope}[shift={(5.2+0.50, 0.50)}]
        \plusshape{red!80}{red!50!black}{0.14}{0.045}
    \end{scope}

    \begin{scope}[shift={(5.2+0.50, -0.50)}, rotate=45]
        \plusshape{blue!80}{blue!50!black}{0.14}{0.045}
    \end{scope}

    \draw[->, gray!40, line width=0.9pt, dashed, rounded corners=2pt]
        (5.2-0.50, -0.38) -- (5.2-0.50, 0.50) -- (5.2+0.38, 0.50);
    \draw[->, gray!40, line width=0.9pt, dashed, rounded corners=2pt]
        (5.2-0.38, -0.50) -- (5.2+0.38, -0.50);

    \draw[decorate, decoration={brace, amplitude=7pt, mirror}, gray!45, line width=1pt]
        (-0.9, -1.25) -- (3.6, -1.25);

    \node[prefbox] (predbox) at (1.35, -3.6) {};
    \node[font=\small\bfseries, teal!60!black] at (1.35, -4.5) {Predicted Preferences};

    \draw[->, teal!65, line width=1.8pt, densely dashed]
        (1.35, -1.6) -- (1.35, -2.85);

    \node[font=\small\itshape, teal!60!black, align=right, anchor=east] at (1.2, -2.05) {latent policy\\gradient};

    \begin{scope}[shift={(1.35, -3.7)}]
        \fill[red!70] (-0.55, -0.30) rectangle (-0.15, 0.35);
        \draw[red!60!black, line width=0.6pt] (-0.55, -0.30) rectangle (-0.15, 0.35);
        \begin{scope}[shift={(-0.35, 0.05)}, scale=0.25]
            \fill[white]
                (-0.08, -0.28) -- (-0.08, -0.08) -- (-0.28, -0.08) -- (-0.28, 0.08) --
                (-0.08, 0.08) -- (-0.08, 0.28) -- (0.08, 0.28) -- (0.08, 0.08) --
                (0.28, 0.08) -- (0.28, -0.08) -- (0.08, -0.08) -- (0.08, -0.28) -- cycle;
        \end{scope}

        \fill[blue!70] (0.15, -0.30) rectangle (0.55, -0.02);
        \draw[blue!60!black, line width=0.6pt] (0.15, -0.30) rectangle (0.55, -0.02);
        \begin{scope}[shift={(0.35, -0.16)}, rotate=45, scale=0.18]
            \fill[white]
                (-0.08, -0.28) -- (-0.08, -0.08) -- (-0.28, -0.08) -- (-0.28, 0.08) --
                (-0.08, 0.08) -- (-0.08, 0.28) -- (0.08, 0.28) -- (0.08, 0.08) --
                (0.28, 0.08) -- (0.28, -0.08) -- (0.08, -0.08) -- (0.08, -0.28) -- cycle;
        \end{scope}

        \draw[gray!50, line width=0.6pt] (-0.65, -0.30) -- (0.65, -0.30);
        \node[font=\tiny, red!60!black] at (-0.35, 0.48) {71\%};
        \node[font=\tiny, blue!60!black] at (0.35, 0.12) {29\%};
    \end{scope}

    \node[prefbox, draw=orange!55, fill=orange!5] (empbox) at (5.2, -3.6) {};
    \node[font=\small\bfseries, orange!60!black] at (5.2, -4.5) {Empirical Preferences};

    \draw[->, orange!55, line width=1.3pt, densely dashed]
        ($(meval.south) + (0, -0.08)$) -- ($(empbox.north) + (0, 0.08)$);

    \begin{scope}[shift={(5.2, -3.7)}]
        \fill[red!70] (-0.55, -0.30) rectangle (-0.15, 0.35);
        \draw[red!60!black, line width=0.6pt] (-0.55, -0.30) rectangle (-0.15, 0.35);
        \begin{scope}[shift={(-0.35, 0.05)}, scale=0.25]
            \fill[white]
                (-0.08, -0.28) -- (-0.08, -0.08) -- (-0.28, -0.08) -- (-0.28, 0.08) --
                (-0.08, 0.08) -- (-0.08, 0.28) -- (0.08, 0.28) -- (0.08, 0.08) --
                (0.28, 0.08) -- (0.28, -0.08) -- (0.08, -0.08) -- (0.08, -0.28) -- cycle;
        \end{scope}

        \fill[blue!70] (0.15, -0.30) rectangle (0.55, -0.02);
        \draw[blue!60!black, line width=0.6pt] (0.15, -0.30) rectangle (0.55, -0.02);
        \begin{scope}[shift={(0.35, -0.16)}, rotate=45, scale=0.18]
            \fill[white]
                (-0.08, -0.28) -- (-0.08, -0.08) -- (-0.28, -0.08) -- (-0.28, 0.08) --
                (-0.08, 0.08) -- (-0.08, 0.28) -- (0.08, 0.28) -- (0.08, 0.08) --
                (0.28, 0.08) -- (0.28, -0.08) -- (0.08, -0.08) -- (0.08, -0.28) -- cycle;
        \end{scope}

        \draw[gray!50, line width=0.6pt] (-0.65, -0.30) -- (0.65, -0.30);
        \node[font=\tiny, red!60!black] at (-0.35, 0.48) {73\%};
        \node[font=\tiny, blue!60!black] at (0.35, 0.12) {27\%};
    \end{scope}

    \draw[<->, color={rgb,255:red,184;green,115;blue,51}, line width=1.2pt, densely dotted]
        ($(predbox.east) + (0.08, 0)$) -- ($(empbox.west) + (-0.08, 0)$);

    \end{tikzpicture}
    }
    \end{subfigure}
    \hfill
    \begin{subfigure}[c]{0.48\linewidth}
        \centering
        \begin{tikzpicture}[node distance=0.8cm and 2cm, auto, font=\small]
            \node (tl) at (0,0) [align=center] {Training\\pipeline, $\Omega$};
            \node (tr) at (3.5,0) [align=center] {Agent, $\hat{\pi}_\Omega$};
            \node (bl) at (0,-2) [align=center] {Predicted\\$\Pi_\Omega$};
            \node (br) at (3.5,-2) [align=center] {Induced\\$\hat{\Pi}_\Omega$};
            \draw[->, thick] (tl) -- (tr) node[midway, above, font=\scriptsize] {\emph{Training}};
            \draw[->, thick] (tl) -- (bl) node[midway, left, font=\scriptsize] {\emph{Modelling}};
            \draw[->, thick] (tr) -- (br) node[midway, right, align=center, font=\scriptsize] {\emph{Rollout}};
            \draw[<->, thick, color={rgb,255:red,184;green,115;blue,51}] (bl) -- (br)
                node[midway, above, font=\scriptsize] {\emph{Compare}}
                node[midway, below, font=\scriptsize] {\emph{(KL div.)}};
        \end{tikzpicture}
    \end{subfigure}
    \caption{\textbf{Left: Illustration of our experimental design.} Our experimental design is covered in \cref{sec:experimental_setting}. RL agents are trained on pipelines involving either one or two stages (\emph{e.g.}, trained to pursue \sym{red}{diamond} in stage 1 and \sym{red}{cross} in stage 2). They are then evaluated in out-of-distribution environments containing two objects (\emph{e.g.}, \sym{red}{plus} and \sym{blue}{cross}) to generate an empirical preference distribution. We explore these distributions in \cref{sec:empirical_analysis}. Then, in \cref{sec:modelling_generalisation}, we introduce a method of predicting preference from training pipelines---the \emph{latent policy gradients} method---and compare the fit between the predicted and empirical preferences.
    \textbf{Right: The goal generalisation modelling problem.} Training pipelines are used to train agents, which are then rolled out in evaluation environments to produce an induced preference function. Modelling gives predicted preference functions. We find the fit of our model by comparing the average KL-divergence between the predicted and induced preference functions.}
    \label{fig:intro_overview_and_modelling_problem}
\end{figure}

Reinforcement learning (RL) agents trained in one environment may exhibit very different behaviour when placed in environments not seen in training \citep{kirk_survey_2023,langosco_goal_2023}.
While unpredictable out-of-distribution behaviour can be found across machine learning methods, RL can sometimes lead to agents which continue to show coherent, goal-directed behaviour out-of-distribution.
A particular instance of this phenomenon is \emph{goal misgeneralisation}, in which RL agents may learn to pursue a goal that obtains high reward within training environments, but not on out-of-distribution samples \citep{shah_goal_2022,langosco_goal_2023}.
As agents trained via RL are increasingly deployed in the real world, they will encounter situations outside those they are trained on; understanding and predicting behaviour in these situations in advance is therefore necessary for providing guarantees of safety and reliability \citep{hubinger_risks_2021,hendrycks_overview_2023,ngo_alignment_2025}.

This is particularly pressing in modern frontier AI systems, which are typically trained through a variety of stages including self-supervised pre-training, supervised fine-tuning, reinforcement learning from human feedback \citep{ouyang_training_2022,bai_training_2022,rafailov_direct_2023}, and reinforcement learning against verifiable rewards \citep{deepseek-ai_deepseek-r1_2025}.
Some of these stages are explicitly intended to shape the values the system will act on, as in Constitutional AI \citep{bai_constitutional_2022} and Character Training \citep{anthropic_claudes_2024,maiya_open_2025}.
Additionally, existing empirical work has surfaced unexpected and often concerning behaviours in frontier AI systems, particularly when models undergo further training on new objectives \citep{betley_weird_2025,taylor_school_2025,macdiarmid_natural_2025,betley_emergent_2026,dubinski_conditional_2026}.
These findings illustrate \emph{what} can go wrong with multi-stage training of AI systems deployed out-of-distribution, but we lack a predictive understanding of \emph{how} these behaviours arise.

This paper seeks to advance our understanding of \emph{goal generalisation} in reinforcement learning agents---given the training history of an agent and knowledge of its in-distribution behaviour, or that of similar agents, how can we predict the behaviour of that agent on out-of-distribution samples?
We address this question specifically in the case of \emph{transfer learning}, in which RL agents are trained on multiple tasks in sequence \citep{khetarpal_towards_2022,iman_review_2023}.
Our hope is to understand this question at a fundamental level, building a principled theory of how training pipelines shape what agents learn to value.
Building such an understanding requires solid theoretical foundations, which in turn call for controlled settings where training configurations can be systematically varied. We therefore explore the goal generalisation problem in a deliberately controlled toy setting.

We train RL agents within maze environments to pursue objects that have a particular colour and shape.
Agents are trained either to pursue a single goal object, or sequentially to pursue two different goal objects across two training stages.
We then evaluate these agents on out-of-distribution environments containing two objects, and measure the frequency at which they select one object over another.
This setup is detailed in \cref{sec:experimental_setting}.
In \cref{sec:empirical_analysis} we explore these preferences empirically, highlighting several important phenomena.

We subsequently attempt to understand and explain our observations in \cref{sec:modelling_generalisation}.
Inspired by models of associative learning from behavioural psychology \citep{rescorla_theory_1972}, we posit that agent goals can be understood in terms of a few \emph{latent variables} which evolve during training in a predictable and understandable manner.
We term this evolution \emph{latent policy gradients}.
We first show that our method provides a good fit to our empirical preference data, despite having only a few parameters.
Then, we show that the hyperparameters of our method admit a natural interpretation in terms of feature similarity, and that the evolution of our latent variables can be understood as iterated projection onto hyperplanes.

Illustrated in \cref{fig:intro_overview_and_modelling_problem}, the main contributions of our paper are as follows:
\begin{enumerate}
    \item We provide an empirical analysis of goal generalisation behaviour of RL agents trained with over 100 different training pipelines, each evaluated on over 250 out-of-distribution environments. We demonstrate a number of empirical phenomena through both single case studies and population-level effects.
    \item We provide a novel formulation of the goal generalisation problem as predicting the out-of-distribution behaviour of RL agents based on their training pipeline.
    \item We provide a method for solving this problem---\emph{latent policy gradients}---and show that it achieves good fit with our empirical data while having interpretable dynamics and hyperparameters.\footnote{Code and data for reproducing our experiments are available at \url{https://github.com/jr-brown/latent-policy-gradients}.}
\end{enumerate}

\section{Background}\label{sec:background}

An RL agent exhibits \emph{goal misgeneralisation} when it performs competently but pursues unintended goals on out-of-distribution (OOD) inputs \citep{shah_goal_2022,langosco_goal_2023}---the agent behaves competently, but toward the wrong objective.
Goal misgeneralisation poses a significant gap in our ability to safely deploy RL agents, as it can lead to agents that perform well during training and evaluation but behave unexpectedly once deployed.
In this paper, we investigate the broader problem of \emph{goal generalisation}: given knowledge of an agent's training pipeline, can we predict how the agent's learnt behaviour will generalise to unseen environments?

As discussed in \cref{sec:introduction}, modern frontier AI systems are trained through multi-stage pipelines, some stages of which are explicitly designed to shape the system's general behaviour across many environments.
While transfer and continual learning have been studied extensively in general machine learning \citep{iman_review_2023} and in reinforcement learning specifically \citep{khetarpal_towards_2022}, comparatively little is known about how sequences of goal-directed RL training stages shape the final OOD behaviour of an agent.
This is the regime our work targets.
A broader discussion of related work is provided in \cref{sec:related_work}.

\section{Experimental setting}\label{sec:experimental_setting}

\subsection{The maze task}

We train CNN-based RL agents on procedurally generated maze navigation tasks (see \cref{app:exp_details_agent} for details on the training algorithm and network architecture).
Each episode, the agent (represented by a grey circle, \sym{grey}{circle}) is placed in a procedurally generated $8 \times 8$ maze containing a goal object with a fixed colour and shape, \emph{e.g.}, a red cross, \sym{red}{cross}. Each episode of RL training occurs in a new maze. The agent's observations are $128 \times 128$ RGB pixel images. See \cref{app:exp_details_maze}, \cref{fig:maze_example} for maze generation details and an example maze.

Each turn, the agent can move up, down, left, or right, with movement into a wall cell resulting in no change. A reward of +1 is generated when the agent reaches the goal object, whereby the episode terminates; every other transition results in a reward of -0.1. The environment has a fixed horizon of 200 steps, after which the episode is terminated with no further reward.
For some environments, we also include a distractor object of fixed shape and colour distinct from the goal. Distractor objects are visual and do not affect rewards. The colours of goal and distractor objects can vary between \black{}, \red{}, and \blue{}, and shapes vary between crosses (\Cross{}), diamonds (\Diam{}), plusses (\Plus{}), and rings (\Ring{}), giving a total of 12 different possible objects.
Visualisations of the shapes and colours are given in \cref{app:exp_details_maze}, \cref{fig:features}.

\subsection{Training pipelines and evaluation}\label{sec:data_gathering}

We denote by $\Omega$ a specific sequence of goal-directed RL training stages, which we term a \emph{training pipeline}, and by $\hat{\pi}_\Omega$ the agent that results from training through that pipeline.
To investigate how training pipelines affect goal generalisation, we collect data on the preferences of agents trained through a range of pipelines.
We consider training pipelines consisting of either one or two stages.
In each stage, the agent is trained for 3 million steps with PPO \citep{schulman_proximal_2017} on mazes with a fixed goal object, and either one or no distractor objects. For example, a two-stage pipeline may consist of first training the agent on mazes with \sym{red}{diamond} goal and no distractor, and then subsequently training the agent on mazes with \sym{blue}{cross} goal and \sym{red}{plus} distractor. See \cref{app:supplementary_figures_agent_training}, \cref{fig:training_curves}, for example learning curves from our agents.

For each of the 12 shape and colour combinations we train an agent in the single stage training pipeline. We then fine-tune independent copies of each of these agents with all possible \red{} and black goals to obtain two-stage pipeline results.\footnote{
Since our agents receive single-channel RGB observations, training with \blue{} second-stage goals would be symmetric with \red{} goals (each occupying one colour channel). We therefore omit \blue{} second-stage goals to avoid unnecessary repeats.}
This gives a total of 108 agents\footnote{
There are 3 colours and 4 shapes, resulting in 12 possible first-stage training goal objects. For second stage-training, omitting \blue{} goal objects leaves 2 colours and 4 shapes, and so 8 possible second-stage training goal objects. This gives a total of $12$ single-stage agents and $8 \times 12 = 96$ two-stage agents.
} for our analysis in \cref{sec:empirical_analysis}.

For \cref{sec:modelling_generalisation}, we extend our dataset by training additional agents with distractor objects present during training.
Distractors can have any colour and shape available to training goals, with the addition of \green{}.
This yields 190 unique training pipelines (both single and two stages, with and without distractors), giving a total of 298 agents.
Full details of the distractor pipeline configurations are given in \cref{app:exp_details_agent}.

For each agent, we define its preferences over objects operationally as follows. For each possible pair of objects, we rollout 100 episodes in mazes containing that pair, and record the fraction of times the agent reaches each of the two objects first, and the fraction of times the episode terminates at 200 steps without either object being reached. 
This includes objects with a novel colour (\green{}) and two novel shapes (\Circle{}, \HollowDiam{}) not seen in any training pipeline.
This gives a total of 276 evaluation environments.\footnote{
There are 24 possible objects (4 colours and 6 shapes). There are then $\binom{24}{2} = 276$ evaluation environments.
}
We denote the resulting empirical distribution over outcomes for agent $\hat{\pi}_\Omega$ as its \emph{empirical preference function}, $\hat{\Pi}_\Omega$.
Our modelling problem in \cref{sec:modelling_generalisation} is to predict $\hat{\Pi}_\Omega$ from $\Omega$ alone.

\section{Empirical analysis of agent values}\label{sec:empirical_analysis}

\subsection{Method}\label{sec:method_empirical}

Naively, one might expect agents to act erratically out of distribution, or if they were to act coherently, to simply pursue the last goal they were trained on.
We investigate both of these assumptions.
First, we test whether our trained agents display coherent OOD preferences at all.
\emph{A priori}, there is no reason to expect this: agents are trained to navigate toward a single goal object per environment, and their behaviour on environments containing novel object pairs is unconstrained by the training objective.
Having investigated the coherence of these preferences, we then explore their structure, and how they relate to the agents' training histories.

The notion of coherence we employ is \emph{Boltzmann-rationality}: for each agent, each object is assigned a scalar score, and the probability of preferring object $A$ to $B$ is given by $\sigma(\text{score}(A) - \text{score}(B))$, where $\sigma$ is the logistic function.
This compresses the full set of $\binom{n}{2}$ pairwise preference probabilities into just $n$ scores.
Boltzmann-rationality is a stochastic analogue of the transitivity requirement of classical preference theory \citep{von_neumann_theory_1947}, where if $A$ is preferred to $B$, and $B$ preferred to $C$, then $A$ must be preferred to $C$.
Analogously, Boltzmann-rationality implies that if $A$ is more likely to be chosen than $B$, and $B$ more likely than $C$, then $A$ must be more likely to be chosen than $C$.\footnote{More precisely, Boltzmann-rationality requires $\mathrm{logit}(\mathbb{P}(A \succ C)) = \mathrm{logit}(\mathbb{P}(A \succ B)) + \mathrm{logit}(\mathbb{P}(B \succ C))$.}
If an agent's preferences frequently violate this condition, they cannot be well-described by a set of scores, and the model will be unable to predict held-out pairwise choices.
We test the Boltzmann-rationality assumption by first dividing the data into $K=4$ folds, and then for each fold fitting scores on the other folds using the Elo algorithm \citep{elo_rating_2008}, and evaluating the fitted scores' ability to predict pairwise preferences from the initial held-out fold.
Specifically we compute the directional accuracy of the preferences predicted by the Boltzmann-rationality model, \emph{i.e.}, the proportion of held-out pairs for which the model correctly predicts which object is selected more frequently.
For more details on the fitting procedure, see \cref{app:elo_fitting}.

\subsection{Results}\label{sec:empirical_results}

\paragraph{Out-of-distribution agent preferences are coherent.}
We find our Elo scores have 89.73\% directional accuracy on held-out pairs in our 4-fold cross-validation ($\pm$ 0.18\% SE; see \cref{app:elo_holdout}).
The high validation accuracy confirms that agent preferences over OOD goals are approximately coherent in the sense described above---a non-trivial finding given that nothing in the training objective constrains OOD preferences in this manner.

Having validated the Elo scores as a good summary of agent preferences, we can now use them to investigate the structure of these preferences, and how they relate to training history.
To summarise how much each agent values individual features, we average Elo scores over all objects containing a given feature to produce \emph{marginalised Elo scores}.
We observe three clear patterns in the structure of these preferences, which we summarise here.
Full details, case studies, population-level analyses, and all supporting figures are given in \cref{app:empirical_details}.
In this section we only analyse training pipelines without distractors; \cref{app:supplementary_figures_agent_prefs} contains breakdowns of all marginalised Elo scores across all agents.

\paragraph{Some features are more salient in learning than others.}
Across our single-stage pipelines, we find that when agents are evaluated OOD, they tend to pursue objects sharing specific features of their training goal more than others.
In agreement with prior work on the inductive biases of CNNs \citep{ritter_cognitive_2017,feinman_learning_2018}, shape tends to drive generalisation more strongly than colour, with \Cross{}-shape being the most salient feature and black the least (\cref{fig:feature_saliency_combined}).
We also observe \emph{feature confusion}, where training on one feature can lead to valuing or de-valuing another (\cref{fig:cross_saliencies}).
We will return to this phenomenon in \cref{sec:modelling_generalisation}.

\paragraph{Values persist across training stages.}
In two-stage training pipelines, values learnt for features in the first stage's goal persist after the second stage, even when those features are not part of the second training objective (\cref{fig:value_persistence_combined}).
This holds both whether the two goals share some other feature or not.
Additionally, agents trained on more unique features generalise to pursue more objects (\cref{fig:feature_diversity}).

\paragraph{Repeated goal features' values are strengthened, and inhibit new values forming.}
When a feature is shared between two training stages, that feature's value is strengthened relative to if it only appeared in the second stage.
Conversely, the non-shared feature in the second-stage goal is valued much less than it would be if the shared feature was not present, indicating that existing values for features of the training goal inhibit the formation of new values (\cref{fig:value_inhibition_combined}).
These findings are consistent with known plasticity effects in continual learning \citep{nikishin_primacy_2022,nikishin_deep_2023,dohare_loss_2024} and gradient starvation \citep{pezeshki_gradient_2021}, whereby dominant features suppress learning of others.
As a corollary, we observe strong ordering effects when goals share a feature: flipping the order of training stages can significantly change the agent's final values (\cref{fig:ordering_effects}).

\section{Understanding generalisation through latent policy gradients}\label{sec:modelling_generalisation}

In this section, we attempt to understand the generalisation effects observed in \cref{sec:empirical_results} by modelling agent preferences as arising from low-dimensional latent variables that evolve during training via policy gradient ascent.\footnote{
Prior work in Inverse RL and Game Theory indicate preferences capture the underlying motivations of an agent \citep{von_neumann_theory_1947,skalse_invariance_2023}, we are also partially inspired by models of associative learning from behavioural psychology \citep{rescorla_theory_1972}.
}

\subsection{Problem Formalism}\label{sec:modelling_formalism}

We now provide additional formal detail on the preference function $\hat{\Pi}_\Omega$ introduced in \cref{sec:data_gathering}.
In the set of objects we include the null-object, $\bm{0}$, which corresponds to an episode terminating without an object being reached.
We write $\hat{\Pi}_\Omega\left( \bm{\phi}^{(a)} \middle| \bm{\phi}^{(a)}, \bm{\phi}^{(b)}, \bm{0} \right)$ to denote the empirical probability that when the agent $\hat{\pi}_\Omega$ is placed in an environment containing objects with features $\bm{\phi}^{(a)}, \bm{\phi}^{(b)}$, it terminates the episode by reaching the object with features $\bm{\phi}^{(a)}$.
Further, $\hat{\Pi}_\Omega\left( \bm{0} \middle| \bm{\phi}^{(a)}, \bm{\phi}^{(b)}, \bm{0} \right)$ is used to denote the empirical probability that the episode will terminate at some fixed horizon length without any object being reached (in our case, 200 steps).

The modelling problem can then be formulated as follows: find a mapping from training pipelines to preference functions, $\Omega \mapsto \Pi_\Omega$, which minimises the KL-divergence from the empirically observed preference function.
This is illustrated in \cref{fig:intro_overview_and_modelling_problem}.
In other words, we wish to minimise the following \emph{modelling loss}: 
\begin{equation}\label{eq:modelling_loss}
     D\left( \hat{\Pi}_\Omega\left(\bullet\middle|\bm{\phi}^{(a)}, \bm{\phi}^{(b)}, \bm{0}\right) \middle|\middle| \Pi_\Omega\left(\bullet\middle|\bm{\phi}^{(a)}, \bm{\phi}^{(b)}, \bm{0} \right) \right),
\end{equation}
which we average over $\Omega \sim \mathcal{D}_\text{train},  \left(\bm{\phi}^{(a)}, \bm{\phi}^{(b)}\right) \sim \mathcal{D}_\text{eval}.$

\subsection{The latent policy gradient method}\label{sec:latent_policy_gradient}

There are many ways to specify a mapping from training pipelines to predicted preference functions, $\Omega \to \Pi_\Omega$. In this section, we introduce the core of our modelling approach---the \emph{latent policy gradients} method.

We hypothesise that the preferences of policies can encapsulated by a small collection of \emph{latent variables} which determine agent behaviour, $\bm{w}$. Trivially, a large enough collection of latents can capture the agent's behaviour---indeed, we could take $\bm{w}$ to just be the entire parameter set of our agent's network. Our goal therefore is to find a set of \emph{low-dimensional} latents, which is much smaller than the number of parameters in the network, and which have readily interpretable meaning.

The first step of the latent policy gradients method is to specify a parametrisation of preference functions according to latent variables, $\Pi(\bullet|\bullet; \bm{w})$. This parametrisation may include hyperparameters. Given this parametrisation, we must then specify how the latent parameters, $\bm{w}$, depend on the training pipeline, $\Omega$. In other words, we must specify a mapping $\Omega \mapsto \bm{w}_\Omega$. Ideally, this map itself is again interpretable and meaningful. With some slight abuse of notation, we will then use $\Pi(\bullet|\bullet;\bm{w}_\Omega)$ as our predicted preference function, $\Pi_\Omega(\bullet|\bullet)$.

For each training pipeline, $\Omega$, we specify $\bm{w}_\Omega$ by performing gradient ascent for a fixed number of steps on modified policy objectives \emph{over preference functions}. We also include entropy regularisation in our objective \citep{haarnoja_reinforcement_2017,haarnoja_soft_2019}.
Entropy regularisation allows us to model how effectively we expect the agents to pursue their training objectives via a hyperparameter, $\tau$.
Concretely, when training on an $\Omega$ with a single stage with a single goal, $\bm{\phi}^{(g)}$, we use the following policy objective over preference functions:
\begin{align}\label{eq:modified_policy_objective}
    J(\Pi) = \Pi\left( \bm{\phi}^{(g)} \middle | \bm{\phi}^{(g)}, \bm{0} \right) + \tau h\left( \Pi\left( \bm{\phi}^{(g)} \middle | \bm{\phi}^{(g)}, \bm{0} \right) \right) 
\end{align}
where $h(\bullet)$ is the binary entropy function\footnote{$h(p) = -p \log(p) - (1-p)\log(1 - p)$ is the entropy of a Bernoulli random variable with probability $p$}. When the training environment of the stage includes a distractor object with features $\bm{\phi}^{(d)}$, we include that distractors features in the conditioning set, \emph{i.e.}, replace $\Pi\left(\bm{\phi}^{(g)} \middle| \bm{\phi}^{(g)}, \bm{0} \right)$ with $\Pi\left(\bm{\phi}^{(g)} \middle| \bm{\phi}^{(g)},\bm{\phi}^{(d)}, \bm{0} \right)$ in \cref{eq:modified_policy_objective}.
When the training pipeline contains multiple stages, we fit on corresponding $J$s in sequence, with the final $\bm{w}$ of one gradient ascent initialising the $\bm{w}$ of the next.

\subsection{Preference function parametrisation}\label{sec:parametrisation}

To apply latent policy gradients to our problem, we need to specify a parametrisation of our preference functions, $\Pi$, in terms of latent variables, $\bm{w} \mapsto \Pi(\bullet|\bullet;\bm{w})$. In \cref{sec:empirical_results} we showed that agent preferences are well-described by Elo scores, which assume Boltzmann-rational preferences.
We therefore adopt the same assumption here \citep{bradley_rank_1952,wulfmeier_maximum_2016,christiano_deep_2017,skalse_misspecification_2023}, and additionally parametrise values as linear in both goal features and the latent variables. Specifically, for a goal with features $\bm{\phi}^{(g)}$, the preference function is:
\begin{align}\label{eq:linear_boltzmann_rational}
    \Pi\left( \bm{\phi}^{(g)}\middle|\bm{\phi}^{(g)}, \bm{0}; \bm{w} \right) &= \frac{\exp\left( \bm{\phi}^{(g)} \cdot S\bm{w} \right) }{1 + \exp\left( \bm{\phi}^{(g)} \cdot S\bm{w} \right)}.
\end{align}
We choose the dimension of $\bm{w}$ to be 10, equal to the total number of object features.\footnote{Note that this parametrisation is equivalent to modelling Elo scores as linear in goal features, with $V(\bm{\phi}^{(g)}) = \bm{\phi}^{(g)} \cdot S\bm{w}$.} In \cref{app:latent_dimension} we show that performance saturates at the number of feature dimensions, as expected, and can be reduced to 8 without degradation.
Note that this is much smaller than the 4.8 million parameters of our trained CNN-based agents.
The saliency matrix, $S$, is a hyperparameter that controls which features are learnt to be valued faster, as well as modelling feature confusion.
As discussed in \cref{sec:empirical_results}, we observe empirically that training on one set of features can cause agents to pursue goals with features they were not trained on; the off-diagonal entries of $S$ capture these cross-feature effects (see \cref{app:empirical_details}, \cref{fig:cross_saliencies}).
We constrain it to be upper-triangular to account for rotational invariance in our model.
We also learn an initial value $w_0 \in \mathbb{R}$, so that $\bm{w}$ is initialised to $w_0 \bm{1}$ before simulating each training pipeline.
Like $\tau$, $S$ and $w_0$ are fixed while performing gradient ascent on the policy objective over preference functions, \cref{eq:modified_policy_objective}, but optimised in an outer loop to reduce the modelling loss, \cref{eq:modelling_loss}.
For details on our full algorithm including the hyperparameter fitting procedure, see \cref{app:alg_details}.

\subsection{Results}\label{sec:modelling_results}

\begin{figure}[t]
    \centering
    \includegraphics[width=0.8\linewidth]{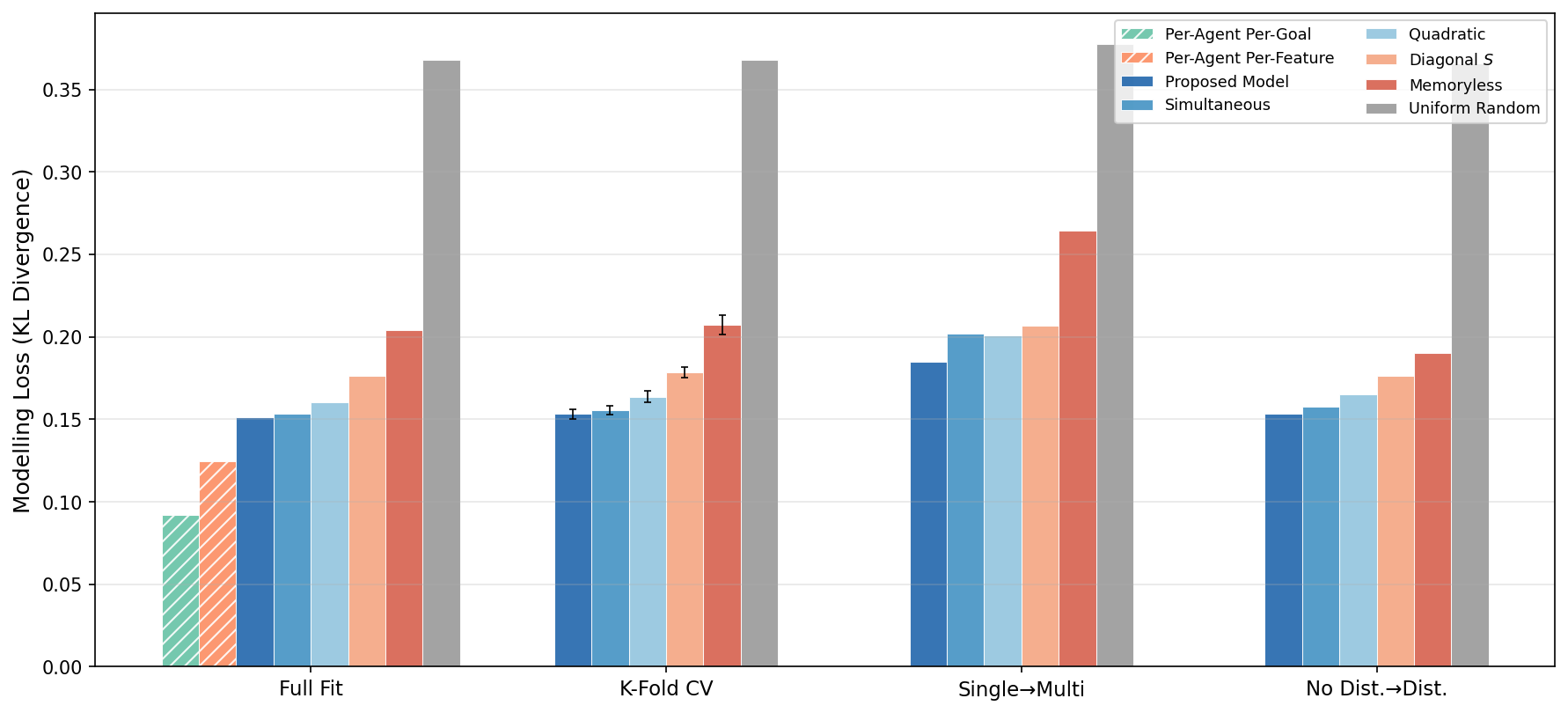}
    \caption{\textbf{Model comparison.} Average modelling loss (\cref{eq:modelling_loss}) across four evaluations (described in main text). Error bars show standard error for K-fold CV. Two per-agent lower bounds (Full Fit only) are also shown. Lower is better. Exact values are given in \cref{tab:model_results}.}
    \label{fig:model_comparison}
\end{figure}

To analyse the suitability of our model, we test it along with four alterations, a baseline, and two lower bounds across four different evaluations.
In each evaluation we fit the hyperparameters of the model to a subset of preference data, and then compute the modelling loss (\cref{eq:modelling_loss}) across some evaluation subset.

We compare against four alternatives: \textbf{Diagonal $S$}, which restricts the saliency matrix to be diagonal (no feature confusion); \textbf{Memoryless}, which fits only on the last training objective; \textbf{Simultaneous}, which ignores the ordering of training stages; and \textbf{Quadratic}, which expands the feature space to include pairwise interactions (with diagonal $S$ to limit parameters).%
\footnote{The expanded feature space contains $n$ base features, $n$ self-interaction terms ($\phi_i^2$), and $\binom{n}{2}$ unique cross-interaction terms ($\phi_i \phi_j = \phi_j \phi_i$ for $i < j$). Since each symmetric pair shares identical learning dynamics, the effective number of saliency parameters is $2n + \binom{n}{2} = 65$ for $n = 10$, giving 67 total hyperparameters.}
As a baseline, we consider the \textbf{uniform random} predictor, which predicts equal probability for each outcome (goal $a$, goal $b$, or neither).
We also report two loss lower bounds which fit values directly to each agent's observed preferences rather than predicting them from training pipelines.
The \textbf{per-agent per-feature} bound fits a value for each (agent, feature) combination and computes goal values as linear combinations, giving the lowest achievable loss for any method which assumes goal values decompose linearly over features---this includes our proposed model and all alternatives except Quadratic.
The \textbf{per-agent per-goal} bound fits an independent value for each (agent, goal) pair---equivalent to fitting Elo scores independently per agent---giving the lowest achievable loss for any Boltzmann-rational model.
Since these bounds fit values to observed preferences rather than predicting them from training pipeline descriptions, they are only applicable to the Full Fit evaluation and require substantially more parameters ($n_\text{agents} \times n_\text{goals}$ and $n_\text{agents} \times n_\text{features}$ respectively).
By contrast, our proposed model fits a single shared set of 57 hyperparameters across all 298 agents.

The four evaluations are: fitting the hyperparameters on the entire dataset and reporting the overall fit; performing 4-fold cross-validation (holding out 25\% of training pipelines for evaluation);
fitting hyperparameters on just the single-stage pipelines and reporting model fit on the two-stage pipelines; fitting hyperparameters on pipelines with no distractors and reporting model fit on the pipelines containing distractors.
These last two evaluations investigate how robust a model is when used to predict the effects of training pipelines that are qualitatively different from those used to fit its hyperparameters.

Our results are given in \cref{fig:model_comparison}.
Our proposed model achieves the lowest modelling loss across all evaluations.
Notably, a model fit only on single-stage pipelines successfully predicts the preferences of agents trained on two-stage pipelines, and a model fit without distractors generalises to pipelines containing them.
We also see from the alternatives that our model captures underlying patterns in the data better than one would be able to making the assumptions corresponding to each alternative, \emph{e.g.}, training is clearly not memoryless.
We do note that in contrast to the empirical evidence that ordering effects exist (\cref{app:empirical_details}, \cref{fig:ordering_effects}), the simultaneous alternative---which explicitly ignores training order---performs close to our proposed model.
This implies that while ordering effects are real, they account for a relatively small fraction of the total variation in preferences, and the specific form of our model perhaps does not yet capture their nature fully.

We additionally validate the fits using total variation distance, Brier score, and directional accuracy, finding that the model ranking is consistent across all metrics (\cref{app:additional_metrics}).
We give our fitted hyperparameters for the full dataset in \cref{app:fitted_hparams}, along with a brief interpretation of them.

\subsection{Interpretation of latent policy gradients}\label{sec:interpretting_model}

As shown in the previous section, the latent policy gradients method provides a good fit to the data.
However, we also claim that it is interpretable.
In this section, we substantiate this claim in two ways. Firstly, we demonstrate that the latent policy gradients dynamics correspond to a sequence of projection operations. Secondly, we show that the hyperparameters learned by our method---specifically the saliency matrix $S$---induce a similarity metric between features. 

\paragraph{Multi-stage training is iterated projection.} For the linear Boltzmann-rational preference function, \cref{eq:linear_boltzmann_rational}, the gradient of the modified policy objective \cref{eq:modified_policy_objective} can be computed analytically. In \cref{app:gradient_derivation}, we show that, in the case of no distractor object:
\begin{equation}\label{eq:latent_gradient}
    \nabla_\omega J = \left( 1 - \tau v^g \right) \sigma\left( v^g \right)\left( 1 - \sigma\left( v^g \right) \right) S^T \bm{\phi}^{(g)},  
\end{equation}
where $v^g = \bm{\phi}^{(g)} \cdot S\bm{w}$ is the value assigned to the goal object. This implies that the latents are exclusively updated in the direction $S^T \bm{\phi}^{(g)}$. Although we perform only a finite number of gradient updates, we can use \cref{eq:latent_gradient} to approximate the final weights by solving $\nabla_\omega J = 0$. This gives $v^g = \bm{\phi}^{(g)} \cdot S\bm{w} = \tau^{-1}.$ 
Therefore (see \cref{app:equilibrium_derivation}), if we begin with weights $\bm{w}_0$, the updated weights are
\begin{equation}\label{eq:analytic_solution}
    \bm{w} = \bm{w}_0 + \left( \frac{\tau^{-1} - \bm{\phi}^{(g)} \cdot S\bm{w}_0}{||S^T \bm{\phi}^{(g)}||^2_2} \right) S^T \bm{\phi}^{(g)}. 
\end{equation}
Consequently, multi-stage training can be viewed as iteratively projecting the weights onto the hyperplanes $\bm{\phi}^{(g)} \cdot S\bm{w} = \tau^{-1}$. We visualise this in \cref{fig:iterated_projection}. 

\definecolor{darkblue}{RGB}{0,70,140}
\definecolor{pastelblue}{RGB}{100,160,220}
\definecolor{darkred}{RGB}{180,60,0}
\definecolor{pastelred}{RGB}{255,150,150} 

\begin{figure}[h]
    \centering
    \resizebox{0.5\linewidth}{!}{
    \begin{tikzpicture}[
        >=Stealth,
        yscale=0.8,
        hyperplane/.style={thick},
        stage1/.style={->, thick, pastelblue},
        stage2/.style={->, thick, pastelred},
        alt/.style={->, thick, pastelred, dashed}, 
        point/.style={circle, fill, inner sep=1.5pt},
        eqlabel/.style={font=\scriptsize},
        label/.style={font=\small},
    ]
    
    \draw[->, gray!40, thin] (-0.5,0) -- (4.5,0);
    \draw[->, gray!40, thin] (0,-0.5) -- (0,4.2);
    
    \coordinate (origin) at (0,0);
    
    \pgfmathsetmacro{\nAx}{1}
    \pgfmathsetmacro{\nAy}{0.4}
    \pgfmathsetmacro{\nBx}{0.3}
    \pgfmathsetmacro{\nBy}{1}
    
    \pgfmathsetmacro{\cA}{2.5}
    \pgfmathsetmacro{\cB}{3.2}
    
    \pgfmathsetmacro{\normAsq}{\nAx*\nAx + \nAy*\nAy}
    \pgfmathsetmacro{\wAx}{\cA*\nAx/\normAsq}
    \pgfmathsetmacro{\wAy}{\cA*\nAy/\normAsq}
    
    \coordinate (w1) at (\wAx, \wAy);
    
    \pgfmathsetmacro{\normBsq}{\nBx*\nBx + \nBy*\nBy}
    \pgfmathsetmacro{\dotBwA}{\nBx*\wAx + \nBy*\wAy}
    \pgfmathsetmacro{\wBx}{\wAx + (\cB - \dotBwA)*\nBx/\normBsq}
    \pgfmathsetmacro{\wBy}{\wAy + (\cB - \dotBwA)*\nBy/\normBsq}
    
    \coordinate (w2) at (\wBx, \wBy);
    
    \pgfmathsetmacro{\wBpx}{\cB*\nBx/\normBsq}
    \pgfmathsetmacro{\wBpy}{\cB*\nBy/\normBsq}
    
    \coordinate (w2prime) at (\wBpx, \wBpy);
    
    \pgfmathsetmacro{\tAxcross}{-\wAy/\nAx}
    \pgfmathsetmacro{\HoneXaxisX}{\wAx + \tAxcross*(-\nAy)}
    \pgfmathsetmacro{\HoneXaxisY}{0}
    
    \pgfmathsetmacro{\tBycross}{\wBx/\nBy}
    \pgfmathsetmacro{\HtwoYaxisX}{0}
    \pgfmathsetmacro{\HtwoYaxisY}{\wBy + \tBycross*\nBx}
    
    \pgfmathsetmacro{\hAscaleUp}{2.4}  %
    \pgfmathsetmacro{\hAscaleDown}{1.1}
    \coordinate (H1top) at (\wAx - \hAscaleUp*\nAy, \wAy + \hAscaleUp*\nAx);
    \coordinate (H1bottom) at (\wAx + \hAscaleDown*\nAy, \wAy - \hAscaleDown*\nAx);
    
    \pgfmathsetmacro{\hBscaleRight}{1.8}
    \pgfmathsetmacro{\hBscaleLeft}{3.2}
    \coordinate (H2left) at (\wBx - \hBscaleLeft*\nBy, \wBy + \hBscaleLeft*\nBx);
    \coordinate (H2right) at (\wBx + \hBscaleRight*\nBy, \wBy - \hBscaleRight*\nBx);
    
    \draw[hyperplane, darkblue] (H1top) -- (H1bottom);
    \draw[hyperplane, darkred] (H2left) -- (H2right);
    
    \node[eqlabel, darkblue, below right] at (\HoneXaxisX, \HoneXaxisY) 
        {$\bm{\phi}^{(g_1)}.S\bm{w} = \tau^{-1}$};
    \node[eqlabel, darkred, above right, xshift=-1pt] at (\HtwoYaxisX, \HtwoYaxisY) 
        {$\bm{\phi}^{(g_2)}.S\bm{w} = \tau^{-1}$};
    
    \draw[stage1] (origin) -- (w1);
    \node[label, pastelblue, above, yshift=-1pt] at ($(origin)!0.5!(w1)$) {$S^T\bm{\phi}^{(g_1)}$};
    
    \draw[stage2] (w1) -- (w2);
    \node[label, pastelred, right, xshift=2pt] at ($(w1)!0.5!(w2)$) {$S^T\bm{\phi}^{(g_2)}$};
    
    \draw[alt] (origin) -- (w2prime);
    \node[label, pastelred, left, xshift=-2pt] at ($(origin)!0.5!(w2prime)$) {$S^T\bm{\phi}^{(g_2)}$};
    
    \node[point] at (origin) {};
    \node[below left, label] at (origin) {$\bm{w}_0$};
    
    \node[point, pastelblue] at (w1) {};
    \node[below right, label, pastelblue] at (w1) {$\bm{w}_1$};
    
    \node[point, pastelred] at (w2) {};
    \node[above right, label, pastelred] at (w2) {$\bm{w}_2$};
    
    \node[point, pastelred] at (w2prime) {};  %
    \node[below left, label, pastelred] at (w2prime) {$\bm{w}_2'$};
    
    \end{tikzpicture}
    }
    \caption{\textbf{Multi-stage training is iterated projection.} Latent policy gradient shifts the latent variables $\bm{w}$ in the $S^T \bm{\phi}^{(g)}$ direction until they intersect with the hyperplane $\bm{\phi}^{(g)} \cdot S\bm{w} = \tau^{-1}$. The result of training (to convergence) first on $\bm{\phi}^{(g_1)}$, and then on $\bm{\phi}^{(g_2)}$ is shown by $\bm{w}_2$. The result of training to convergence on $\bm{\phi}^{(g_2)}$ alone is shown by $\bm{w}_2'$.}
    \label{fig:iterated_projection}
\end{figure}
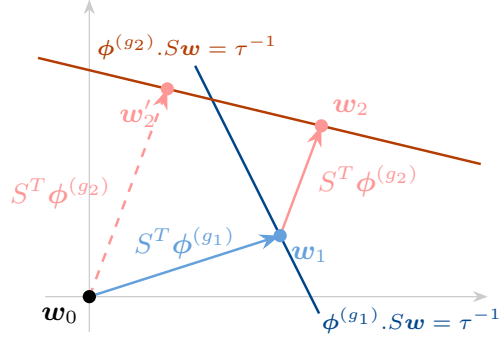

\paragraph{The latent saliency matrix induces a natural similarity metric over feature space.} Given \cref{eq:analytic_solution}, we can ask how training on a goal $\bm{\phi}^{(g)}$ effects the \emph{value} that the model assigns to another object, $\bm{\phi}'$, $\bm{\phi}' \cdot S\bm{w}$. Note from \cref{eq:linear_boltzmann_rational} that these values determine the preferences of the policy. If $\bm{w}_0 = 0$, the value assigned to $\bm{\phi}'$ is 
\begin{equation}\label{eq:similarity_weights}
    \tau^{-1} \bm{\phi}' \cdot SS^T \bm{\phi}^{(g)}/||S^T \bm{\phi}^{(g)}||^2_2.
\end{equation}
\Cref{eq:similarity_weights} tells us that training on one goal increases the value of other objects in accordance with their similarity with the goal, as measured by their inner product with respect to the matrix $SS^T$. We can therefore interpret $SS^T$ as an \emph{object similarity metric}, as judged by the policy network architecture.
In particular, we can interpret the diagonal elements of $SS^T$ as feature-specific learning rates---the relative strengths of each feature in determining generalisation.

\section{Discussion}

This paper formulates the goal generalisation problem as predicting OOD behaviour from training pipelines, and introduces latent policy gradients as a solution.
The best-performing parametrisation uses a saliency matrix to capture differing feature learning rates and cross-feature confusion---phenomena observed in our empirical analysis.
Beyond empirical fit, latent policy gradients admit interpretation: their dynamics correspond to iterated projection onto hyperplanes, and their hyperparameters define a feature similarity metric.
Unlike standard explainability methods, which characterise what a given model has learned, latent policy gradients predict how preferences form during training.

\paragraph{Neural Network Plasticity.}
Two of the phenomena we reported---value persistence and value inhibition---are consistent with plasticity effects in continual learning \citep{nikishin_primacy_2022,nikishin_deep_2023,dohare_loss_2024,lyle_disentangling_2024}, gradient starvation \citep{pezeshki_gradient_2021}, and the emergence of dormant neurons \citep{sokar_dormant_2023}.
We show the impact these effects have on goal generalisation in multi-stage goal-directed RL, and further show that they can be captured by a simple latent model with interpretable structure.

\paragraph{Limitations and future work.}
Our experiments used a single architecture and algorithm (PPO with a CNN) on a single domain (maze navigation) with one- or two-stage pipelines; extending to other architectures and algorithms (particularly action-value methods such as DQN \citep{mnih_playing_2013}), domains, and longer pipelines is important future work.
Our saliency findings (\emph{e.g.}, shape over colour) reflect CNN inductive biases, though the LPG framework itself is architecture-agnostic.
Our agents were each trained with a single random seed, though the large number of pipeline permutations provides symmetries that serve a similar role.
The preference function parametrisation we used assumes that goal values decompose linearly over features. This is well-suited to the factorial structure of our domain, but may not hold in settings where goals have more complex, non-compositional structure. The framework accommodates alternative parametrisations, but we have not yet explored them.
Finally, our predictions of agent preferences were not perfect, and we welcome future work which improves performance on the goal generalisation problem on our dataset.

\paragraph{Broader impacts.}
This work is motivated by AI safety: predicting how training pipelines shape OOD behaviour is necessary for safe deployment of RL systems. By providing interpretable methods for predicting goal generalisation, we aim to contribute to building more understandable AI systems. We do not foresee negative societal consequences from this foundational research.

\paragraph{Conclusion.}
The values of RL agents are path dependent, and prior training can significantly influence how a model generalises OOD.
However, this path-dependence has underlying structure. Latent policy gradients provide a means of predicting goal generalisation via the evolution of a small number of latent variables.
Our ultimate motivation is understanding how multi-stage training pipelines shape the values of frontier AI systems, and we see the problem formulation, evaluation methodology, and baseline method established here as a foundation for this goal.
The mathematical structure of latent policy gradients is general, and we believe it provides a strong foundation for extending this work to more complex systems.

\bibliography{references}
\bibliographystyle{plainnat}

\newpage
\appendix
\crefalias{section}{appendix}
\crefalias{subsection}{appendix}
\crefname{appendix}{Appendix}{Appendices}

\section{Related Work}\label{sec:related_work}

\paragraph{Transfer learning and AI alignment.}
Transfer learning is a common paradigm in modern machine learning \citep{iman_review_2023}.
The most notable example is the use of LLMs pre-trained on large text corpora and then RL fine-tuned to build helpful assistants \citep{bai_training_2022} or powerful reasoners \citep{deepseek-ai_deepseek-r1_2025}, but applications of foundation models are also being explored in robotics \citep{firoozi_foundation_2025}.
Systems trained via transfer learning are at the frontier of AI developments, with concerns over the safety of powerful AI systems \citep{amodei_concrete_2016,hubinger_risks_2021,ngo_alignment_2025,hendrycks_overview_2023} motivating a variety of complex approaches to try and better align these models with human values \citep{bai_constitutional_2022,guan_deliberative_2025,maiya_open_2025,rafailov_direct_2023}.

However, little is understood about how these complex training pipelines, from pre-training to alignment and reasoning post-training, end up shaping the final values and behaviours of the models they produce.
Despite seeming to work at the surface level, research on jailbreaks \citep{yi_jailbreak_2024}, emergent misalignment \citep{betley_emergent_2026,taylor_school_2025}, and misalignment in more agentic settings \citep{lynch_agentic_2025,andriushchenko_agentharm_2025,naik_agentmisalignment_2025} show there is still lots more work to be done.
That said, alignment approaches that consider the entire training pipeline are starting to be explored \citep{tice_alignment_2026}, though we lack a principled theory to understand the empirically observed effects.

One particular failure mode is \emph{goal-misgeneralisation} \citep{langosco_goal_2023}, where even a correctly specified reward function fails to induce the intended behaviour outside of the training distribution.
In our paper we try to better understand goal generalisation, and how it is influenced by multi-stage training pipelines.

\paragraph{Inferring the values of agents.}
In our work, we try to infer the values of AI agents behaviourally, and then try to predict them with just knowledge of the training pipeline.
Whilst our methods are simple, extensive work in the subfield of inverse RL (IRL) has detailed and analysed many more approaches for inferring latent values underpinning agentic behaviour, and the reward structures that might incentivise it \citep{wulfmeier_maximum_2016,skalse_misspecification_2023,ramachandran_bayesian_nodate,hadfield-menell_cooperative_2016,adams_survey_2022,ziebart_maximum_2008,abbeel_apprenticeship_2004,brown_learning_2025}.
However, this has mostly been restricted to robotics or game-playing tasks, and focuses on building reward functions, rather than understanding the dynamics of agent value formation.

More recent work on trying to infer the values of frontier AIs includes that of \citet{mazeika_utility_2025}, who analyse the preferences of LLMs expressed over various hypothetical scenarios, and try to infer their underlying values.
Focussing on more real-world settings, \citet{andriushchenko_agentharm_2025} analyses how LLMs can engage in harmful behaviours when given access to external tools, and \citet{lynch_agentic_2025,naik_agentmisalignment_2025} both show agentic models deployed to perform harmless tasks might engage in risky and malicious behaviour to achieve their goals.

\citet{mini_understanding_2023} uses mechanistic interpretability \citep{bereska_mechanistic_2024} to discover motivational-circuits within a maze-solving agent, showing how methods other than evaluating behaviour can be used to infer the value structure of an AI agent.
Extending this philosophy, we directly consider the relationship between learnt values and the training pipeline that produced them, and believe this is an untapped well of potential insight.

\paragraph{Generalisation in machine learning.}
Beyond values and goals, there is a large body of existing literature on how RL systems generalise \citep{cobbe_quantifying_2019,zhao_investigating_2019,kirk_survey_2023}, though this is mostly focussed on generalisation of agent capabilities, rather than their values.
Prominent works highlight the issue of overfitting, and how to deal with it \citep{cobbe_quantifying_2019,cobbe_leveraging_2020}.
This motivated our use of procedural generation for our tasks, to force our agents to generalise their capabilities as best they could to novel mazes with unfamiliar goals.

In accordance with our results on value persistence and strengthening, other works have found that early training data can have outsized impacts on the learning of deep RL agents \citep{nikishin_primacy_2022}, and that when models are successfully architected to transfer learn successfully, often existing low-level representations are re-used \citep{rusu_progressive_2022}.
A related and rapidly growing literature investigates the \emph{loss of plasticity} in continual and multi-stage neural network training---the tendency of networks to lose their ability to adapt to new data as training progresses \citep{dohare_loss_2024,lyle_disentangling_2024,nikishin_deep_2023}.
Relatedly, \citet{pezeshki_gradient_2021} identify \emph{gradient starvation}, whereby easy-to-learn features dominate gradient updates and suppress learning of other informative features---a phenomenon that parallels our finding that salient features inhibit the formation of new values.
This has been linked to mechanisms such as the emergence of dormant neurons in deep RL networks \citep{sokar_dormant_2023}.
Our findings on value persistence and value inhibition are consistent with these phenomena, and we view our latent policy gradients method as providing a complementary, predictive account of their effects in the specific setting of multi-stage goal-directed RL with compositional goals.

Additionally, our results surrounding the importance of feature salience draws parallels with previous work showing how deep learning models often exploit shortcuts and spurious correlations in their training data in order to achieve high performance on a task, as these are often easier than generalising robustly \citep{geirhos_shortcut_2020,haan_causal_2019}.
Contrary to this, more contemporary work suggests that extended training might overcome this effect \citep{power_grokking_2022,nakkiran_deep_2021}, and that if we trained our models for a far greater amount of time then feature saliencies and previously learnt values might become less important in determining generalisation behaviour. For CNNs specifically, prior work has found they have a bias for categorising based on shape rather than colour \citep{ritter_cognitive_2017,feinman_learning_2018}, consistent with our saliency analysis.

\paragraph{Behavioural psychology and conditioning.}
Our work has strong connections to several neuroscientific works.
Our latent policy gradient method, and the dynamics of value learning it predicts, bears strong resemblance to the Rescorla-Wagner model of Pavlovian conditioning, and similarly predicts dynamics such as blocking and overshadowing \citep{rescorla_theory_1972}.
Other theories of conditioning, such as Configural theory \citep{pearce_model_1987}, diverge from our specific parametrisation (though do not conflict with the intuition behind latent policy gradients at a high level), and might be interesting sources of inspiration for future models of agent values.
Finally, several works investigate stimulus generalisation in a way that bears resemblance to how we think about feature confusion, and the non-diagonal entries of the saliency matrix \citep{shepard_toward_1987,mclaren_associative_2002}.

\section{Experimental Details}

\subsection{Agent}\label{app:exp_details_agent}

Our agent is implemented using Stable Baselines 3 \citep{raffin_stable-baselines3_2021}.
We use the PPO algorithm \citep{schulman_proximal_2017} with a standard CNN-based policy \citep{mnih_human-level_2015}, our hyperparameters are given in \cref{tab:ppo_hparams}.
The policy uses a shared CNN feature extractor, and then separate heads for the actor and critic networks respectively:
\begin{verbatim}
ActorCriticCnnPolicy(
  (features_extractor): NatureCNN(
    (cnn): Sequential(
      (0): Conv2d(3, 32, kernel_size=(8, 8), stride=(4, 4))
      (1): ReLU()
      (2): Conv2d(32, 64, kernel_size=(4, 4), stride=(2, 2))
      (3): ReLU()
      (4): Conv2d(64, 64, kernel_size=(3, 3), stride=(1, 1))
      (5): ReLU()
      (6): Flatten(start_dim=1, end_dim=-1)
    )
    (linear): Sequential(
      (0): Linear(in_features=9216, out_features=512, bias=True)
      (1): ReLU()
    )
  )
  (action_net): Linear(in_features=512, out_features=4, bias=True)
  (value_net): Linear(in_features=512, out_features=1, bias=True)
)
\end{verbatim}
The network has a total of 4.8 million parameters, all of which are randomly initialised for the first stage of training.

\begin{table}[h]
\centering
\caption{PPO hyperparameters used for training all agents.}
\label{tab:ppo_hparams}
\begin{tabular}{ll}
\toprule
\textbf{Hyperparameter} & \textbf{Value} \\
\midrule
Learning rate & 0.0001 \\
Steps per rollout & 256 \\
Batch size & 64 \\
Epochs per update & 10 \\
Discount factor ($\gamma$) & 0.95 \\
GAE lambda & 0.9 \\
Clip range & 0.2 \\
Entropy coefficient & 0.01 \\
Value function coefficient & 0.5 \\
Max gradient norm & 0.5 \\
\bottomrule
\end{tabular}
\end{table}

\FloatBarrier
\subsection{The Maze Environment}\label{app:exp_details_maze}

To generate mazes, we sample $8 \times 8$ grids where each cell is a wall independently with probability 0.2. Mazes with inaccessible vacant squares are discarded and re-generated. The goal object and agent are placed in random vacant cells. Each cell is rendered as a $16 \times 16$ pixel image, giving the agent $128 \times 128$ RGB observations.

\begin{figure}[h]
    \centering
    \includegraphics[width=0.4\linewidth]{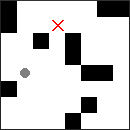}
    \caption{\textbf{An example 8x8 maze environment.} The agent is represented by \sym{grey}{circle} and the \sym{red}{cross} is the goal object. Black squares are impassable walls. The agent cannot move off the edges of the maze; the outer wall shown is for illustrative purposes and is not included as part of the agent's observation. The total observation size is $128\times128$ pixels with 3 colour channels (RGB).}
    \label{fig:maze_example}
\end{figure}

\begin{figure}[htbp]
  \centering
  \includegraphics[width=0.6\textwidth]{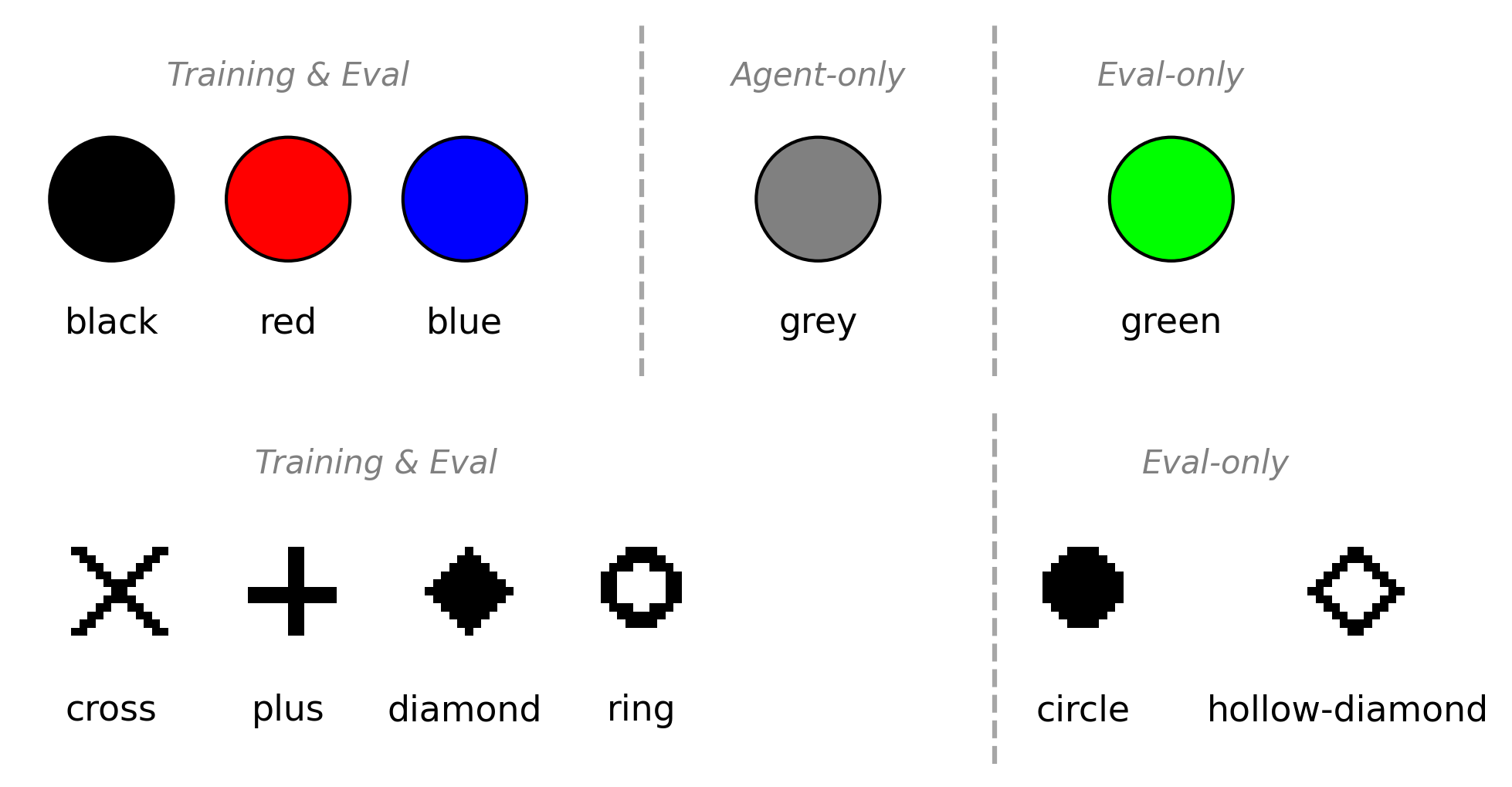}
  \caption{\textbf{Visual features used in the Maze environment}.
    \textbf{Colours} (top): Three colours (black, red, blue) are used for goal objects during both training and evaluation; grey is used exclusively to render the agent; green appears only in evaluation environments to test generalisation to novel colours. All red, blue, and green use a single RGB input channel. 
    \textbf{Shapes} (bottom): Four shapes (cross, plus, diamond, ring) are used during training and evaluation; circle and hollow-diamond appear only in evaluation to test generalisation to novel shapes.
    Each cell of the maze is rendered as a $16 \times 16$ pixel image, thus the rendered shapes have a low resolution.}
  \label{fig:features}                                                                             
\end{figure} 

\FloatBarrier

\FloatBarrier
\section{Methodological Details}\label{app:alg_details}

\subsection{Elo Score Computation}\label{app:elo_fitting}

Under the Elo model, each object receives a scalar score $V$, and the probability that an agent selects object $A$ over object $B$ is given by $1 / (1 + 10^{-(V_A - V_B)/400})$.
This is equivalent to a logistic function of the score difference with a growth rate of $\frac{\ln 10}{400}$, and can be derived from the Bradley-Terry model of pairwise comparisons \citep{bradley_rank_1952}.

Since our evaluation data includes a third outcome of the episode terminating without either object being reached, we convert each three-way observation into three pairwise comparisons.
This is done by masking out each outcome in turn and normalising the remaining two rates.
A ``no goal'' baseline is included as a competitor in the Elo system.
After fitting, we shift all scores so that the no-goal score is zero, utilising Elo's invariance to uniform shifts.
This ensures that each object's Elo score reflects both its relative preference and its absolute likelihood of being pursued, while remaining comparable across agents.

\subsection{Latent Policy Gradient Fitting}\label{app:lpg_fitting}

We detail the model fitting procedure in \cref{alg:lpg_fitting}.
The outer loop optimises the hyperparameters $S$, $\tau$, and $w_0$\footnote{In practice, we store $\log \tau$ rather than $\tau$ directly, ensuring $\tau > 0$ and improving numerical stability during optimisation.} by minimising the modelling
loss (\cref{eq:modelling_loss}) via gradient descent.
To construct $\mathcal{D}_\text{train}$ we add a separate entry for each evaluation environment of each agent, so for our 298 agents each with 276 evaluation environments we obtain 82k datapoints.

For each training example, \cref{alg:simulate_pipeline} computes the latent weights $\bm{w}_\Omega$ by simulating gradient ascent with $n_\text{integration-steps}=100$ on the modified policy objective (\cref{eq:modified_policy_objective}) through each stage of the training pipeline.
Since this inner simulation is differentiable with respect to the hyperparameters, we're able to optimise them with gradient descent in the outer loop.

The algorithm is generic across different validation scenarios.
For full-data fitting, $\mathcal{D}_\text{train} = \mathcal{D}_\text{eval}$.
For K-fold cross-validation, agents are partitioned and the procedure is repeated K times with different train/eval splits. For transfer evaluations (single-stage $\to$ multi-stage, or no-distractor $\to$ distractor), training uses only the simpler pipelines while evaluation uses the more complex ones.

We have given \cref{alg:lpg_fitting} in the form of full-batch gradient descent for simplicity, however, in practice we mini-batch with batch size 64 and use the Adam optimiser \citep{kingma_adam_2014} with a learning rate of 0.03 and a $\beta_1$ and $\beta_2$ of 0.9 and 0.999 respectively.
We perform only a single epoch as we found it gave similar performance to more epochs, but was much faster.

\begin{algorithm}[h]
\caption{Fitting the Latent Policy Gradient Model}\label{alg:lpg_fitting}
\begin{algorithmic}[1]
\REQUIRE Training data $\mathcal{D}_\text{train} = \{(\Omega_i, \bm{\phi}^{(a)}_i, \bm{\phi}^{(b)}_i, \hat{\Pi}_i)\}_{i=1}^N$, where $\Omega_i$ is a training pipeline, $(\bm{\phi}^{(a)}_i, \bm{\phi}^{(b)}_i)$ is an evaluation goal pair, and $\hat{\Pi}_i$ is the observed choice distribution
\REQUIRE Evaluation data $\mathcal{D}_\text{eval}$ (same format)
\REQUIRE Learning rate $\eta$, number of epochs $E$, number of features $n$
\ENSURE Fitted saliency matrix $S$, temperature $\tau$, initial value $w_0$; evaluation loss $\mathcal{L}_\text{eval}$

\STATE \textbf{Initialise:} $S \gets I_n$ (upper-triangular), $\tau \gets 1$, $w_0 \gets 0$

\FOR{epoch $= 1, \ldots, E$}
    \STATE $\mathcal{L}_\text{train} \gets 0$
    \FOR{each $(\Omega, \bm{\phi}^{(a)}, \bm{\phi}^{(b)}, \hat{\Pi}) \in \mathcal{D}_\text{train}$}
        \STATE $\bm{w}_\Omega \gets \textsc{SimulatePipeline}(\Omega, S, \tau, w_0)$ \COMMENT{Alg.~\ref{alg:simulate_pipeline}}
        \STATE $\Pi_\Omega \gets \text{softmax}\left([\bm{\phi}^{(a)} \cdot S\bm{w}_\Omega, \, \bm{\phi}^{(b)} \cdot S\bm{w}_\Omega, \, 0]\right)$
        \STATE $\mathcal{L}_\text{train} \gets \mathcal{L}_\text{train} + D_\text{KL}\left( \hat{\Pi} \,\|\, \Pi_\Omega \right)$
    \ENDFOR
    \STATE Update $S, \tau, w_0$ via gradient descent on $\mathcal{L}_\text{train}$ with learning rate $\eta$
\ENDFOR

\STATE \textbf{Evaluate:} $\mathcal{L}_\text{eval} \gets 0$
\FOR{each $(\Omega, \bm{\phi}^{(a)}, \bm{\phi}^{(b)}, \hat{\Pi}) \in \mathcal{D}_\text{eval}$}
    \STATE $\bm{w}_\Omega \gets \textsc{SimulatePipeline}(\Omega, S, \tau, w_0)$
    \STATE $\Pi_\Omega \gets \text{softmax}\left([\bm{\phi}^{(a)} \cdot S\bm{w}_\Omega, \, \bm{\phi}^{(b)} \cdot S\bm{w}_\Omega, \, 0]\right)$
    \STATE $\mathcal{L}_\text{eval} \gets \mathcal{L}_\text{eval} + D_\text{KL}\left( \hat{\Pi} \,\|\, \Pi_\Omega \right)$
\ENDFOR
\STATE $\mathcal{L}_\text{eval} \gets \mathcal{L}_\text{eval} / |\mathcal{D}_\text{eval}|$

\STATE \textbf{return} $S$, $\tau$, $w_0$, $\mathcal{L}_\text{eval}$\end{algorithmic}
\end{algorithm}

\begin{algorithm}[h]
\caption{\textsc{SimulatePipeline}: Compute Latent Weights via Policy Gradient Ascent}\label{alg:simulate_pipeline}
\begin{algorithmic}[1]
\REQUIRE Pipeline $\Omega = \{(\bm{\phi}^{(g_t)}, \bm{\phi}^{(d_t)})\}_{t=1}^T$ with $T$ stages, each having goal features $\bm{\phi}^{(g_t)}$ and optional distractor features $\bm{\phi}^{(d_t)}$
\REQUIRE Saliency matrix $S$, temperature $\tau$, initial value $w_0$
\ENSURE Latent weights $\bm{w}$

\STATE $\bm{w} \gets w_0 \bm{1}$
\FOR{stage $t = 1, \ldots, T$}
    \FOR{step $= 1, ..., n_\text{integration-steps}$}
        \STATE $v^g \gets \bm{\phi}^{(g_t)} \cdot S\bm{w}$ \COMMENT{Goal value}
        \IF{distractor $\bm{\phi}^{(d_t)}$ present}
            \STATE $v^d \gets \bm{\phi}^{(d_t)} \cdot S\bm{w}$ \COMMENT{Distractor value}
            \STATE $\pi_G \gets \exp(v^g) / (1 + \exp(v^g) + \exp(v^d))$
        \ELSE
            \STATE $\pi_G \gets \sigma(v^g)$ \COMMENT{Goal probability}
        \ENDIF
        \STATE $J \gets \pi_G + \tau \, h(\pi_G)$ \COMMENT{Modified policy objective, \cref{eq:modified_policy_objective}}
        \STATE $\bm{w} \gets \bm{w} + \nabla_{\bm{w}} J$ \COMMENT{Gradient ascent step}
    \ENDFOR
\ENDFOR
\STATE \textbf{return} $\bm{w}$ \end{algorithmic}
\end{algorithm}

\FloatBarrier

\section{Fitted Model Parameters}
\label{app:fitted_hparams}

Here we present the fitted hyperparameters for our proposed model when fit on all our data.

For the entropy regularisation coefficient, $\tau$, we obtain a value of 0.698, showing our agents were generally modelled as moderately effective at achieving their goals.
For the initial latent value, $w_0$, we obtain a value of $-0.372$, indicating that agents begin with a slight prior against pursuing any goal, consistent with untrained agents not yet having learned goal-directed behaviour.
The entries of $S$ and $SS^T$ are given in \cref{tab:saliency-matrix,tab:sst-matrix} respectively.
$S$ alone is not very interpretable, but as discussed in \cref{sec:interpretting_model}, $SS^T$ induces an interpretable natural similarity metric over features.
In agreement with the empirical results, we see that shapes are stronger than colours at driving generalisation, with cross being by far the strongest. Hollow-diamond is the weakest shape and black is the weakest feature overall.

\begin{table}[h]
\centering
\caption{\textbf{Saliency Matrix}, $S$ (upper-triangular). This maps between latent space and goal-feature space.}
\label{tab:saliency-matrix}
\resizebox{\textwidth}{!}{%
\begin{tabular}{l|rrrrrrrrrr}
\toprule
$S_{ij}$ & \multicolumn{10}{c}{$\leftarrow \bm{w} \rightarrow$} \\
\midrule
black & 0.994 & 0.349 & 0.289 & 0.142 & $-$0.253 & 0.065 & 0.461 & 0.092 & 0.239 & 0.157 \\
blue & 0 & 2.088 & $-$0.379 & $-$0.318 & 0.188 & 0.099 & 0.114 & 0.341 & 0.206 & 0.133 \\
green & 0 & 0 & 1.549 & $-$0.394 & 0.131 & 0.325 & 0.097 & 0.116 & 0.211 & 0.076 \\
red & 0 & 0 & 0 & 2.080 & 0.044 & 0.011 & 0.183 & 0.216 & 0.277 & 0.181 \\
\midrule
circle & 0 & 0 & 0 & 0 & 1.445 & $-$0.084 & 0.689 & 0.417 & 0.115 & 0.100 \\
cross & 0 & 0 & 0 & 0 & 0 & 2.794 & $-$0.476 & $-$0.468 & $-$0.536 & $-$0.587 \\
diamond & 0 & 0 & 0 & 0 & 0 & 0 & 1.781 & 0.157 & 0.280 & $-$0.476 \\
hollow-dia. & 0 & 0 & 0 & 0 & 0 & 0 & 0 & 1.157 & $-$0.158 & 0.294 \\
plus & 0 & 0 & 0 & 0 & 0 & 0 & 0 & 0 & 2.428 & $-$0.470 \\
ring & 0 & 0 & 0 & 0 & 0 & 0 & 0 & 0 & 0 & 2.478 \\
\bottomrule
\end{tabular}%
}
\end{table}

\begin{table}[h]
\centering
\caption{\textbf{Induced similarity metric}, $SS^\top$ This captures the effective learning rate interactions between features. Diagonal entries (shown in bold) represent the relative strengths of features, while off-diagonal entries represent cross-feature interactions.}
\label{tab:sst-matrix}
\resizebox{\textwidth}{!}{%
\begin{tabular}{l|rrrr|rrrrrr}
\toprule
$[SS^\top]_{ij}$ & black & blue & green & red & circle & cross & diamond & hollow-dia. & plus & ring \\
\midrule
black & \textbf{1.585} & 0.685 & 0.498 & 0.485 & 0.028 & $-$0.300 & 0.828 & 0.114 & 0.507 & 0.389 \\
blue & 0.685 & \textbf{4.838} & $-$0.300 & $-$0.477 & 0.522 & $-$0.125 & 0.250 & 0.402 & 0.437 & 0.331 \\
green & 0.498 & $-$0.300 & \textbf{2.750} & $-$0.695 & 0.309 & 0.649 & 0.213 & 0.123 & 0.476 & 0.189 \\
red & 0.485 & $-$0.477 & $-$0.695 & \textbf{4.516} & 0.329 & $-$0.414 & 0.352 & 0.259 & 0.587 & 0.449 \\
\midrule
circle & 0.028 & 0.522 & 0.309 & 0.329 & \textbf{2.767} & $-$0.877 & 1.277 & 0.493 & 0.231 & 0.249 \\
cross & $-$0.300 & $-$0.125 & 0.649 & $-$0.414 & $-$0.877 & \textbf{8.882} & $-$0.792 & $-$0.630 & $-$1.025 & $-$1.456 \\
diamond & 0.828 & 0.250 & 0.213 & 0.352 & 1.277 & $-$0.792 & \textbf{3.500} & $-$0.003 & 0.903 & $-$1.178 \\
hollow-dia. & 0.114 & 0.402 & 0.123 & 0.259 & 0.493 & $-$0.630 & $-$0.003 & \textbf{1.449} & $-$0.522 & 0.729 \\
plus & 0.507 & 0.437 & 0.476 & 0.587 & 0.231 & $-$1.025 & 0.903 & $-$0.522 & \textbf{6.115} & $-$1.166 \\
ring & 0.389 & 0.331 & 0.189 & 0.449 & 0.249 & $-$1.456 & $-$1.178 & 0.729 & $-$1.166 & \textbf{6.141} \\
\bottomrule
\end{tabular}%
}
\end{table}

\FloatBarrier

\section{Latent Dimension Analysis}
\label{app:latent_dimension}

\begin{figure}[h]
    \centering
    \includegraphics[width=\linewidth]{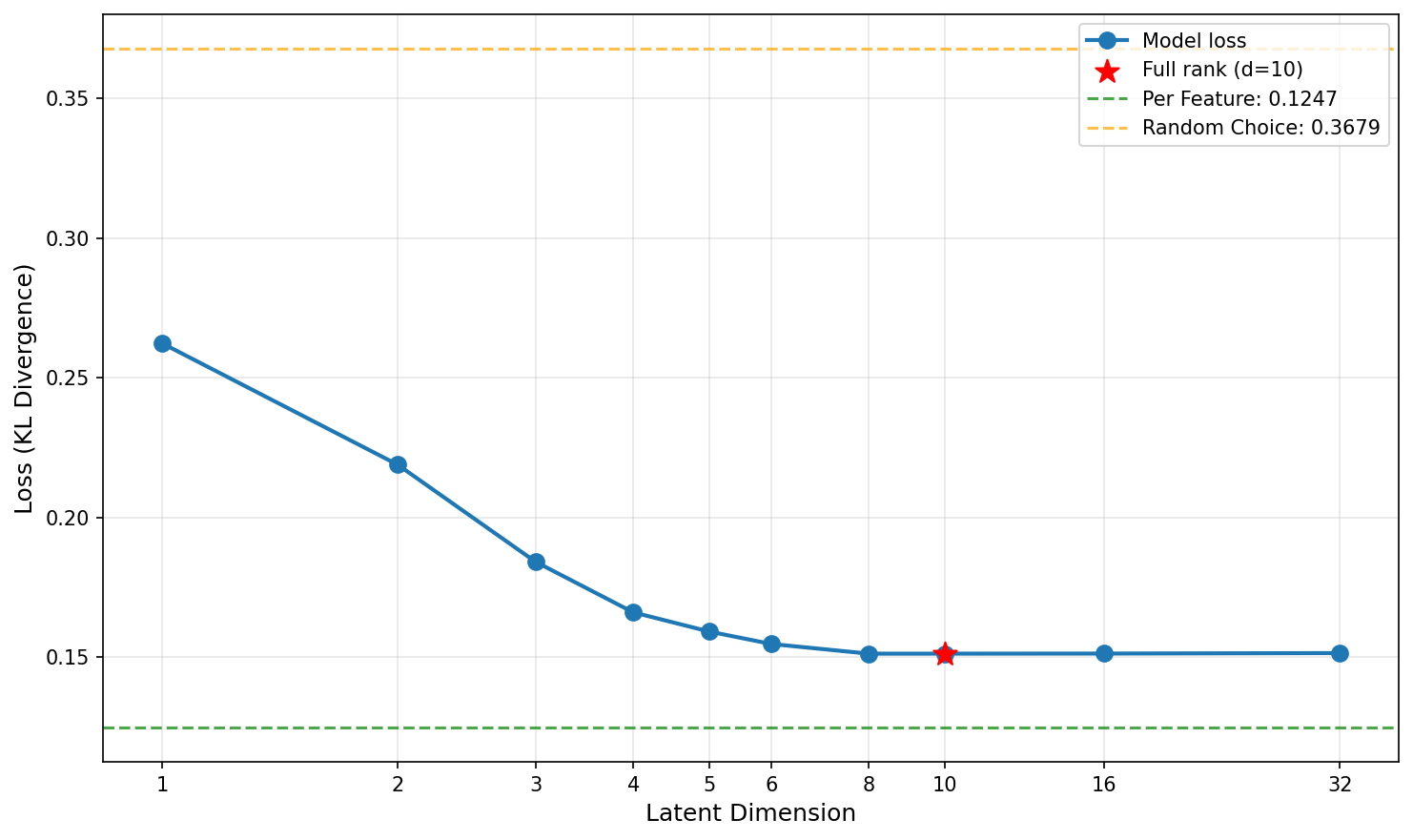}
    \caption{Model loss (KL divergence) as a function of the latent dimension $d$. The dashed lines show the uniform random baseline (upper) and the per-agent per-feature lower bound (lower). Loss plateaus at $d = 8$; the red star marks $d = 10 = n_\text{features}$, the value used in the main text.}
    \label{fig:latent_dim_sweep}
\end{figure}

We sweep the latent dimension $d$ (the size of the weight vector $\bm{w}$) from 1 to 32, keeping all other hyperparameters at their default values.
\Cref{fig:latent_dim_sweep} shows that loss decreases monotonically up to $d = 8$ and plateaus thereafter, with no benefit from dimensions beyond $n_\text{features} = 10$.
We use $d = 10$ in the main text, which sits at the plateau and matches the number of object features, giving each latent dimension a natural interpretation.
The clean plateau suggests that the underlying preference structure has moderate effective dimensionality and that the choice of $d$ is not sensitive provided $d \geq n_\text{features}$.

\FloatBarrier

\section{Additional Model Evaluation Metrics}
\label{app:additional_metrics}

Throughout this appendix, we evaluate model predictions using the following metrics.
We report these in two variants: over the full \textbf{three-way} distribution (goal $a$, goal $b$, neither), which is the distribution our models are trained to predict; and over a renormalised \textbf{two-way} distribution (goal $a$ vs goal $b$), which removes the null-goal component and isolates how well the model captures \emph{relative} goal preferences, independent of its ability to predict the overall goal-reaching rate.

For a single evaluation example with observed distribution $\hat{p} = (\hat{p}_1, \ldots, \hat{p}_K)$ and predicted distribution $q = (q_1, \ldots, q_K)$ where $K$ is either 3 (three-way) or 2 (two-way):

\textbf{KL divergence} (training objective in the main text):
\begin{equation}
    D_\text{KL}(\hat{p} \| q) = \sum_{k=1}^{K} \hat{p}_k \log \frac{\hat{p}_k}{q_k}.
\end{equation}

\textbf{Total variation (TV) distance}, which provides a scale-interpretable bound on the maximum prediction error for any event:
\begin{equation}
    \text{TV}(\hat{p}, q) = \frac{1}{2} \sum_{k=1}^{K} |\hat{p}_k - q_k|.
\end{equation}

\textbf{Brier score}, the mean squared error between predicted and observed probabilities, which penalises confident incorrect predictions:
\begin{equation}
    \text{BS}(\hat{p}, q) = \frac{1}{K}\sum_{k=1}^{K} (\hat{p}_k - q_k)^2.
\end{equation}

\textbf{Directional accuracy}, the fraction of non-trivial evaluation pairs\footnote{We exclude pairs where the observed normalised rate difference is less than 10 percentage points, as these are near-ties where directional prediction is not meaningful.} for which the model correctly predicts which goal the agent prefers:
\begin{equation}
    \text{Dir. Acc.} = \frac{1}{|\mathcal{D}'|}\sum_{i \in \mathcal{D}'} \mathbf{1}\left[\operatorname{sign}(q_1^{(i)} - q_2^{(i)}) = \operatorname{sign}(\hat{p}_1^{(i)} - \hat{p}_2^{(i)})\right],
\end{equation}
where $\mathcal{D}'$ is the set of examples exceeding the directional threshold.
All metrics are averaged over all evaluation examples. Lower is better for KL, TV, and Brier; higher is better for directional accuracy.

\subsection{Elo Holdout Validation}
\label{app:elo_holdout}

To validate that Elo scores reliably summarise the pairwise preference data, we perform 4-fold cross-validation.
For each agent, we partition the 276 pairwise comparisons into 4 folds, fit Elo scores on 3 folds, and predict win probabilities on the held-out fold using $P(\text{goal}_a) = 1 / (1 + 10^{-(V_a - V_b)/400})$.
We report the mean and standard error of each metric across all 298 agents.

\begin{table}[h]
\centering
\caption{Elo 4-fold cross-validation results (297 agents). Elo scores fitted on 75\% of pairwise comparisons accurately predict the held-out 25\%.}
\label{tab:elo_holdout}
\begin{tabular}{lcc}
\toprule
\textbf{Metric} & \textbf{Mean} & \textbf{SE} \\
\midrule
KL divergence & 0.0511 & 0.0010 \\
Total variation distance & 0.0922 & 0.0007 \\
Brier score & 0.0339 & 0.0006 \\
Directional accuracy & 89.73\% & 0.18\% \\
\bottomrule
\end{tabular}
\end{table}

As shown in \cref{tab:elo_holdout}, Elo scores are a strong predictor of held-out pairwise outcomes: the model correctly predicts which goal is preferred in nearly 90\% of non-trivial matchups, with low KL divergence and well-calibrated probability estimates (Brier score 0.034).

\subsection{Elo Scores vs Model-Predicted Values}
\label{app:elo_vs_model}

\begin{figure}[h]
    \centering
    \includegraphics[width=0.75\linewidth]{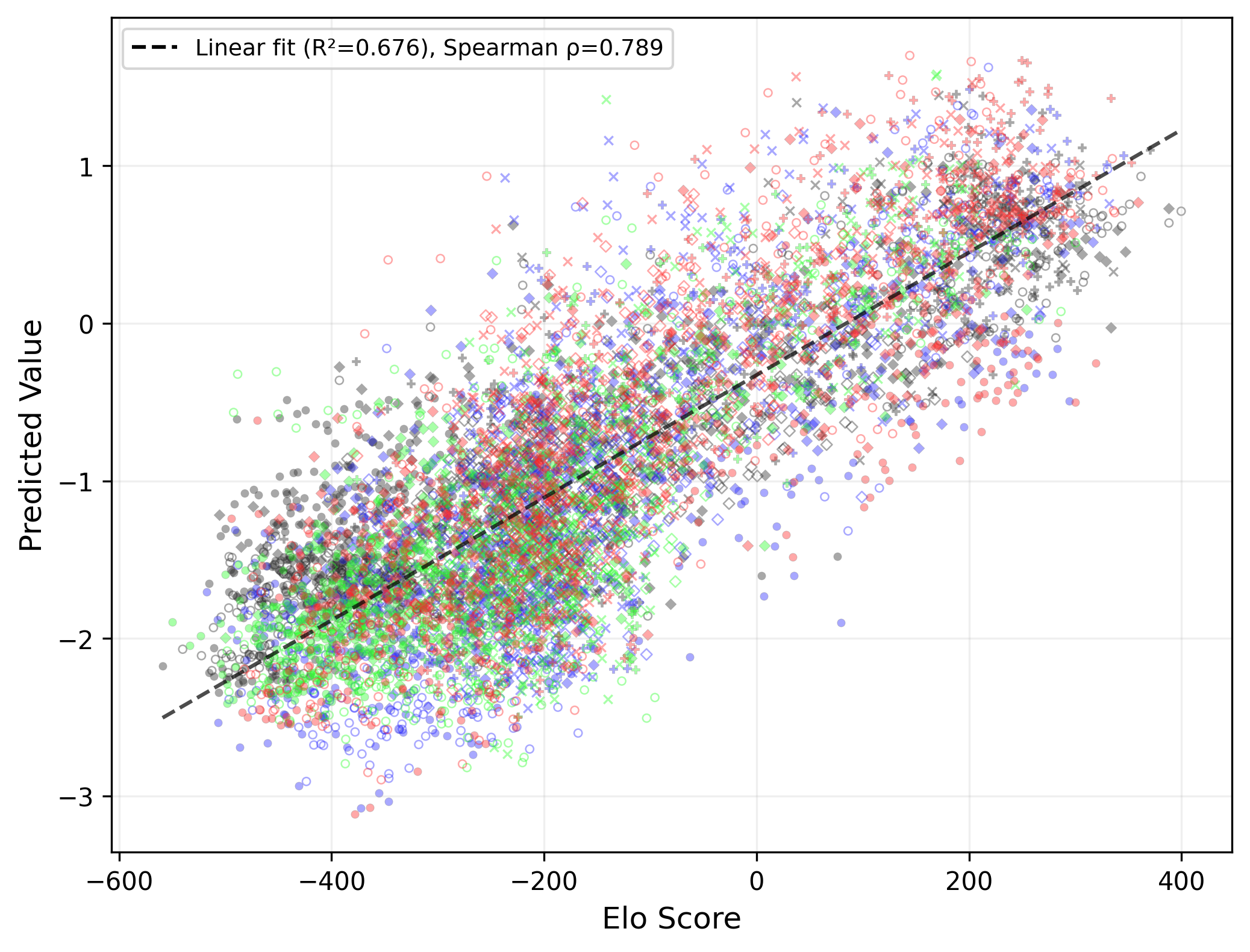}
    \caption{Empirical Elo scores vs model-predicted values for all 298 agents across 24 goals. Each point is an (agent, goal) pair, coloured by the goal's colour and shaped by the goal's shape.}
    \label{fig:elo_vs_model_raw}
\end{figure}

As a further validation that our model captures meaningful preference structure, we compare each agent's empirical Elo scores with the values predicted by our proposed model.
\Cref{fig:elo_vs_model_raw} shows the unnormalised comparison, where each point is an (agent, goal) pair.
The strong positive correlation (Spearman $\rho = 0.789$, $R^2 = 0.676$, $n = 7{,}152$) confirms that the model's predicted values track the empirically observed preference rankings.
This plot reflects the full three-way evaluation: the absolute value levels encode information about the agent's overall goal-reaching rate (analogous to the no-goal probability), in addition to relative goal preferences.

\begin{figure}[h]
    \centering
    \includegraphics[width=0.75\linewidth]{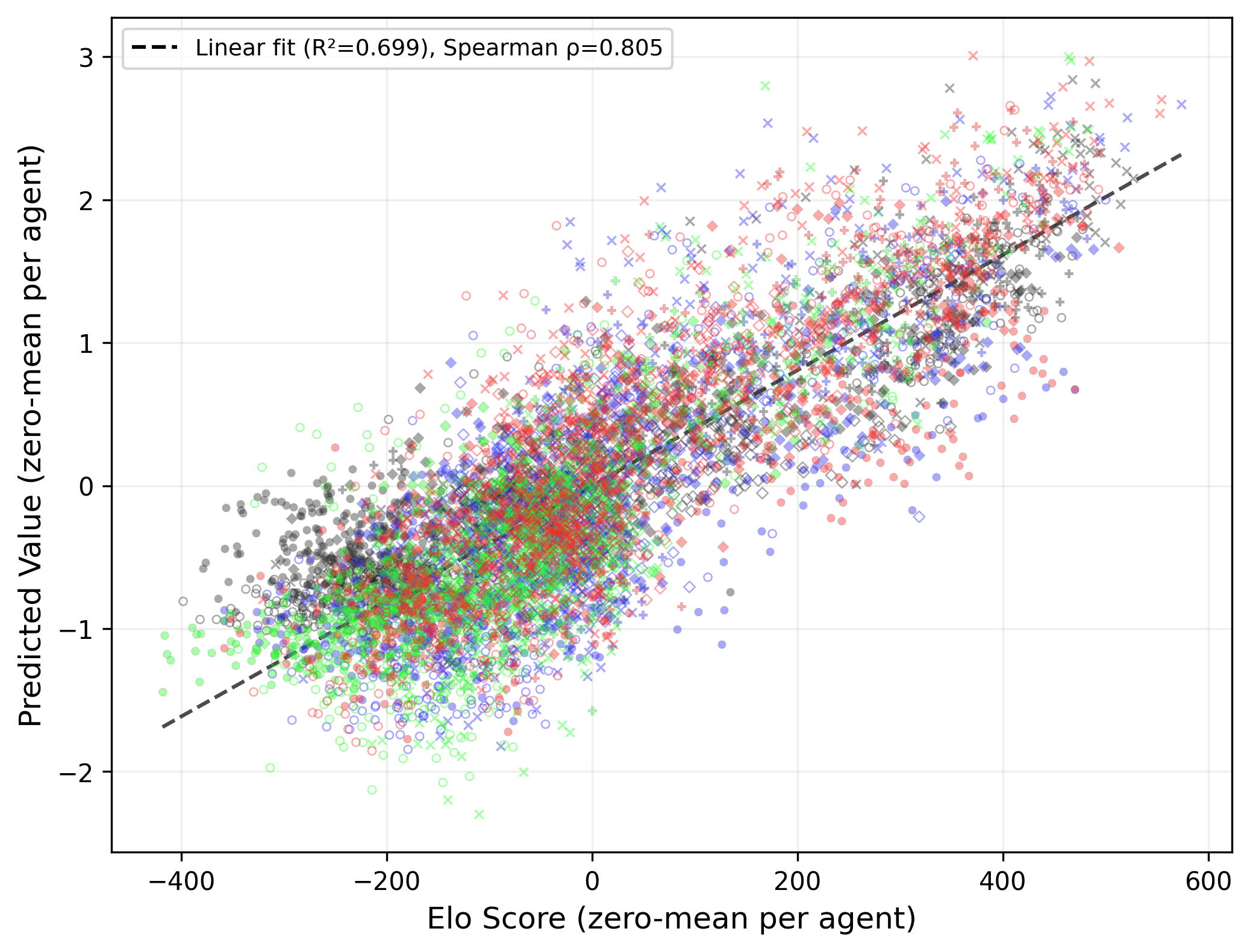}
    \caption{Zero-mean normalised Elo scores vs model-predicted values. Per-agent means are subtracted from both axes, isolating relative goal preferences within each agent.}
    \label{fig:elo_vs_model_norm}
\end{figure}

\Cref{fig:elo_vs_model_norm} shows the same data after subtracting the per-agent mean from both Elo scores and model values.
This zero-mean normalisation removes the per-agent baseline, isolating how well the model captures \emph{relative} goal preferences within each agent---analogous to the two-way evaluation that ignores no-goal probabilities.
The improved correlation (Spearman $\rho = 0.805$, $R^2 = 0.699$) indicates that the model captures relative preferences somewhat better than absolute value levels.
In both plots, points are coloured by goal colour and shaped by goal shape, and the visual clustering confirms that the model learns meaningful colour and shape representations.

\subsection{Supplementary Evaluation Metrics}
\label{app:supplementary_metrics}

\Cref{tab:model_results} gives the exact numerical values corresponding to \cref{fig:model_comparison} in the main text.

\begin{table}[h]
\centering
\caption{Average modelling loss (\cref{eq:modelling_loss}) across our proposed model, four alternatives, a baseline, and two lower bounds, for a variety of evaluations. Lower is better, \textbf{best} model in column is shown in bold. For K-fold Cross Validation we give the standard error computed over the 4 folds. For the latter two evaluations, their hyperparameters are fit on training pipelines matching the upper descriptor, but then we report the modelling loss evaluated over training pipelines matching the lower descriptor. The per-agent lower bounds fit values directly to each agent's observed preferences rather than predicting them from training pipelines, so they are only applicable to the Full Fit evaluation.}
\label{tab:model_results}
\resizebox{\textwidth}{!}{%
\begin{tabular}{lccccc}
\toprule
\multirow{2}{*}{\textbf{Method}} & \textbf{Hyper-} & \multirow{2}{*}{\textbf{Full Fit}} & \textbf{K-Fold} & \textbf{Single-Stage Fit} & \textbf{No Distractors Fit} \\
                                 & \textbf{parameters} & & \textbf{Cross-Validation} & \textbf{Two-Stage Loss} & \textbf{Distractors Loss} \\
\midrule
\textbf{Proposed Model} & 57 & \textbf{0.1513} & \textbf{0.1532} $\pm$ 0.0031 & \textbf{0.1851} & \textbf{0.1535} \\
Simultaneous      & 57 & 0.1534 & 0.1554 $\pm$ 0.0028 & 0.2018 & 0.1577 \\
Quadratic         & 67 & 0.1605 & 0.1637 $\pm$ 0.0036 & 0.2009 & 0.1653 \\
Diagonal $S$      & 12 & 0.1765 & 0.1786 $\pm$ 0.0033 & 0.2069 & 0.1762 \\
Memoryless        & 57 & 0.2041 & 0.2073 $\pm$ 0.0059 & 0.2644 & 0.1900 \\
\midrule
Uniform Random (Baseline)   & 0 & 0.3679 & 0.3679 $\pm$ 0.0000 & 0.3776 & 0.3656 \\
\midrule
Per-Agent Per-Goal (Lower Bound)    & 7{,}152 & 0.0921 & N/A & N/A & N/A \\
Per-Agent Per-Feature (Lower Bound) & 2{,}980 & 0.1247 & N/A & N/A & N/A \\
\bottomrule
\end{tabular}%
}
\end{table}

In the main text, we report the KL divergence as our primary evaluation metric, since this is the objective used to fit the model's hyperparameters.
To provide additional confidence that the model's performance is not an artefact of optimising for a single metric, we report all four metrics defined above in \cref{tab:additional_metrics_3way,tab:additional_metrics_2way}.
These metrics are computed post-hoc on the full dataset using saved model parameters, and are not used during hyperparameter fitting.

\paragraph{Two-way vs three-way evaluation.}
\Cref{tab:additional_metrics_3way} reports metrics over the full three-way distribution (goal $a$, goal $b$, neither), which is the distribution our model is trained to predict.
\Cref{tab:additional_metrics_2way} reports metrics after renormalising to a two-way distribution (goal $a$ vs goal $b$), removing the null-goal component. This isolates how well the model captures \emph{relative} goal preferences, independent of its ability to predict the overall goal-reaching rate.

\begin{table}[h]
\centering
\caption{Post-hoc evaluation metrics over the \textbf{three-way} distribution (Full Fit). KL divergence is the training objective; other metrics are reported for validation.}
\label{tab:additional_metrics_3way}
\begin{tabular}{lcccc}
\toprule
\textbf{Method} & \textbf{KL} & \textbf{TV Distance} & \textbf{Brier Score} & \textbf{Dir. Accuracy} \\
\midrule
\textbf{Proposed Model} & \textbf{0.1536} & \textbf{0.2036} & \textbf{0.0993} & \textbf{87.71\%} \\
Simultaneous      & 0.1558 & 0.2059 & 0.1006 & 87.10\% \\
Quadratic         & 0.1627 & 0.2105 & 0.1054 & 85.61\% \\
Diagonal $S$      & 0.1789 & 0.2242 & 0.1164 & 84.16\% \\
Memoryless        & 0.2070 & 0.2373 & 0.1328 & 77.28\% \\
\midrule
Uniform Random    & 0.3679 & 0.3745 & 0.2459 & 0.00\% \\
\bottomrule
\end{tabular}
\end{table}

\begin{table}[h]
\centering
\caption{Post-hoc evaluation metrics over the \textbf{two-way} (renormalised) distribution (Full Fit). This isolates relative goal preference prediction, removing the null-goal component.}
\label{tab:additional_metrics_2way}
\begin{tabular}{lcccc}
\toprule
\textbf{Method} & \textbf{KL} & \textbf{TV Distance} & \textbf{Brier Score} & \textbf{Dir. Accuracy} \\
\midrule
\textbf{Proposed Model} & \textbf{0.0750} & \textbf{0.1323} & \textbf{0.0597} & \textbf{79.03\%} \\
Simultaneous      & 0.0779 & 0.1355 & 0.0622 & 78.58\% \\
Diagonal $S$      & 0.0823 & 0.1415 & 0.0684 & 75.26\% \\
Quadratic         & 0.0856 & 0.1412 & 0.0676 & 77.29\% \\
Memoryless        & 0.1057 & 0.1612 & 0.0895 & 71.69\% \\
\midrule
Uniform Random    & 0.1423 & 0.2107 & 0.1270 & 0.00\% \\
\bottomrule
\end{tabular}
\end{table}

The model ranking is consistent across all metrics and both evaluation variants, with the proposed model achieving the best performance on every metric.
The directional accuracy of 87.71\% (three-way) indicates that the model correctly predicts the preferred goal in nearly 9 out of 10 non-trivial matchups.

\FloatBarrier

\section{Proofs for Latent Policy Gradient Analysis}\label{app:latent_policy_gradient_proofs}

\subsection{Derivation of the Latent Gradient (Equation~\ref{eq:latent_gradient})}\label{app:gradient_derivation}

We derive the gradient of the modified policy objective with respect to the latent variables $\bm{w}$. 

\paragraph{Case 1: No distractor.} Recall the modified policy objective:
\begin{equation}
    J = \Pi\left( \bm{\phi}^{(g)} \middle | \bm{\phi}^{(g)}, \bm{0} \right) + \tau h\left( \Pi\left( \bm{\phi}^{(g)} \middle | \bm{\phi}^{(g)}, \bm{0} \right) \right)
\end{equation}
where $h(p) = -p \log p - (1-p)\log(1-p)$ is the binary entropy function. Under our linear Boltzmann-rational parametrisation, $\Pi = \sigma(v^g)$ where $v^g = \bm{\phi}^{(g)} \cdot S\bm{w}$.

We compute the gradient via the chain rule:
\begin{equation}
    \nabla_{\bm{w}} J = \frac{\partial J}{\partial \Pi} \cdot \frac{\partial \Pi}{\partial v^g} \cdot \frac{\partial v^g}{\partial \bm{w}}
\end{equation}

\textbf{Step 1:} The derivative of the entropy function is:
\begin{equation}
    \frac{dh}{dp} = -\log p - 1 + \log(1-p) + 1 = \log\left(\frac{1-p}{p}\right)
\end{equation}
Therefore:
\begin{equation}
    \frac{\partial J}{\partial \Pi} = 1 + \tau \log\left(\frac{1-\Pi}{\Pi}\right)
\end{equation}
Since $\Pi = \sigma(v^g) = \frac{e^{v^g}}{1 + e^{v^g}}$, we have $\frac{1-\Pi}{\Pi} = e^{-v^g}$, and thus:
\begin{equation}
    \frac{\partial J}{\partial \Pi} = 1 - \tau v^g
\end{equation}

\textbf{Step 2:} The derivative of the sigmoid function is:
\begin{equation}
    \frac{\partial \Pi}{\partial v^g} = \sigma(v^g)(1 - \sigma(v^g))
\end{equation}

\textbf{Step 3:} From $v^g = \bm{\phi}^{(g)} \cdot S\bm{w}$:
\begin{equation}
    \frac{\partial v^g}{\partial \bm{w}} = S^T \bm{\phi}^{(g)}
\end{equation}

Combining these three steps:
\begin{equation}
    \nabla_{\bm{w}} J = (1 - \tau v^g) \sigma(v^g)(1 - \sigma(v^g)) S^T \bm{\phi}^{(g)}
\end{equation}
as claimed. \hfill $\square$

\paragraph{Case 2: With distractor.} When a distractor object with features $\bm{\phi}^{(d)}$ is present, the agent chooses between three options: the goal, the distractor, and doing nothing. The natural extension of our Boltzmann-rational parametrisation is:
\begin{equation}
    \Pi\left( \bm{\phi}^{(g)} \middle | \bm{\phi}^{(g)}, \bm{\phi}^{(d)}, \bm{0} \right) = \frac{\exp(v^g)}{\exp(v^g) + \exp(v^d) + 1}
\end{equation}
where $v^g = \bm{\phi}^{(g)} \cdot S\bm{w}$ and $v^d = \bm{\phi}^{(d)} \cdot S\bm{w}$ are the values assigned to the goal and distractor respectively.

For notational convenience, let $\Pi^g = \Pi\left( \bm{\phi}^{(g)} \middle | \bm{\phi}^{(g)}, \bm{\phi}^{(d)}, \bm{0} \right)$ denote the probability of choosing the goal. The modified policy objective becomes:
\begin{equation}
    J = \Pi^g + \tau h(\Pi^g)
\end{equation}

As before, we have:
\begin{equation}
    \frac{\partial J}{\partial \Pi^g} = 1 + \tau \log\left( \frac{1 - \Pi^g}{\Pi^g} \right)
\end{equation}

To compute $\frac{\partial \Pi^g}{\partial \bm{w}}$, let $Z = \exp(v^g) + \exp(v^d) + 1$ be the partition function. Then:
\begin{align}
    \frac{\partial \Pi^g}{\partial v^g} &= \frac{\exp(v^g) \cdot Z - \exp(v^g) \cdot \exp(v^g)}{Z^2} = \Pi^g (1 - \Pi^g) \\
    \frac{\partial \Pi^g}{\partial v^d} &= \frac{0 - \exp(v^g) \cdot \exp(v^d)}{Z^2} = -\Pi^g \Pi^d
\end{align}
where $\Pi^d = \frac{\exp(v^d)}{Z}$ is the probability of choosing the distractor.

Therefore:
\begin{align}
    \frac{\partial \Pi^g}{\partial \bm{w}} &= \frac{\partial \Pi^g}{\partial v^g} \frac{\partial v^g}{\partial \bm{w}} + \frac{\partial \Pi^g}{\partial v^d} \frac{\partial v^d}{\partial \bm{w}} \\
    &= \Pi^g(1 - \Pi^g) S^T \bm{\phi}^{(g)} - \Pi^g \Pi^d S^T \bm{\phi}^{(d)} \\
    &= \Pi^g \left[ (1 - \Pi^g) S^T \bm{\phi}^{(g)} - \Pi^d S^T \bm{\phi}^{(d)} \right]
\end{align}

Combining with $\frac{\partial J}{\partial \Pi^g}$, the full gradient is:
\begin{equation}\label{eq:distractor_gradient}
    \nabla_{\bm{w}} J = \left( 1 + \tau \log\left( \frac{1 - \Pi^g}{\Pi^g} \right) \right) \Pi^g \left[ (1 - \Pi^g) S^T \bm{\phi}^{(g)} - \Pi^d S^T \bm{\phi}^{(d)} \right]
\end{equation}

Note that in the distractor case, the gradient is no longer purely in the direction $S^T \bm{\phi}^{(g)}$, but contains an additional component in the direction $S^T \bm{\phi}^{(d)}$. This reflects the fact that training with a distractor present affects how the agent values features associated with the distractor. \hfill $\square$

\subsection{Analytic Solution for Equilibrium Weights (Equation~\ref{eq:analytic_solution})}\label{app:equilibrium_derivation}

We derive the closed-form expression for the weights after convergence to equilibrium in the no-distractor case.

From the gradient expression derived above, the updates to $\bm{w}$ are always in the direction $S^T \bm{\phi}^{(g)}$. Therefore, starting from initial weights $\bm{w}_0$, the final weights must take the form:
\begin{equation}
    \bm{w} = \bm{w}_0 + \alpha S^T \bm{\phi}^{(g)}
\end{equation}
for some scalar $\alpha \in \mathbb{R}$.

At equilibrium, setting $\nabla_{\bm{w}} J = 0$ and noting that $\sigma(v^g)(1-\sigma(v^g)) > 0$ for finite $v^g$, we require:
\begin{equation}
    1 - \tau v^g = 0 \implies v^g = \tau^{-1}
\end{equation}

Substituting the equilibrium condition $v^g = \bm{\phi}^{(g)} \cdot S\bm{w} = \tau^{-1}$:
\begin{equation}
    \bm{\phi}^{(g)} \cdot S\left(\bm{w}_0 + \alpha S^T \bm{\phi}^{(g)}\right) = \tau^{-1}
\end{equation}

Expanding:
\begin{equation}
    \bm{\phi}^{(g)} \cdot S\bm{w}_0 + \alpha \bm{\phi}^{(g)} \cdot SS^T \bm{\phi}^{(g)} = \tau^{-1}
\end{equation}

Noting that $\bm{\phi}^{(g)} \cdot SS^T \bm{\phi}^{(g)} = (S^T\bm{\phi}^{(g)})^T(S^T\bm{\phi}^{(g)}) = \|S^T \bm{\phi}^{(g)}\|_2^2$, we solve for $\alpha$:
\begin{equation}
    \alpha = \frac{\tau^{-1} - \bm{\phi}^{(g)} \cdot S\bm{w}_0}{\|S^T \bm{\phi}^{(g)}\|_2^2}
\end{equation}

Substituting back:
\begin{equation}
    \bm{w} = \bm{w}_0 + \left(\frac{\tau^{-1} - \bm{\phi}^{(g)} \cdot S\bm{w}_0}{\|S^T \bm{\phi}^{(g)}\|_2^2}\right) S^T \bm{\phi}^{(g)}
\end{equation}
as claimed. \hfill $\square$

\FloatBarrier
\section{Supplementary Figures}\label{app:supplementary_figures}

\FloatBarrier
\subsection{Detailed Empirical Analysis}\label{app:empirical_details}

This section provides the full figures and detailed analysis supporting the empirical findings summarised in \cref{sec:empirical_results}.

\begin{figure}[h]
    \centering
    \begin{subfigure}[c]{0.48\linewidth}
        \centering
        \includegraphics[width=\linewidth]{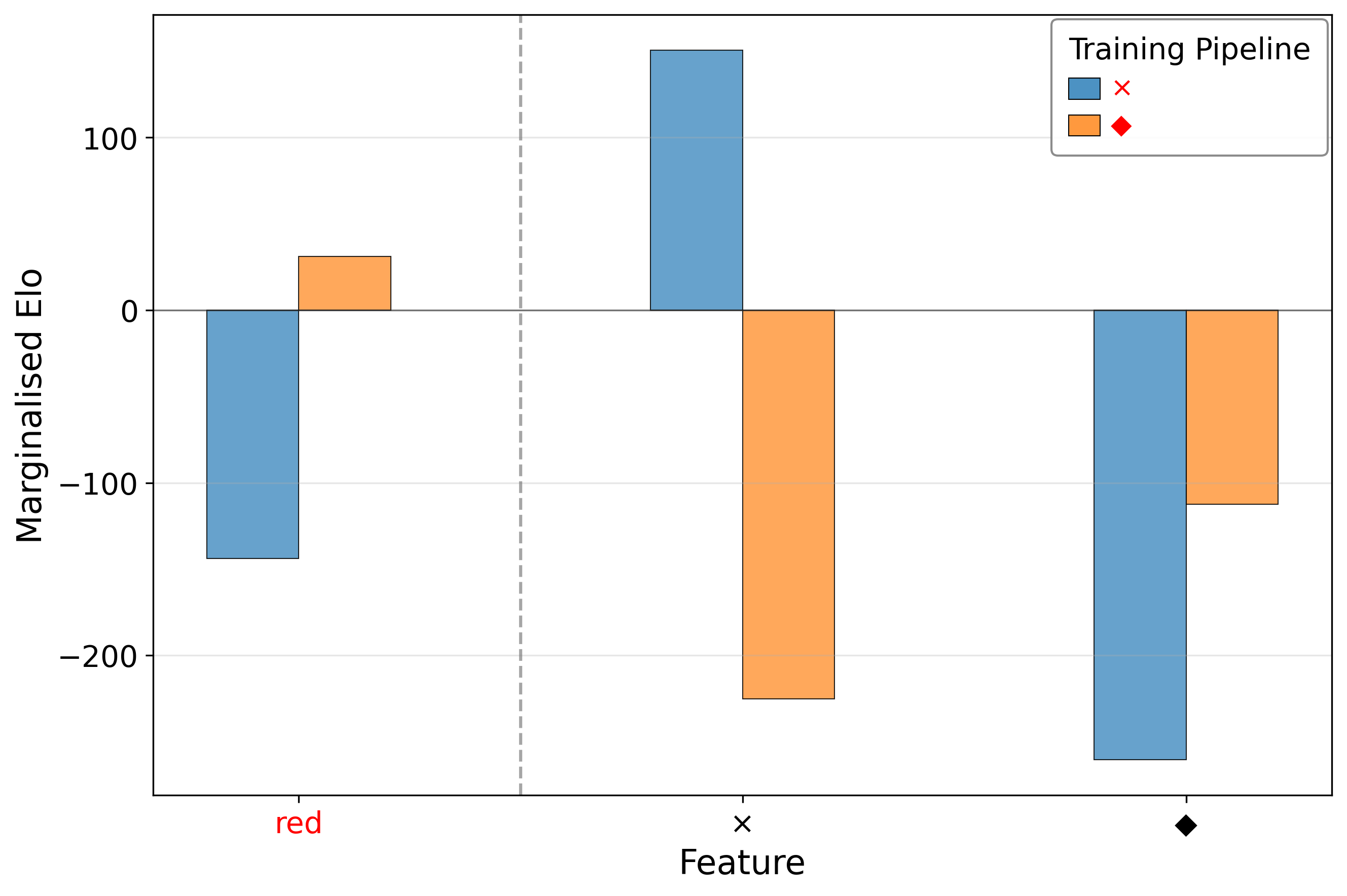}
        \label{fig:red_cross_vs_red_diamond}
    \end{subfigure}
    \hfill
    \begin{subfigure}[c]{0.48\linewidth}
        \centering
        \includegraphics[width=\linewidth]{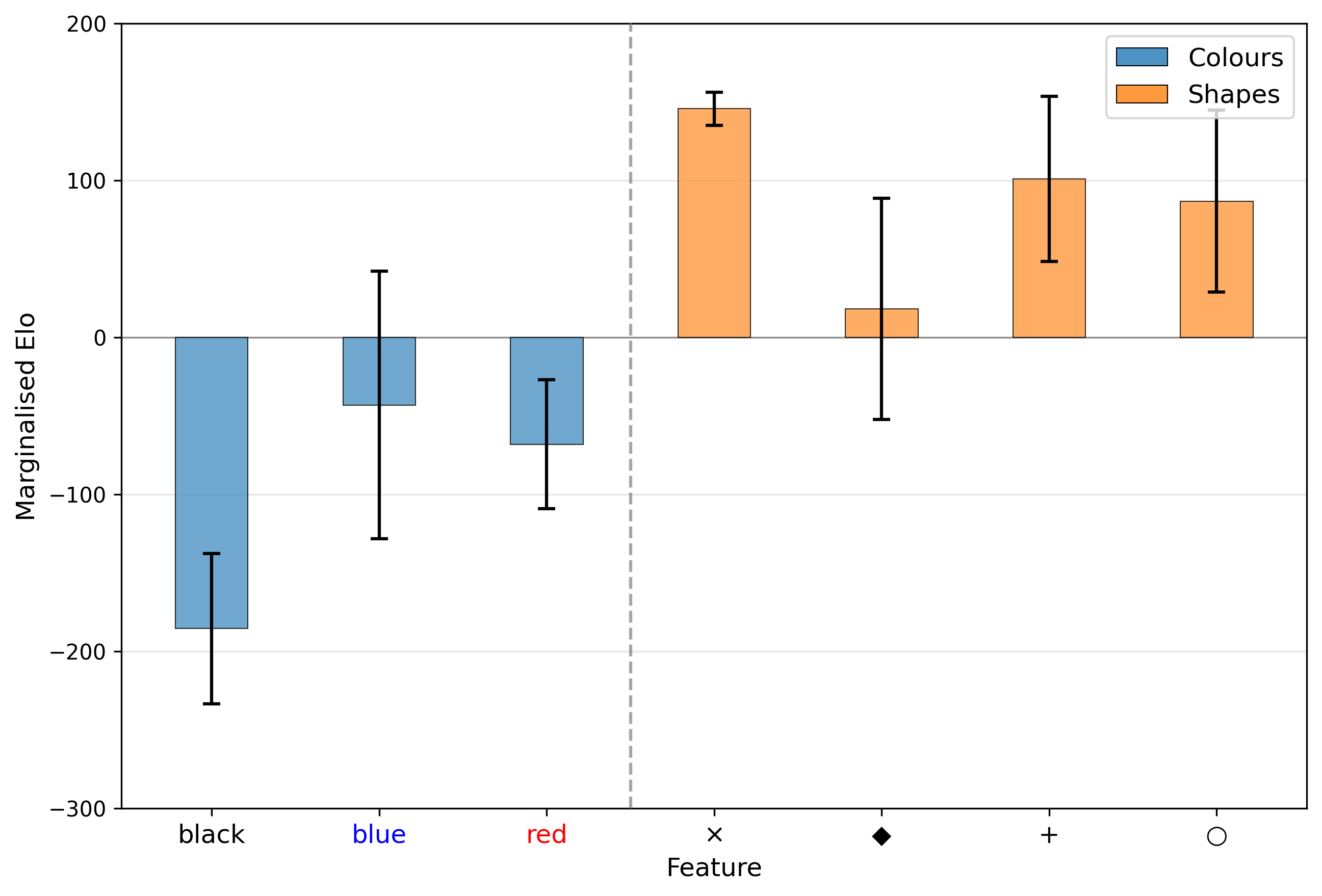}
        \label{fig:saliencies}
    \end{subfigure}
    \caption{\textbf{Left: Generalisation differences between \sym{red}{cross} and \sym{red}{diamond} training goals.} Marginalised feature Elo scores for the \red{}, \Cross-shape, and \Diam{}-shape features for the single-stage agent trained on \sym{red}{cross}, and the single-stage agent trained on \sym{red}{diamond}.
    \textbf{Right: Some features drive generalisation more strongly and are more salient to the model.} Average marginalised Elo with standard error for each feature across agents that have been trained on a goal containing that feature. The average is taken over all single-stage runs without distractors.}
    \label{fig:feature_saliency_combined}
\end{figure}

\paragraph{Feature saliency.}
\Cref{fig:red_cross_vs_red_diamond} illustrates how different training goals lead to different generalisation patterns.
The agent trained on \sym{red}{cross} strongly prefers \Cross{}-shaped objects over \red{}-coloured objects, whereas the agent trained on \sym{red}{diamond} strongly prefers \red{}-coloured objects and only weakly prefers \Diam{}-shaped objects.
\Cref{fig:saliencies} quantifies this effect across all single-stage agents: for each feature, we average the marginalised Elo over all agents trained on goals containing that feature.
Shape tends to drive generalisation more strongly than colour, with \Cross{}-shape being the most salient and \Diam{}-shape the least.
Black---the colour of the maze walls---has the lowest average marginalised Elo, as agents trained on black goals generalise over shape rather than colour.

\begin{figure}[h]
    \centering
    \includegraphics[width=0.55\textwidth]{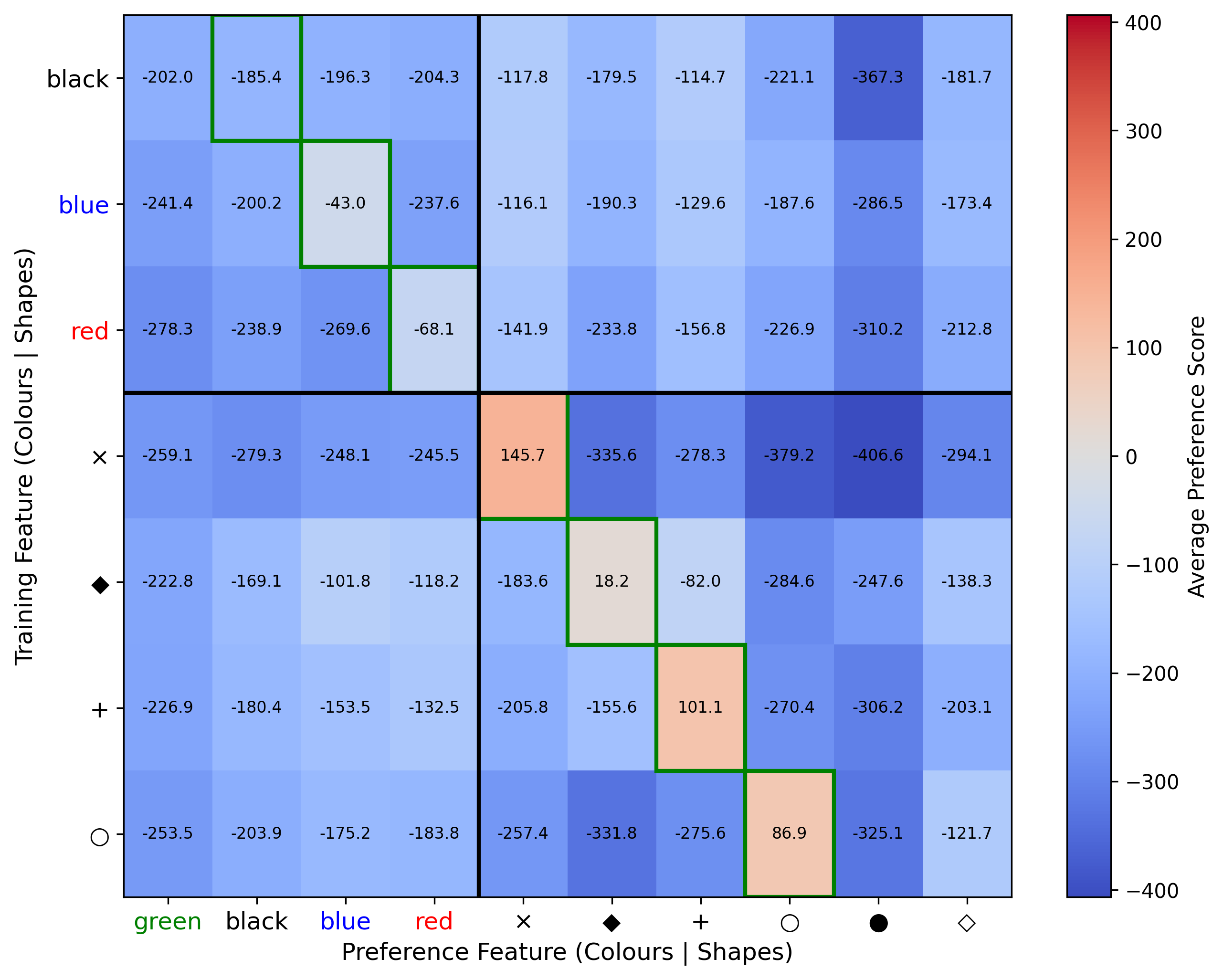}
    \caption{\textbf{Training on one feature can lead to another being valued more or less strongly than average.} Rows indicate a feature being trained on and columns indicate a feature being evaluated, the value being the average preference score for goals containing the evaluation feature across models trained on goals containing the training feature. The average is taken over all single-stage runs without distractors.}
    \label{fig:cross_saliencies}
\end{figure}

\paragraph{Feature cross-saliency and confusion.}\label{app:feature_cross_saliency}
We also computed the marginalised Elo for each pair of features, measuring the marginalised Elo of single-stage agents for one feature if they were trained on a goal containing the other.
\Cref{fig:cross_saliencies} shows this cross-saliency matrix.
We observe \emph{feature confusion}, where training on one feature can lead to valuing or de-valuing another---a phenomenon captured by the off-diagonal entries of the saliency matrix $S$ in our latent policy gradients model (\cref{sec:modelling_generalisation}).

\begin{figure}[h]
    \centering
    \begin{subfigure}[c]{0.48\linewidth}
        \centering
        \includegraphics[width=\linewidth]{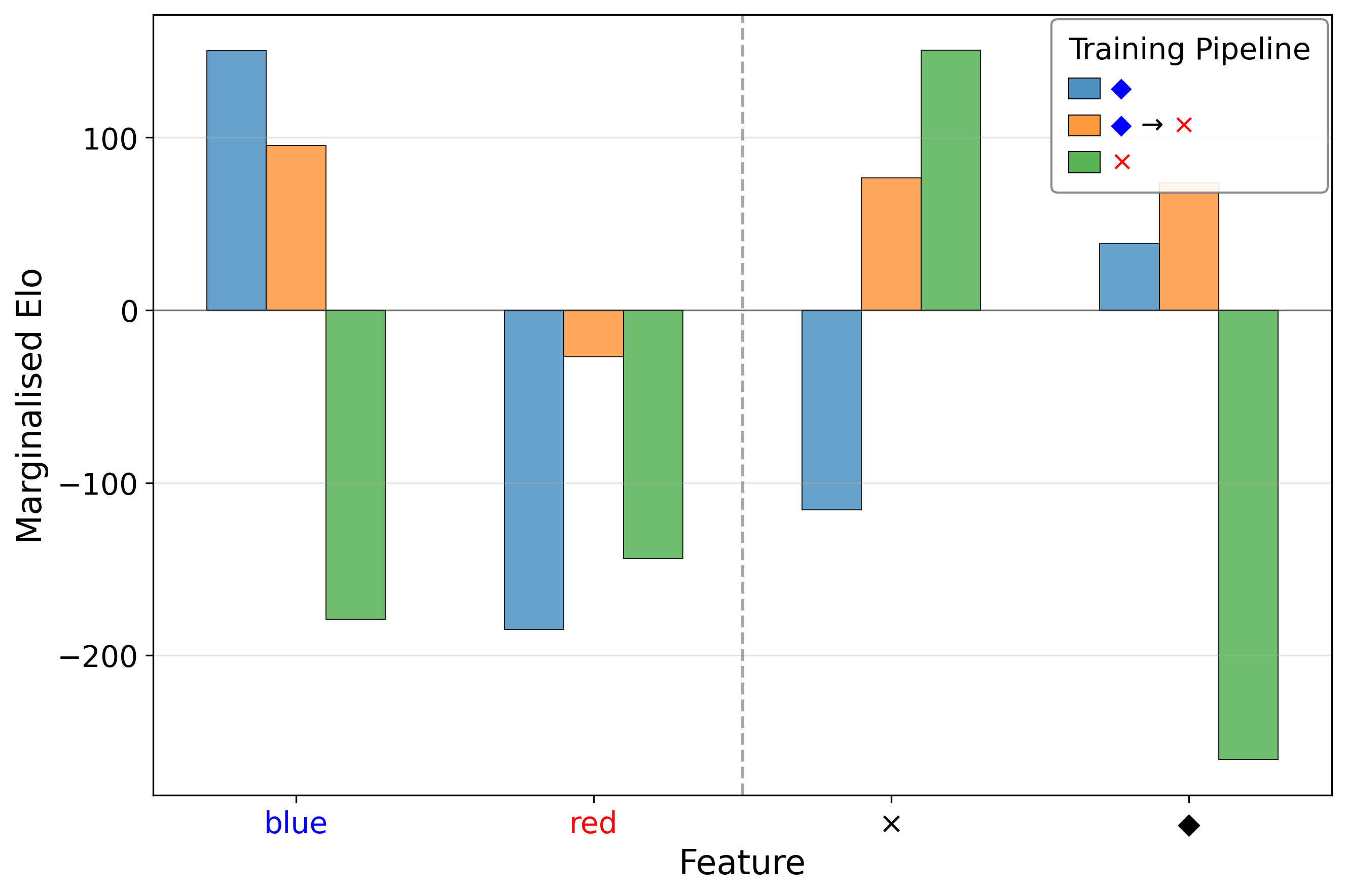}
        \label{fig:blue_diamond_vs_blue_diamond_then_red_cross_vs_red_cross}
    \end{subfigure}
    \hfill
    \begin{subfigure}[c]{0.48\linewidth}
        \centering
        \includegraphics[width=\linewidth]{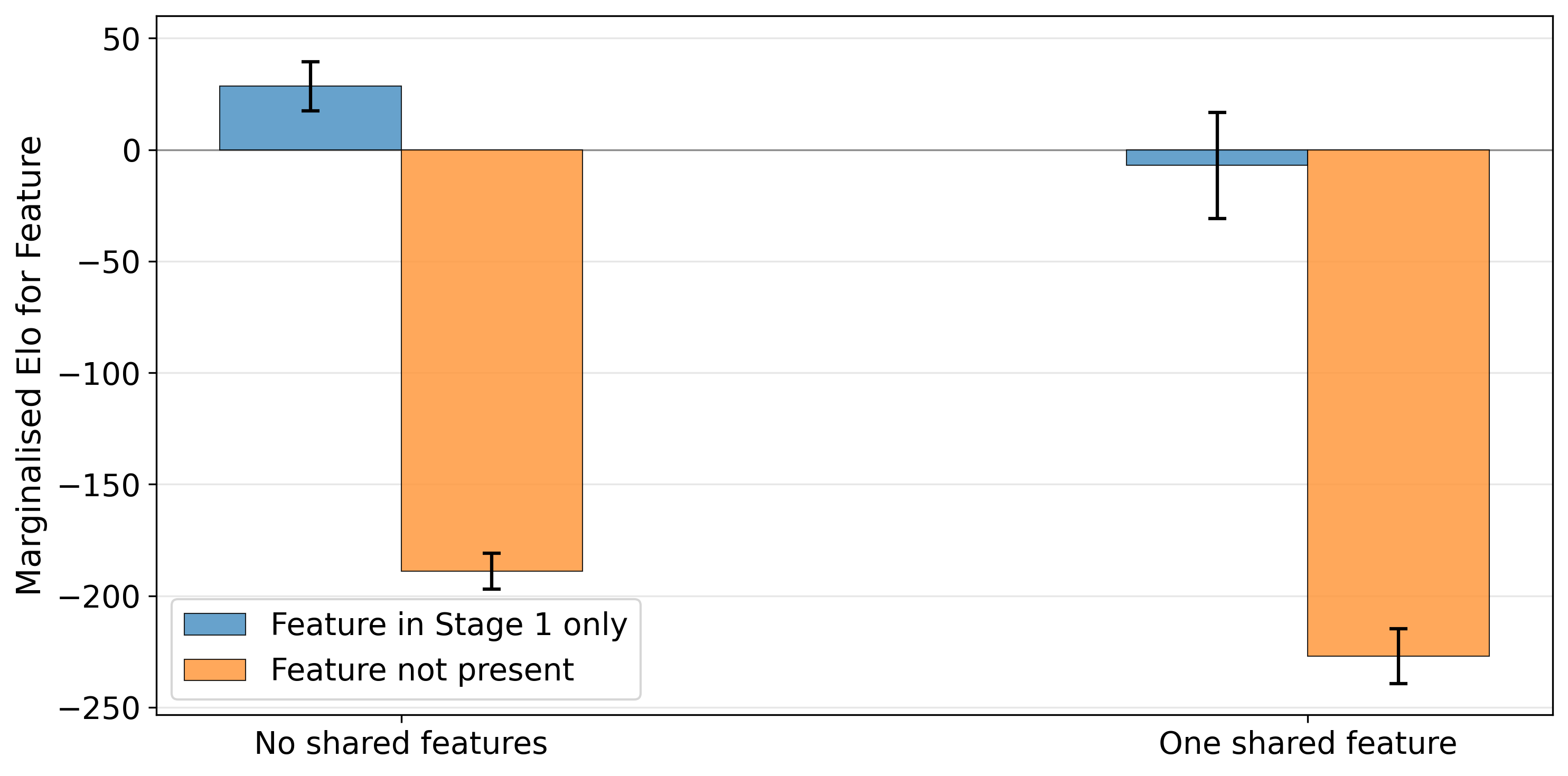}
        \label{fig:value_persistence}
    \end{subfigure}
    \caption{\textbf{Left: \Diam{}-shape and \blue{} values persist after training on \sym{red}{cross}.} Marginalised feature Elo scores for \red{}, \blue{}, \Diam{}-shape, and \Cross{}-shape features for three different agents: single-stage agent trained on \sym{blue}{diamond}; a two-stage agent, \sym{blue}{diamond}$\to$\sym{red}{cross}; a single-stage agent trained just on \sym{red}{cross}.
    \textbf{Right: Values for early training objectives persist.} Marginalised Elo with standard error for features that are only in the first goal in two-stage training pipelines compared to pipelines where they are not present at any stage.
    We measure this by averaging the marginalised Elo for a given feature across agents with that feature only in their first goal, and comparing this to agents with the same training pipeline aside from the feature of interest, which is substituted for something else.
    \emph{E.g.}, the \blue{} value of \sym{blue}{cross}$\to$\sym{red}{diamond} is compared with that of \sym{black}{cross}$\to$\sym{red}{diamond} and \sym{red}{cross}$\to$\sym{red}{diamond} to determine how much a \blue{}-coloured first goal causes agents to value \blue{}-coloured objects.
    Here we only consider two-stage pipelines without distractors.
    On the left of each pair we show this restricted to pipelines whose goals don't share any features, and on the right we restrict it to pipelines where the stage's goals share some other feature than the one being measured.
    Value persistence is shown by the large differences between the bars within each pair---when a feature is only present in the first stage, the downstream value for that feature is much higher than would be expected if that feature was not present.}
    \label{fig:value_persistence_combined}
\end{figure}

\paragraph{Value persistence.}
\Cref{fig:blue_diamond_vs_blue_diamond_then_red_cross_vs_red_cross} shows a case study: an agent first trained on \sym{blue}{diamond} and then \sym{red}{cross} (\sym{blue}{diamond}$\to$\sym{red}{cross}) retains strong preferences for \blue{}-coloured and \Diam{}-shaped objects, in addition to valuing \Cross{}-shaped objects, compared to an agent trained only on \sym{red}{cross}.
\Cref{fig:value_persistence} confirms this pattern across all two-stage pipelines: features present only in the first stage's goal have much higher downstream Elo than controls where that feature is substituted.
This holds both when the two goals share a feature and when they do not.

\begin{figure}[h]
    \centering
    \includegraphics[width=0.6\linewidth]{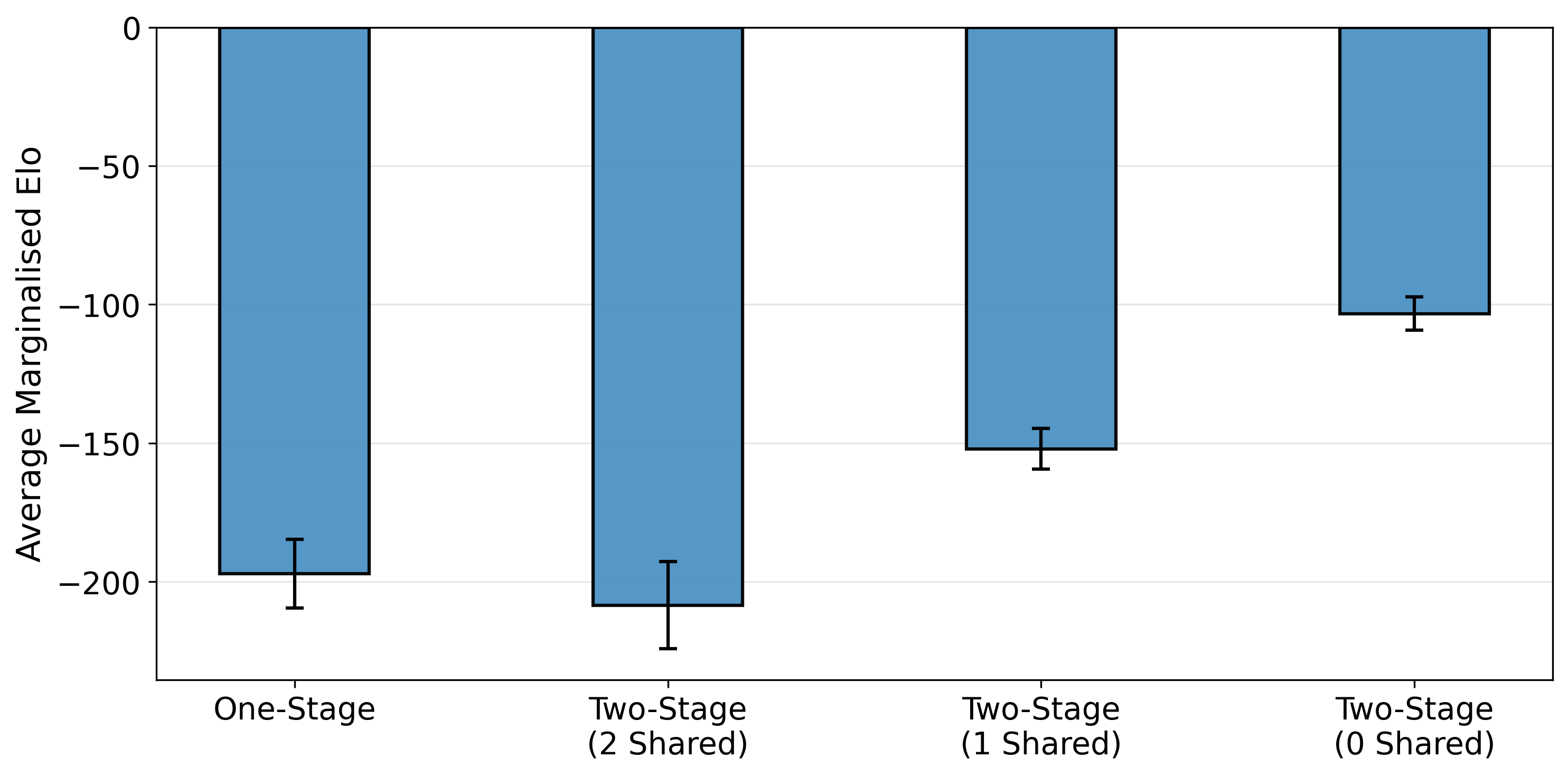}
    \caption{\textbf{Agents trained on more diverse feature sets pursue more goals.} Marginalised Elo with standard error across all features for agents that have been trained on a different number of total unique features. Single-stage pipelines always have two unique features in their goals. Two-stage pipelines can have goals which share both, one, or neither of their features.}
    \label{fig:feature_diversity}
\end{figure}

\paragraph{Feature diversity and goal pursuit.}\label{app:supplementary_figures_feature_diversity}
Agents trained on more unique features generalise to pursue more objects.
Whilst agents typically don't pursue most objects (corresponding to negative Elos on average), this effect reduces with longer and more feature-diverse training pipelines (\cref{fig:feature_diversity}).

\begin{figure}[h]
    \centering
    \begin{subfigure}[c]{0.48\linewidth}
        \centering
        \includegraphics[width=\linewidth]{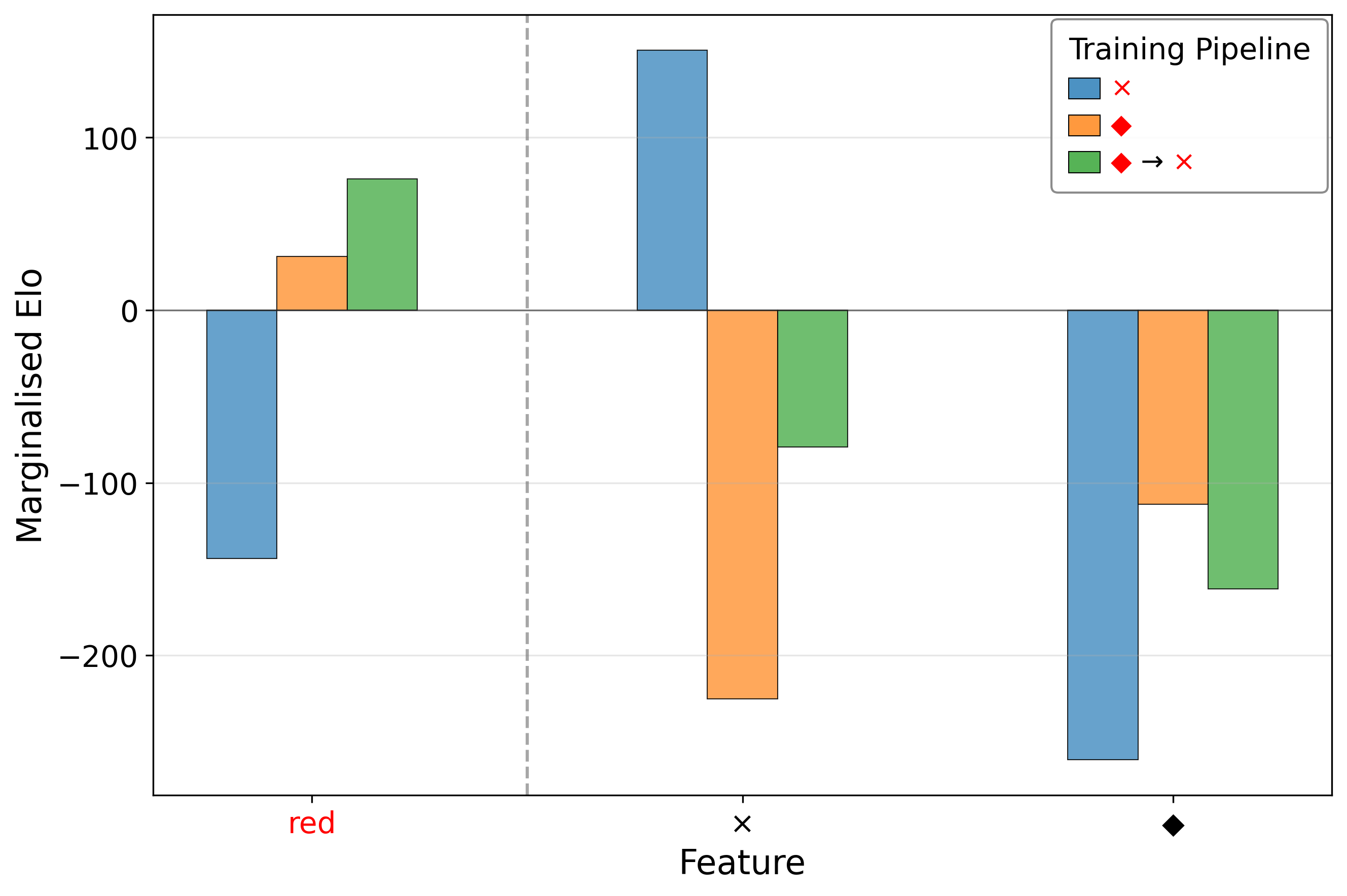}
        \label{fig:red_cross_vs_red_diamond_vs_red_diamond_then_red_cross}
    \end{subfigure}
    \hfill
    \begin{subfigure}[c]{0.48\linewidth}
        \centering
        \includegraphics[width=\linewidth]{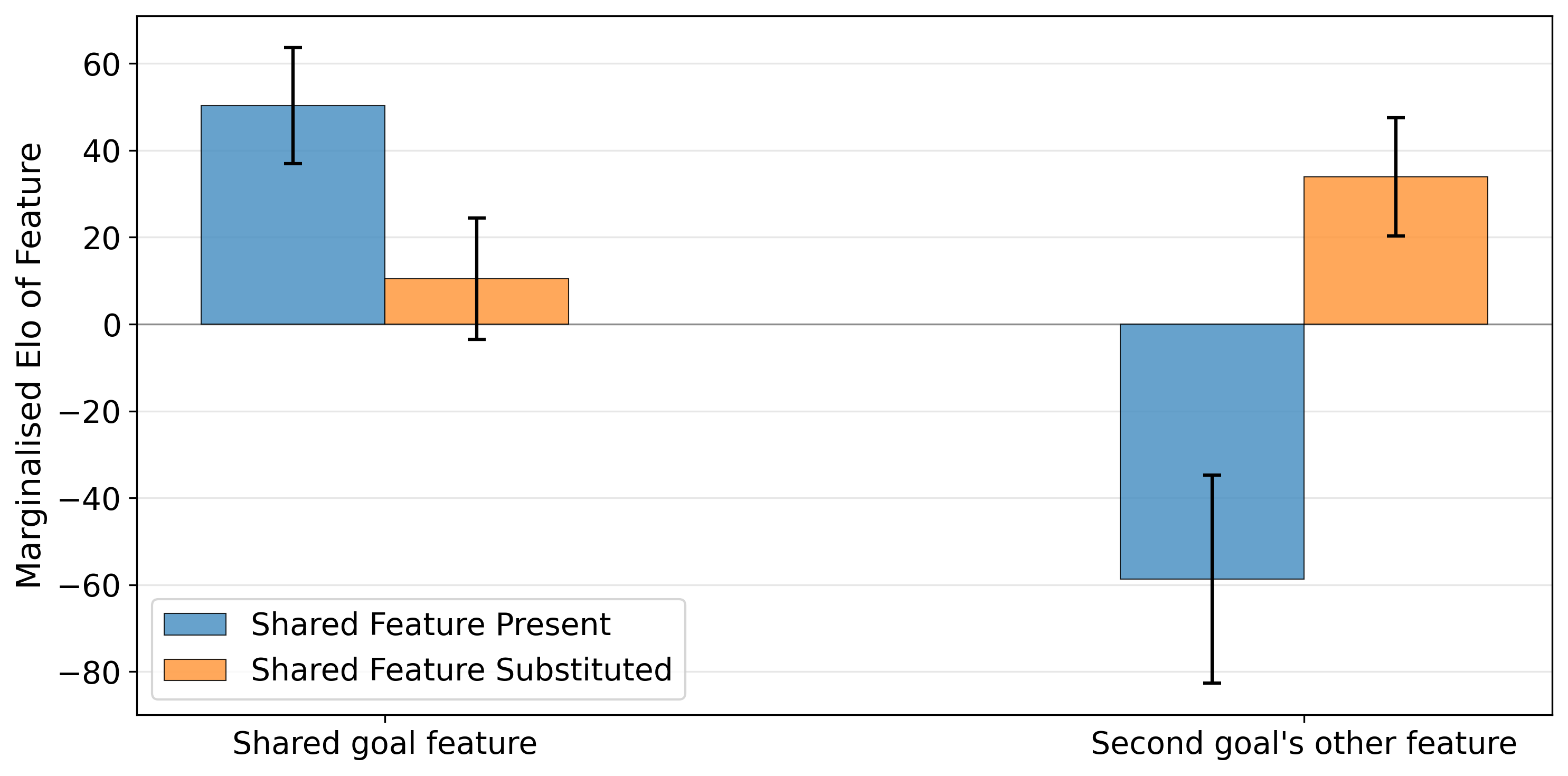}
        \label{fig:value_inhibition}
    \end{subfigure}
    \caption{\textbf{Left: Training on \sym{red}{diamond}$\to$\sym{red}{cross} does not cause a strong value for \Cross{}-shape.} Marginalised feature Elo scores for \red{}, \Cross{}-shape, and \Diam{}-shape for three different agents: a single-stage agent trained just on a \sym{red}{cross}; a single-stage agent trained on a \sym{red}{diamond}; a two-stage agent trained on \sym{red}{diamond}$\to$\sym{red}{cross}.
    \textbf{Right: Repeated goal features' values are strengthened, and inhibit new values forming.}
    The left bars within each pair show the marginalised Elo with standard error for the two features of the last goal in two-stage training pipelines where a feature is shared between the two goals.
    The left group of bars plots the marginalised Elo of the shared feature itself, whereas the right group of bars plots the marginalised Elo of the other feature in the second stage goal.
    The right bars within each pair show the marginalised Elo for the same features but averaged over training pipelines where they are no longer present in the first stage goal (controlling for the other features in the training pipeline).
    Here we only consider two-stage pipelines without distractors.
    In the left pair we see that when a feature is shared between two training stages, its value is strengthened, and much higher than if it were only present in the second stage.
    In the right pair we see that when a feature is shared between two training stages, the other non-shared feature in the second stage goal is valued much less than if the first stage goal did not share any features with the second stage goal.}
    \label{fig:value_inhibition_combined}
\end{figure}

\paragraph{Value strengthening and inhibition.}
\Cref{fig:red_cross_vs_red_diamond_vs_red_diamond_then_red_cross} shows that subsequently training a \sym{red}{diamond} agent on \sym{red}{cross} does \textbf{not} cause it to value \Cross{}-shape nearly as much as a single-stage \sym{red}{cross} agent does. Instead, its value for \red{} is strengthened, and the \Cross{}-shape value is much weaker than it would have been without prior training.
\Cref{fig:value_inhibition} confirms this across all two-stage pipelines: shared features are strengthened (left pair), while the non-shared feature in the second goal is inhibited (right pair).

\begin{figure}[h]
    \centering
    \begin{subfigure}{0.49\linewidth}
        \includegraphics[width=\linewidth]{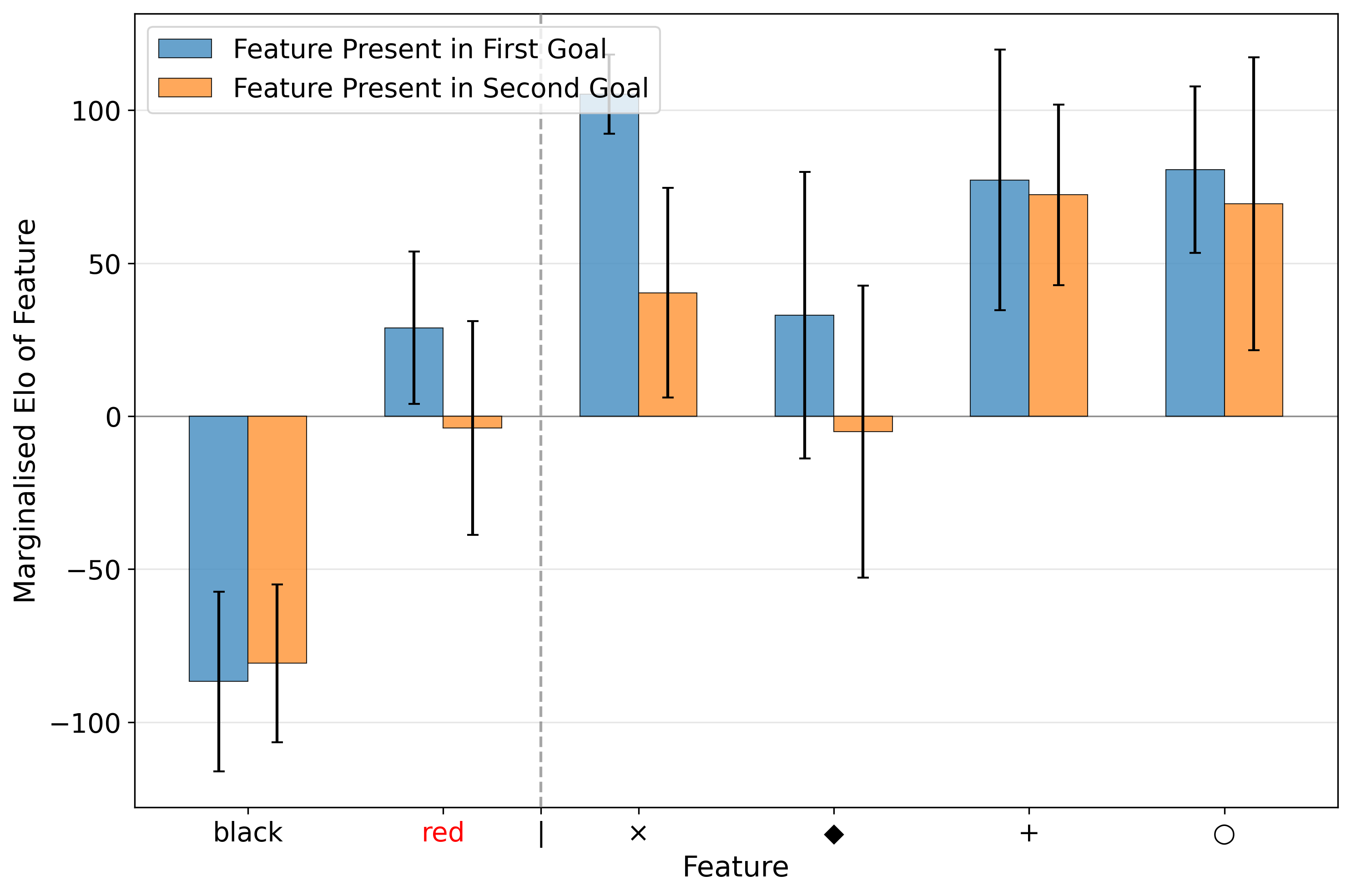}
    \end{subfigure}
    \begin{subfigure}{0.49\linewidth}
        \includegraphics[width=\linewidth]{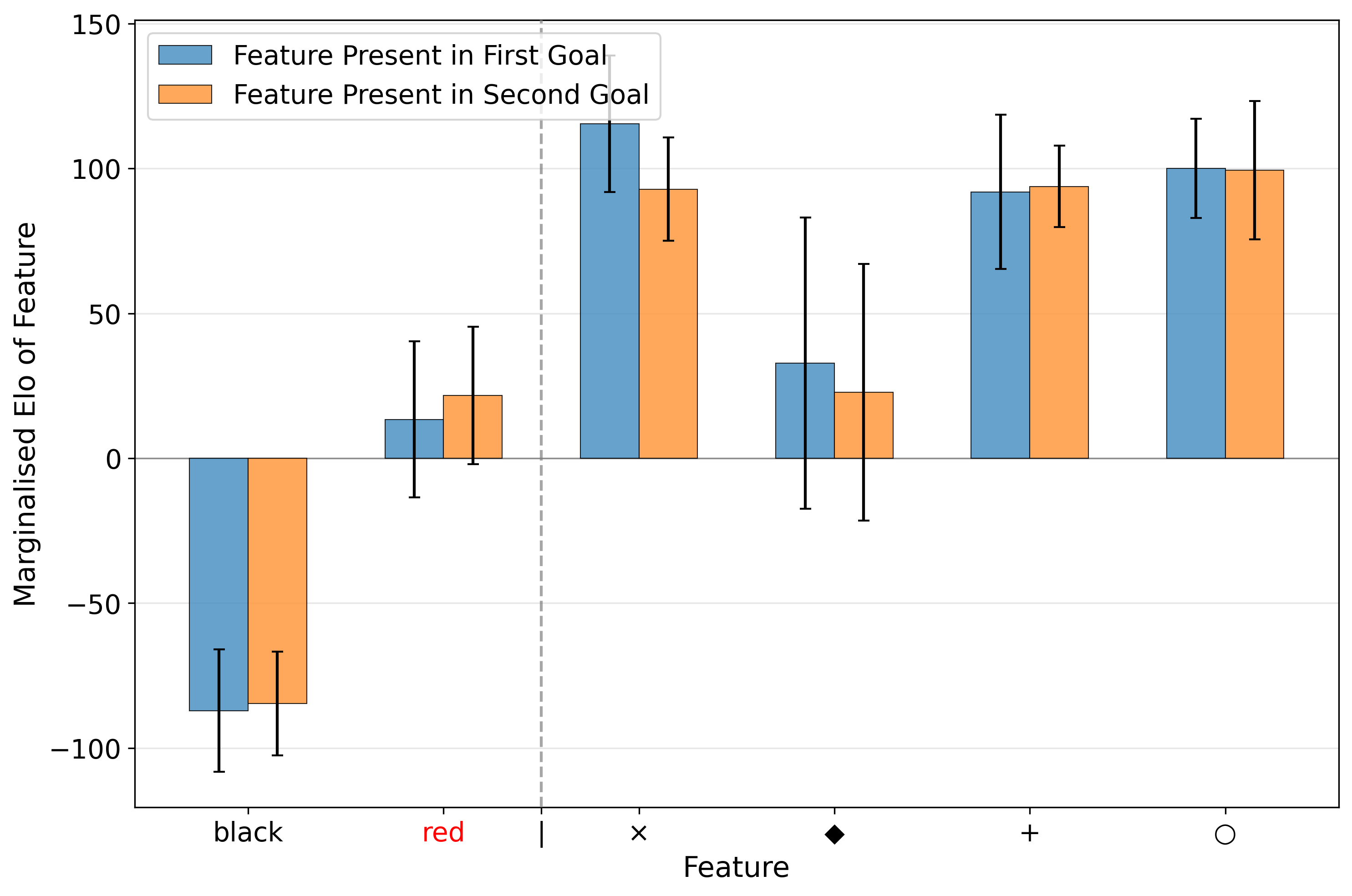}
    \end{subfigure}
    \caption{\textbf{Generalisation behaviour is sensitive to the order of the training objectives when they share a feature.} Marginalised Elo with standard error for each feature stratified by when that feature is present in the first training goal compared to when it is present in the second training goal. Elo is marginalised over pairs of two-stage pipelines without distractors where they are each others reverse.
    \textbf{Left:} Cases where there is a shared feature between the two goals (\emph{e.g.}, the pair \sym{red}{diamond}$\to$\sym{black}{diamond} and \sym{black}{diamond}$\to$\sym{red}{diamond}).
    \textbf{Right:} Cases where there are no shared features between the two goals (\emph{e.g.}, the pair \sym{red}{diamond}$\to$\sym{black}{cross} and \sym{black}{cross}$\to$\sym{red}{diamond}).}
    \label{fig:ordering_effects}
\end{figure}

\paragraph{Ordering effects.}\label{app:supplementary_figures_ordering_effects}
A corollary of value strengthening and inhibition is that when goals share a feature, flipping the order of training stages can significantly change the agent's final values.
\Cref{fig:ordering_effects} shows this effect: when there is a shared feature, the \Cross{}-shape feature is often much less likely to be valued when it is present in the second goal vs.\ the first.
We do not observe strong ordering effects when there is no shared feature between the two goals.

\FloatBarrier
\subsection{Agent Training Curves}\label{app:supplementary_figures_agent_training}

\begin{figure}[h]
      \centering
      \includegraphics[width=0.6\linewidth]{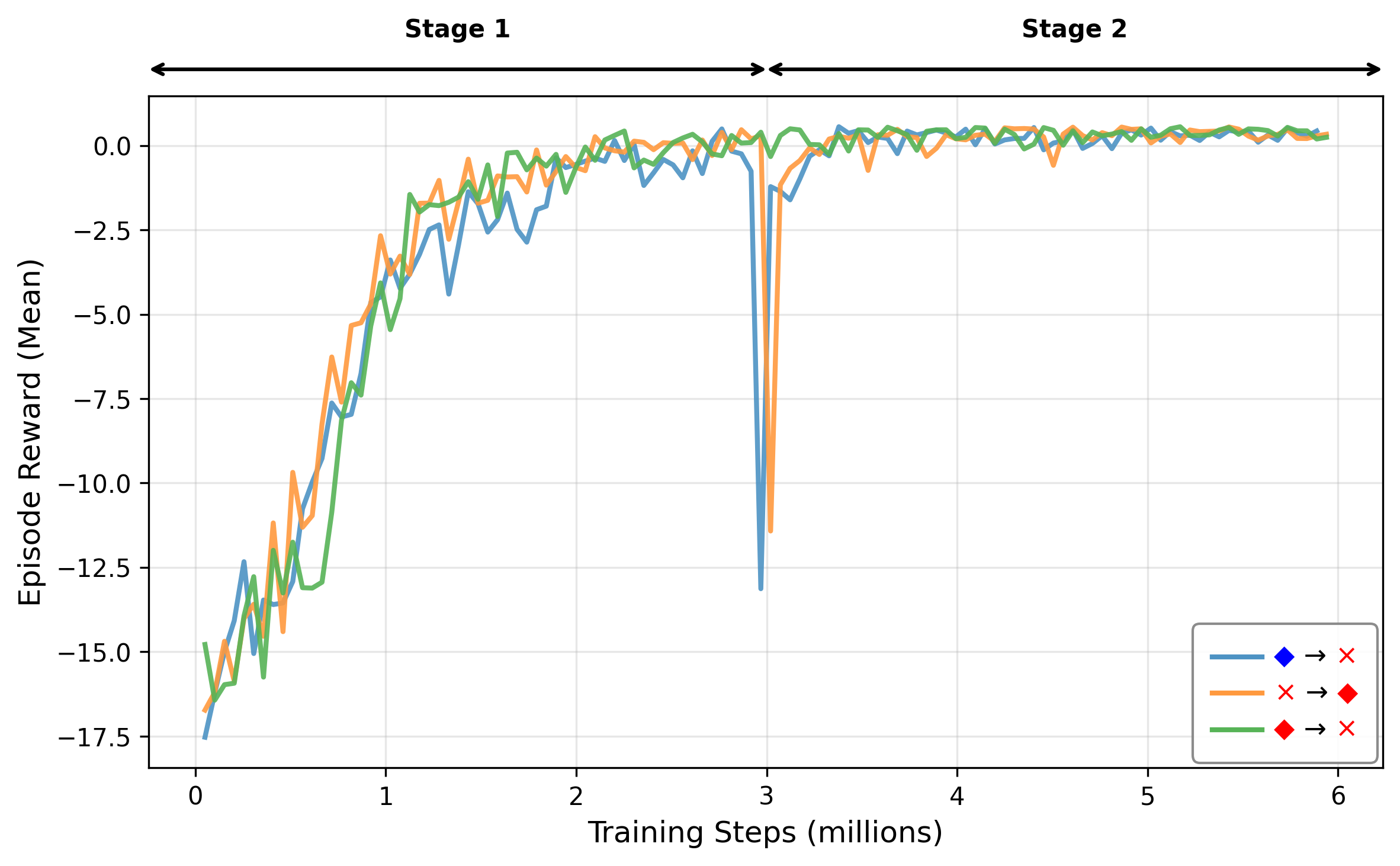}
      \caption{Training curves showing mean episode reward over training steps for selected agents. Two-stage pipelines often exhibit a performance drop at the stage transition when the goal changes, followed by rapid recovery. Interestingly, for some training pipelines this doesn't happen, as the new goal contains a feature that had been generalised to in the first training stage (\emph{e.g.}, \sym{red}{diamond}$\to$\sym{red}{cross}).}
      \label{fig:training_curves}
\end{figure}

\FloatBarrier
\subsection{Agent Colour and Shape Preferences}\label{app:supplementary_figures_agent_prefs}

\begin{figure}[h]
    \centering
    \includegraphics[width=0.6\linewidth]{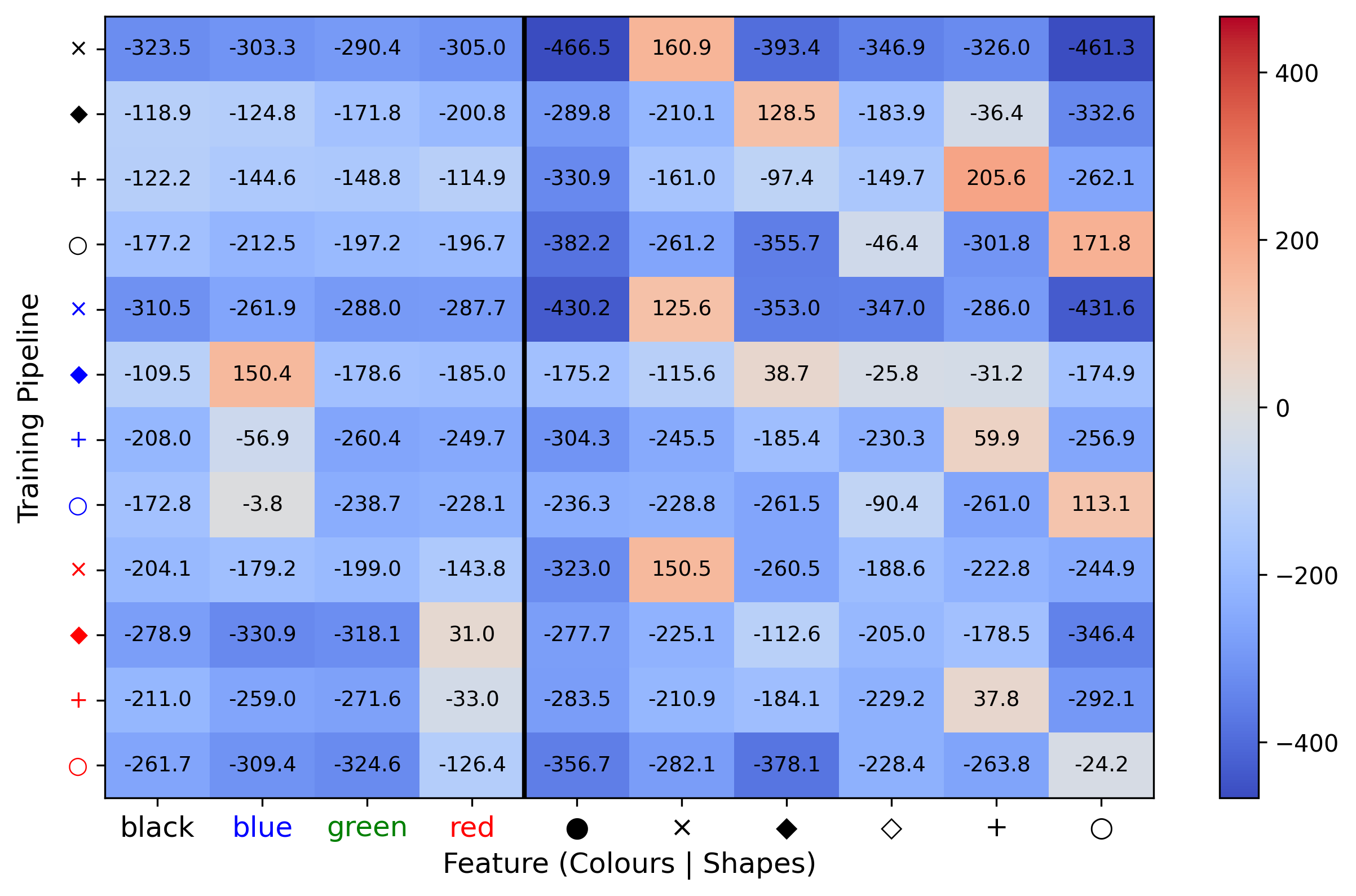}
    \caption{Agent values for single-stage training pipelines without distractors.}
    \label{fig:single_stage_summary}
\end{figure}

\begin{figure}[h]
    \centering
    \begin{subfigure}{0.49\linewidth}
        \includegraphics[width=\linewidth]{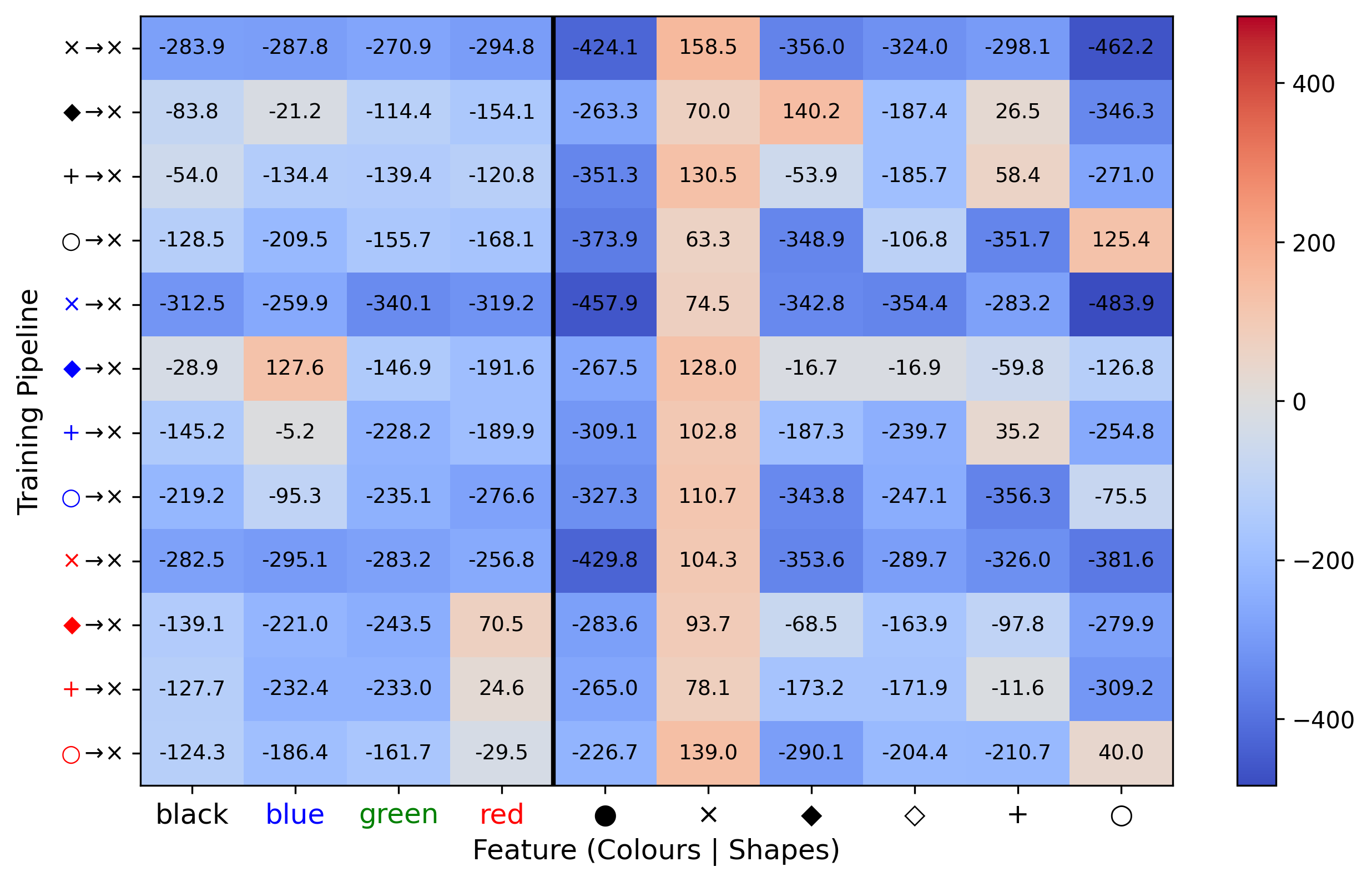}
        \caption{Black cross}
        \label{fig:finetune_black_cross}
    \end{subfigure}
    \begin{subfigure}{0.49\linewidth}
        \includegraphics[width=\linewidth]{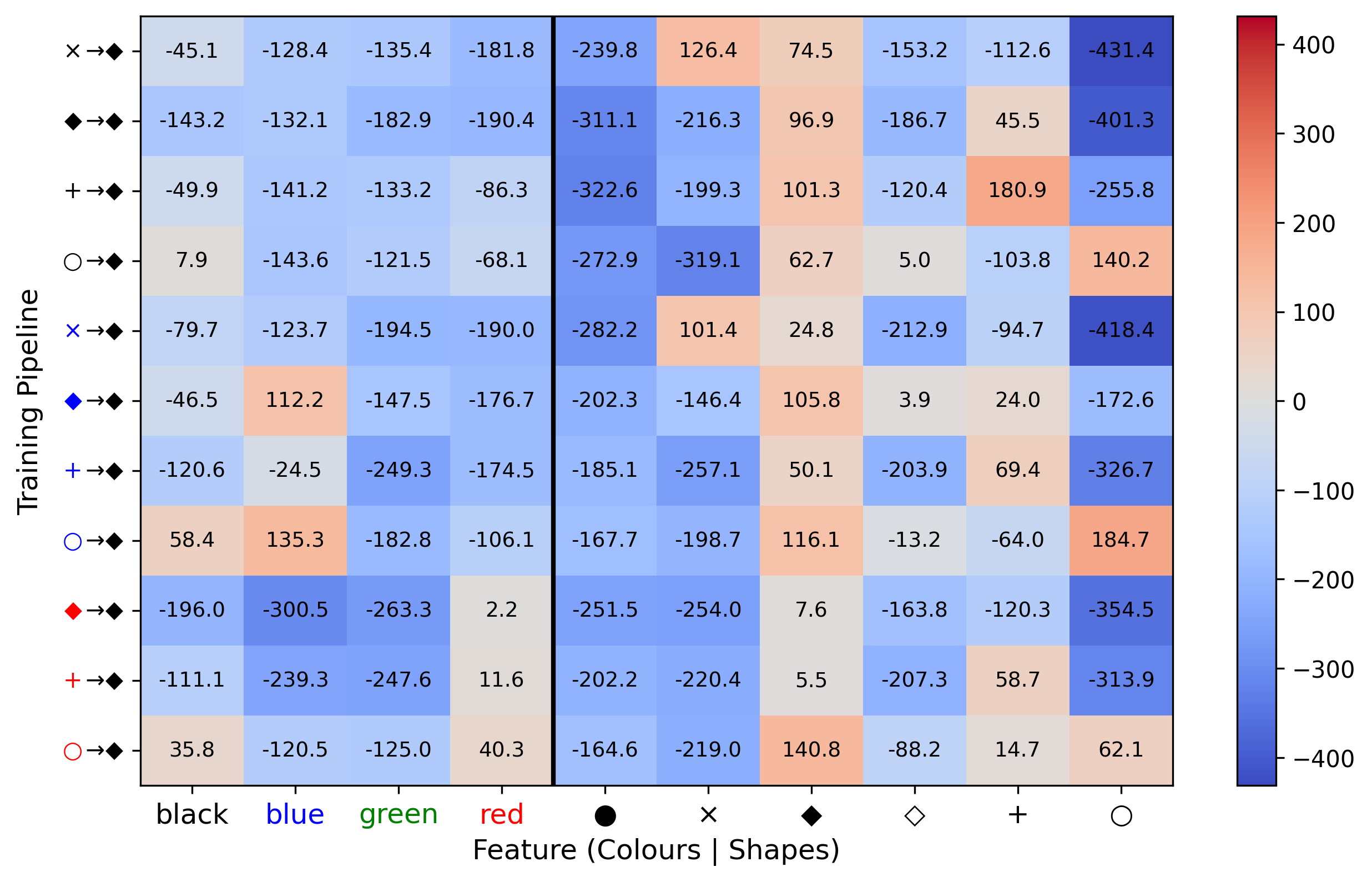}
        \caption{Black diamond}
        \label{fig:finetune_black_diamond}
    \end{subfigure}
    \caption{Agent values after two-stage training without distractors (1/4).}
    \label{fig:finetune_no_distractors_1}
\end{figure}

\begin{figure}[h]
    \centering
    \begin{subfigure}{0.49\linewidth}
        \includegraphics[width=\linewidth]{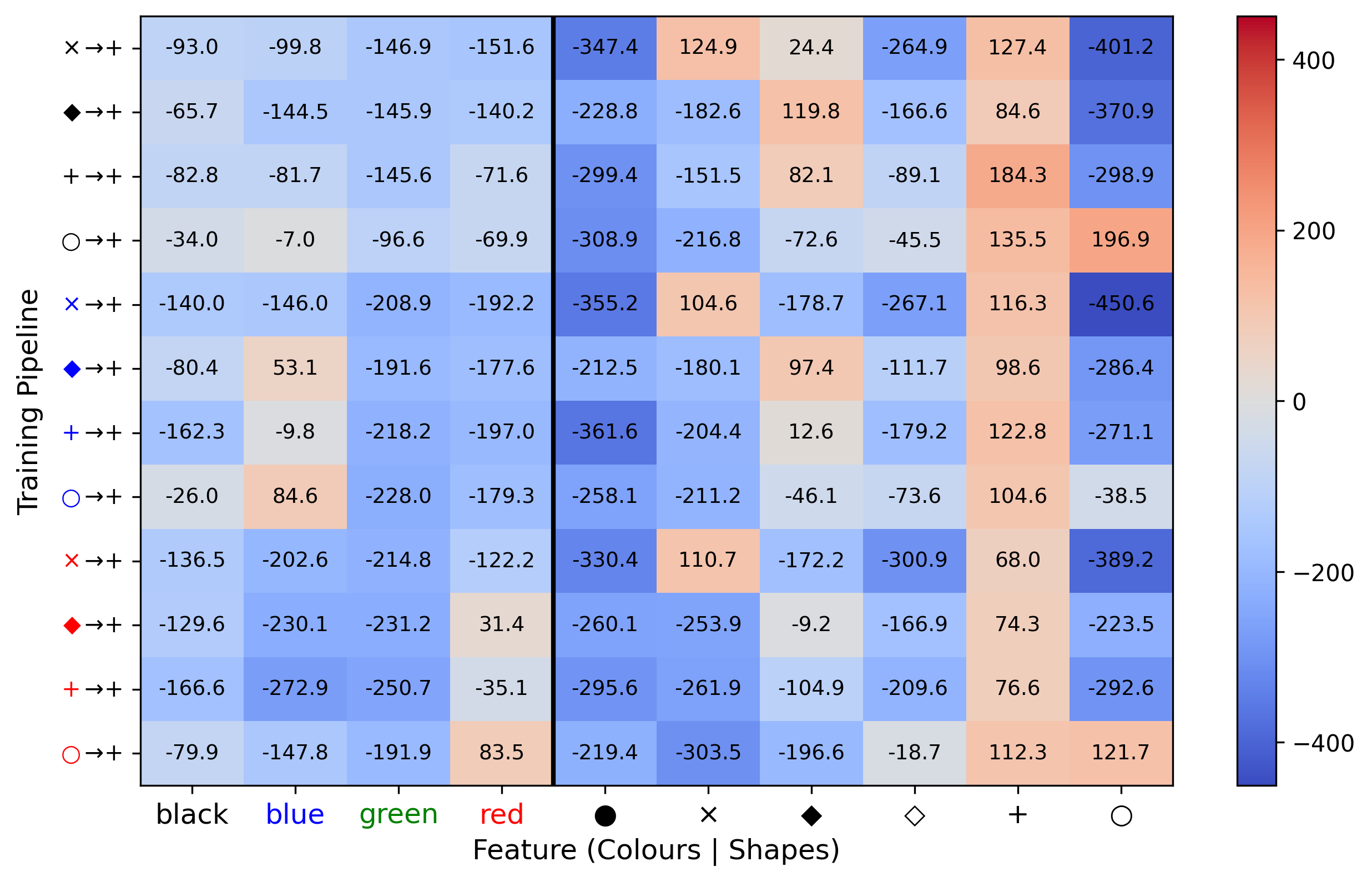}
        \caption{Black plus}
        \label{fig:finetune_black_plus}
    \end{subfigure}
    \begin{subfigure}{0.49\linewidth}
        \includegraphics[width=\linewidth]{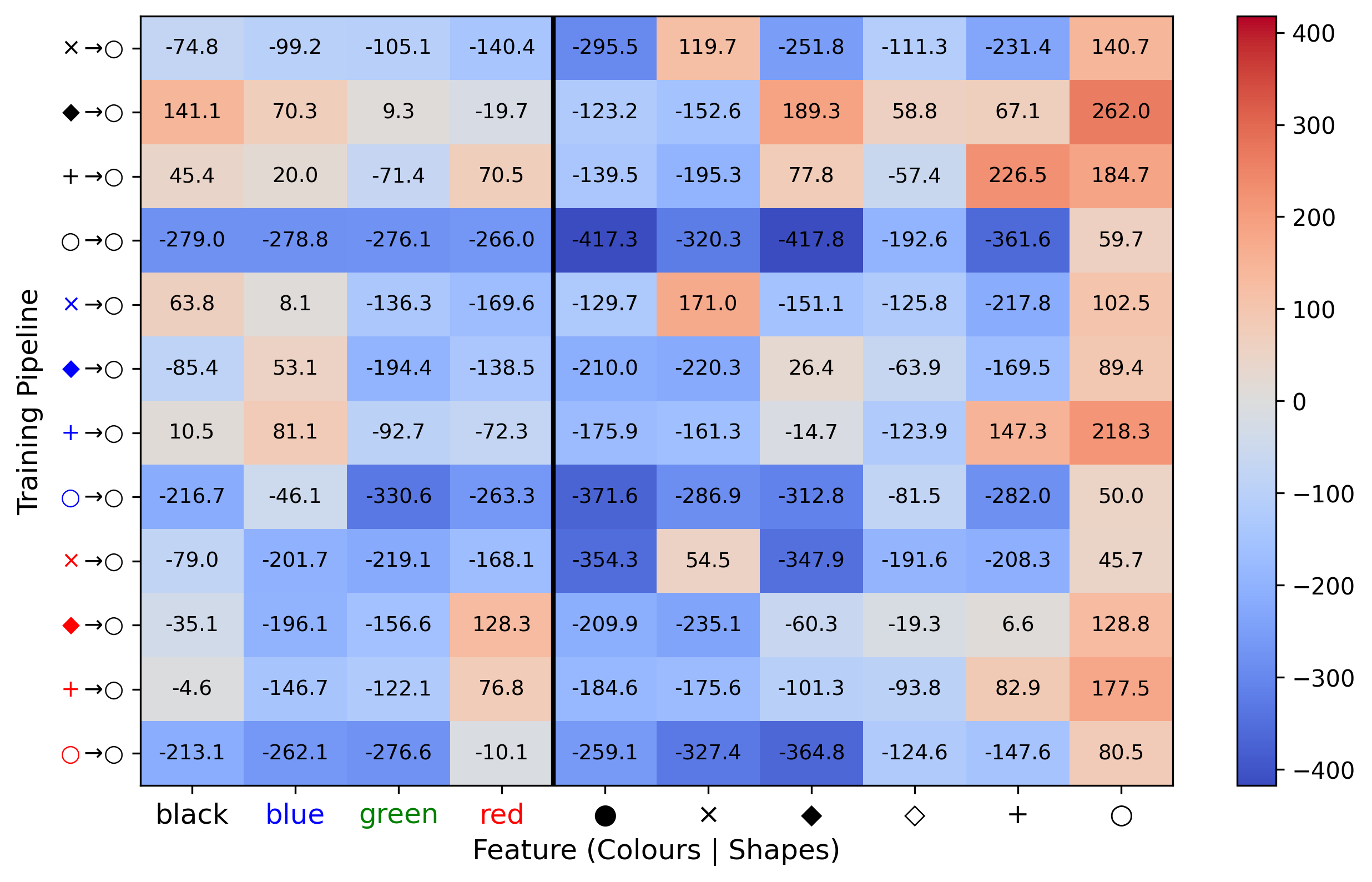}
        \caption{Black ring}
        \label{fig:finetune_black_ring}
    \end{subfigure}
    \caption{Agent values after two-stage training without distractors (2/4).}
    \label{fig:finetune_no_distractors_2}
\end{figure}

\begin{figure}[h]
    \centering
    \begin{subfigure}{0.49\linewidth}
        \includegraphics[width=\linewidth]{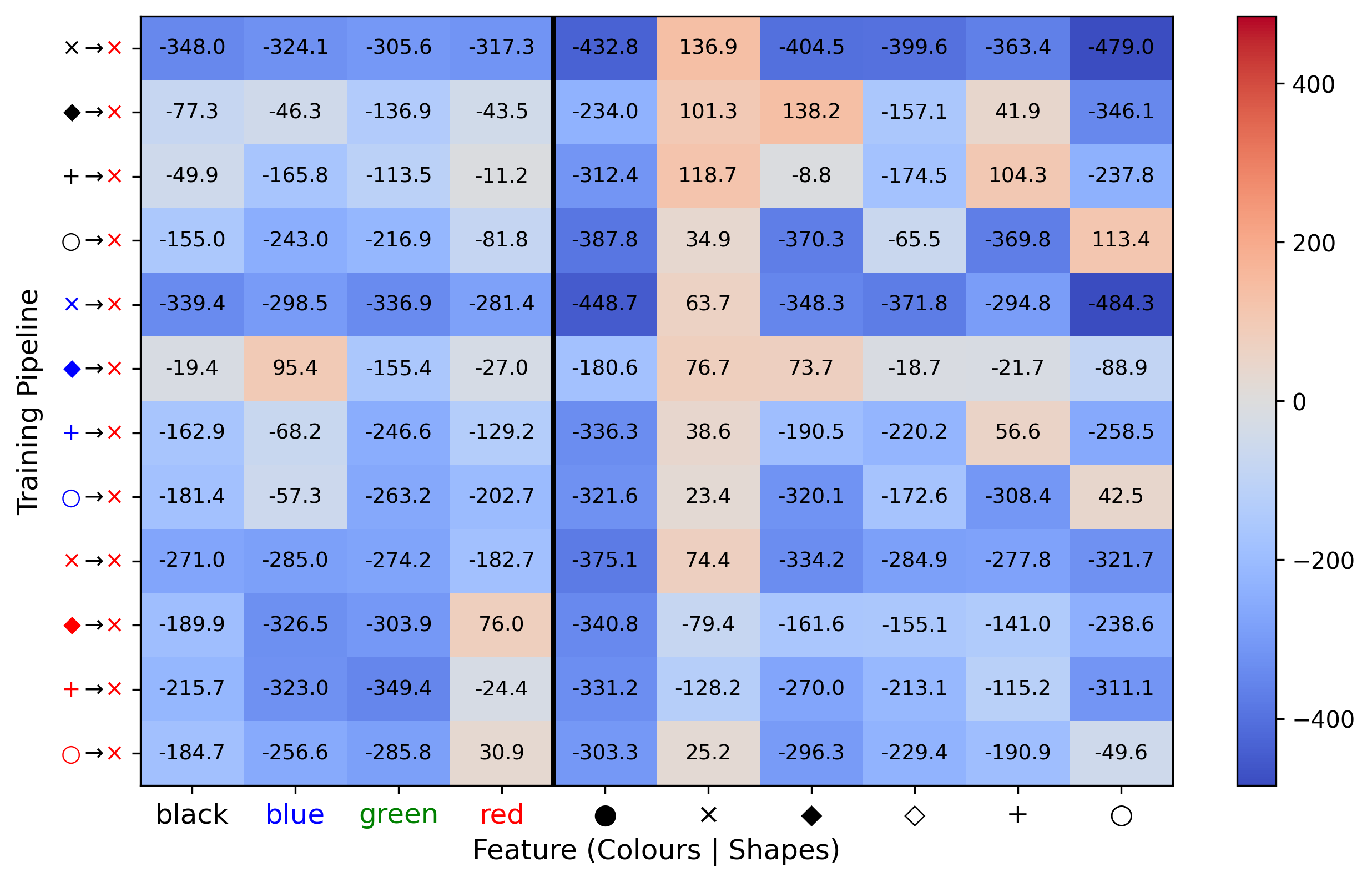}
        \caption{Red cross}
        \label{fig:finetune_red_cross}
    \end{subfigure}
    \begin{subfigure}{0.49\linewidth}
        \includegraphics[width=\linewidth]{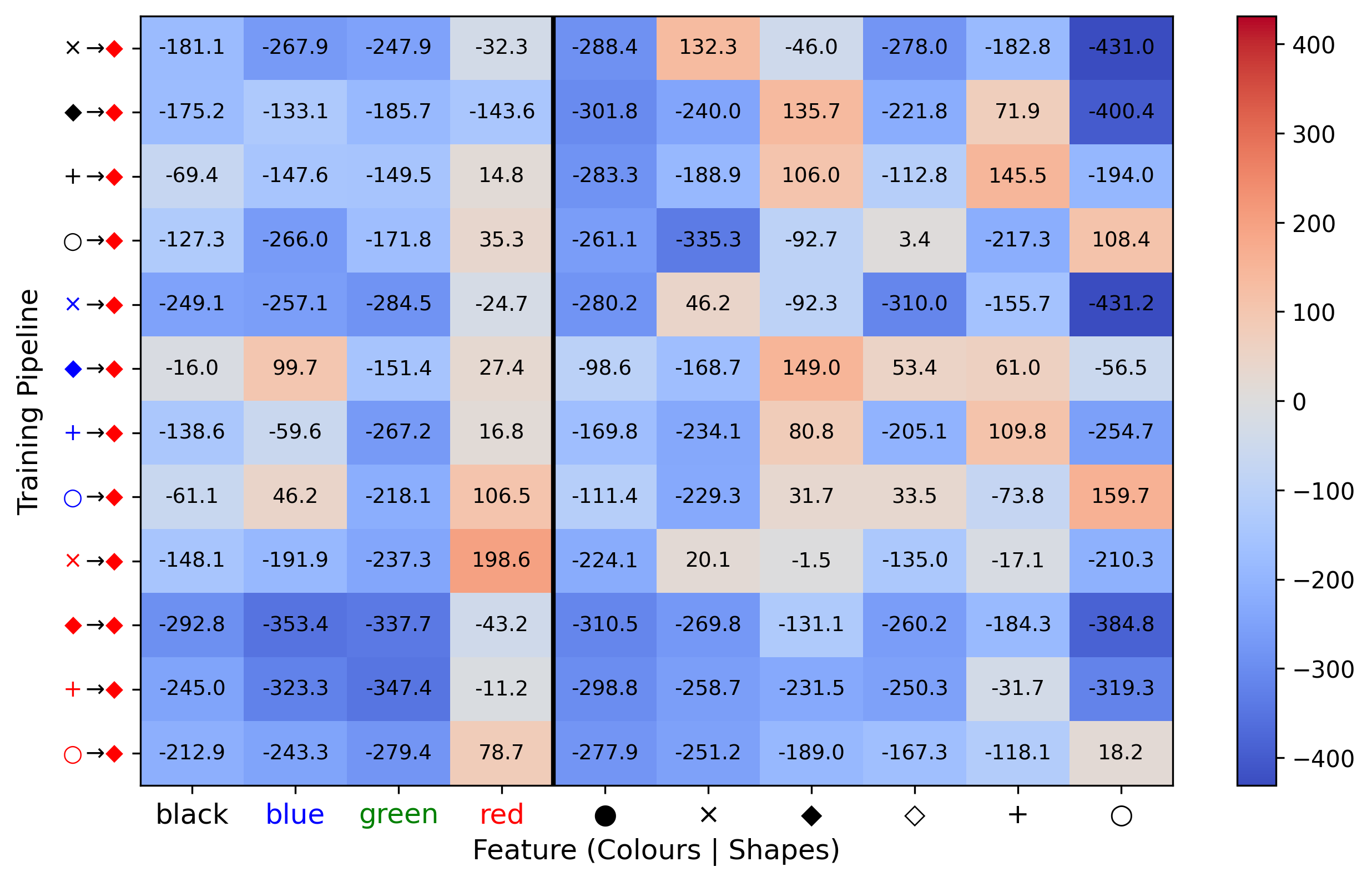}
        \caption{Red diamond}
        \label{fig:finetune_red_diamond}
    \end{subfigure}
    \caption{Agent values after two-stage training without distractors (3/4).}
    \label{fig:finetune_no_distractors_3}
\end{figure}

\begin{figure}[h]
    \centering
    \begin{subfigure}{0.49\linewidth}
        \includegraphics[width=\linewidth]{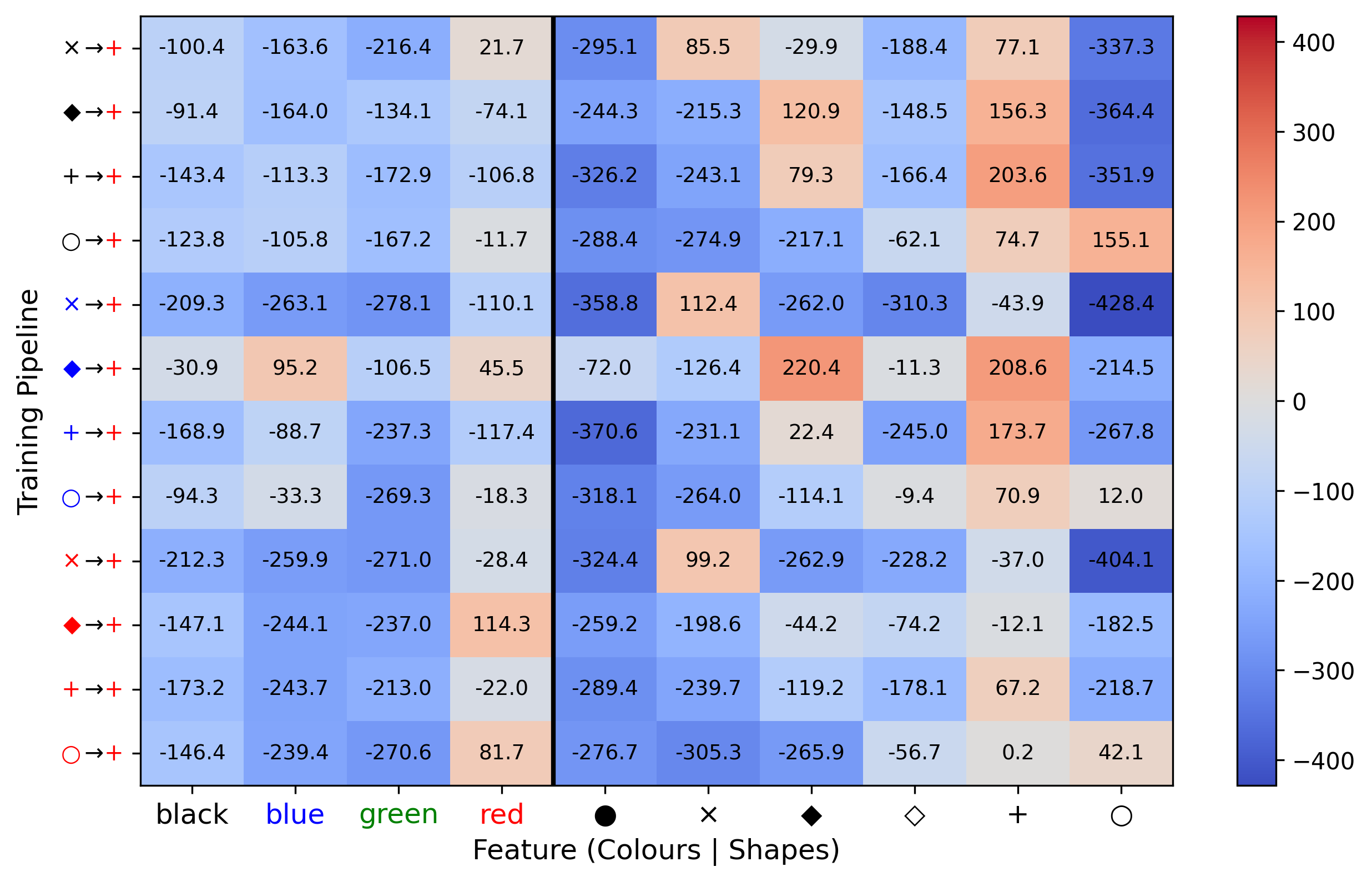}
        \caption{Red plus}
        \label{fig:finetune_red_plus}
    \end{subfigure}
    \begin{subfigure}{0.49\linewidth}
        \includegraphics[width=\linewidth]{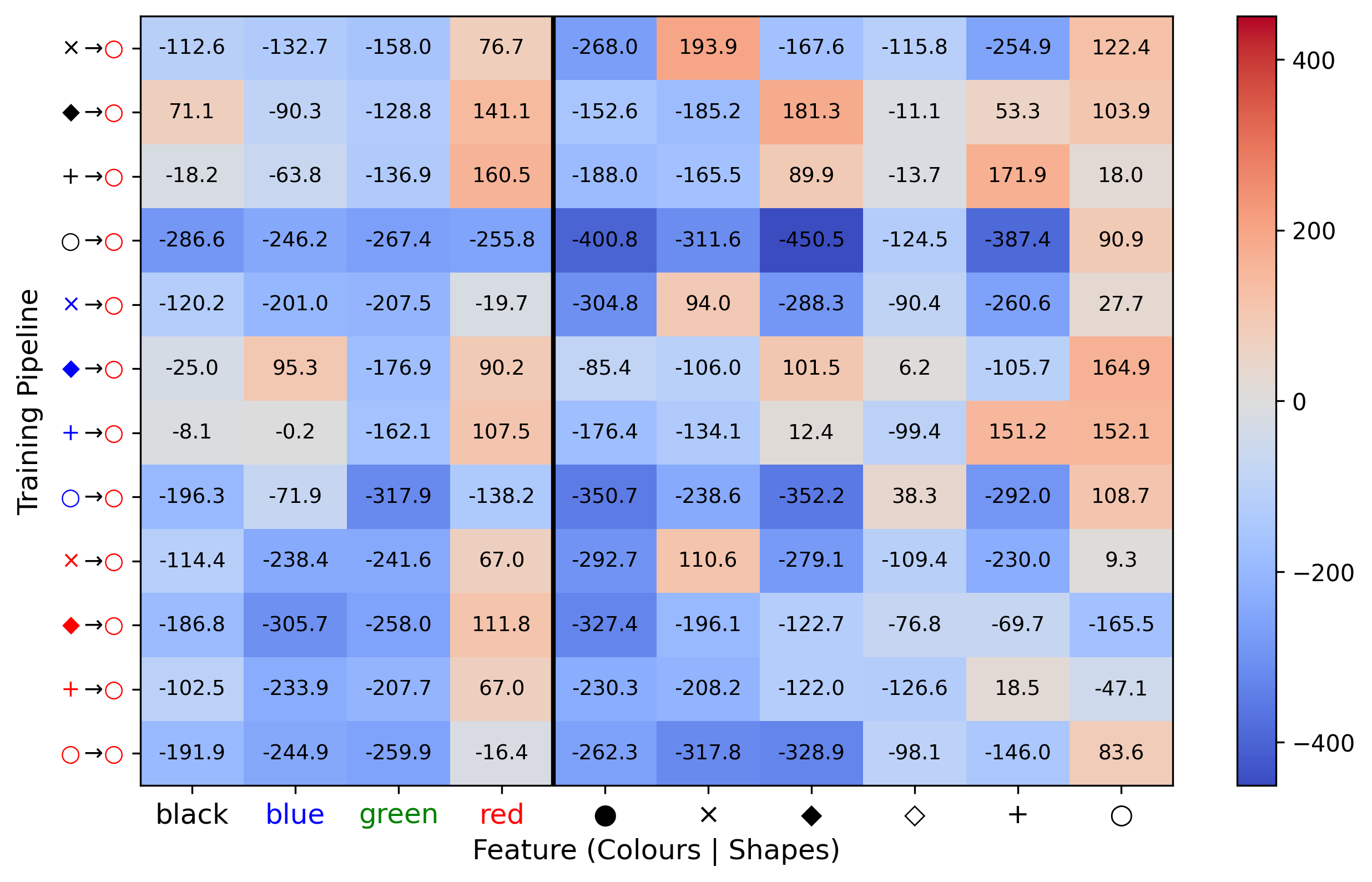}
        \caption{Red ring}
        \label{fig:finetune_red_ring}
    \end{subfigure}
    \caption{Agent values after two-stage training without distractors (4/4).}
    \label{fig:finetune_no_distractors_4}
\end{figure}

\begin{figure}[h]
    \centering
    \begin{subfigure}{0.49\linewidth}
        \includegraphics[width=\linewidth]{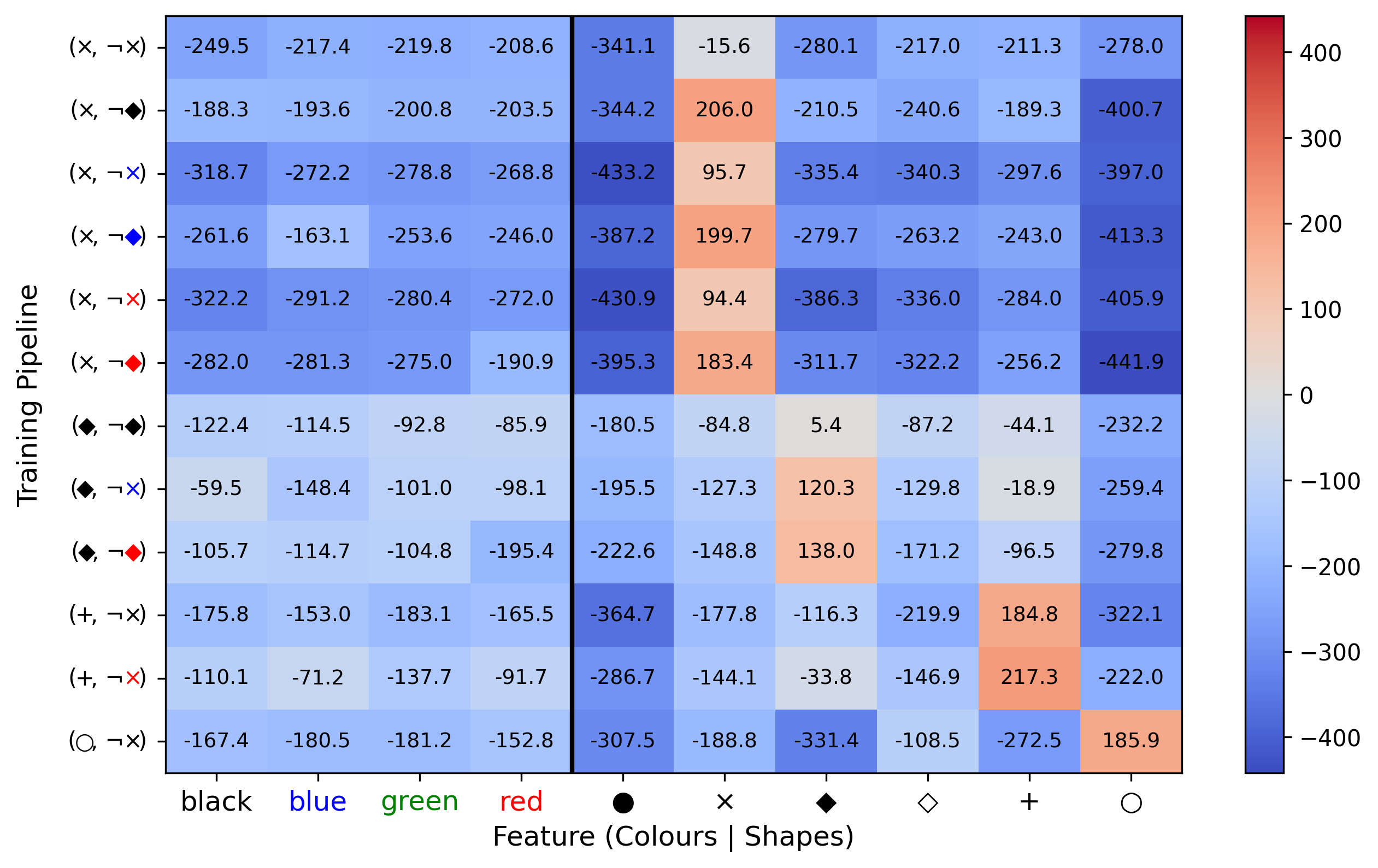}
        \caption{}
        \label{fig:single_stage_distractors_1}
    \end{subfigure}
    \begin{subfigure}{0.49\linewidth}
        \includegraphics[width=\linewidth]{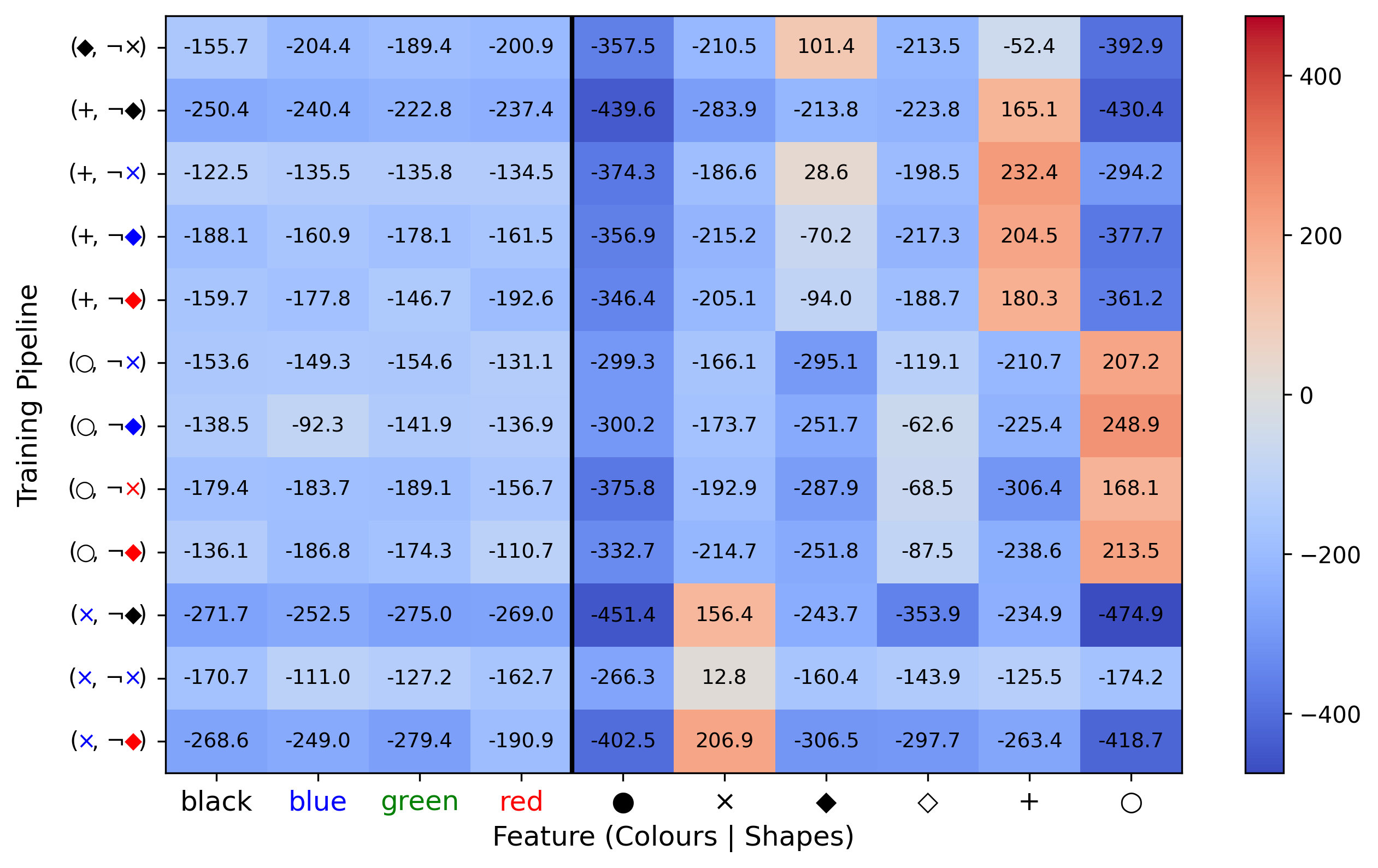}
        \caption{}
        \label{fig:single_stage_distractors_2}
    \end{subfigure}
    \caption{Agent values for single-stage training with distractors (1/3).}
    \label{fig:single_stage_distractors_1_2}
\end{figure}

\begin{figure}[h]
    \centering
    \begin{subfigure}{0.49\linewidth}
        \includegraphics[width=\linewidth]{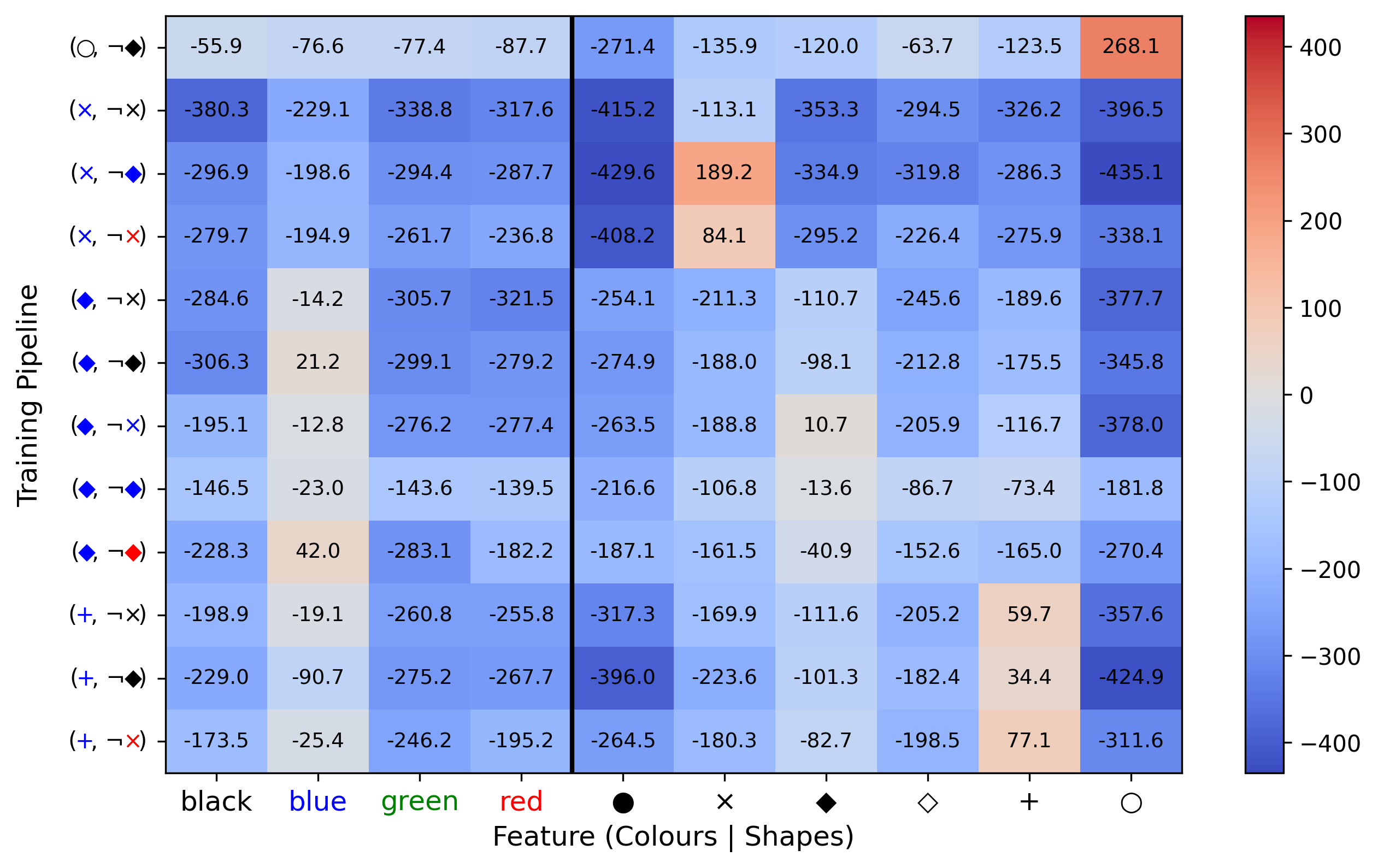}
        \caption{}
        \label{fig:single_stage_distractors_3}
    \end{subfigure}
    \begin{subfigure}{0.49\linewidth}
        \includegraphics[width=\linewidth]{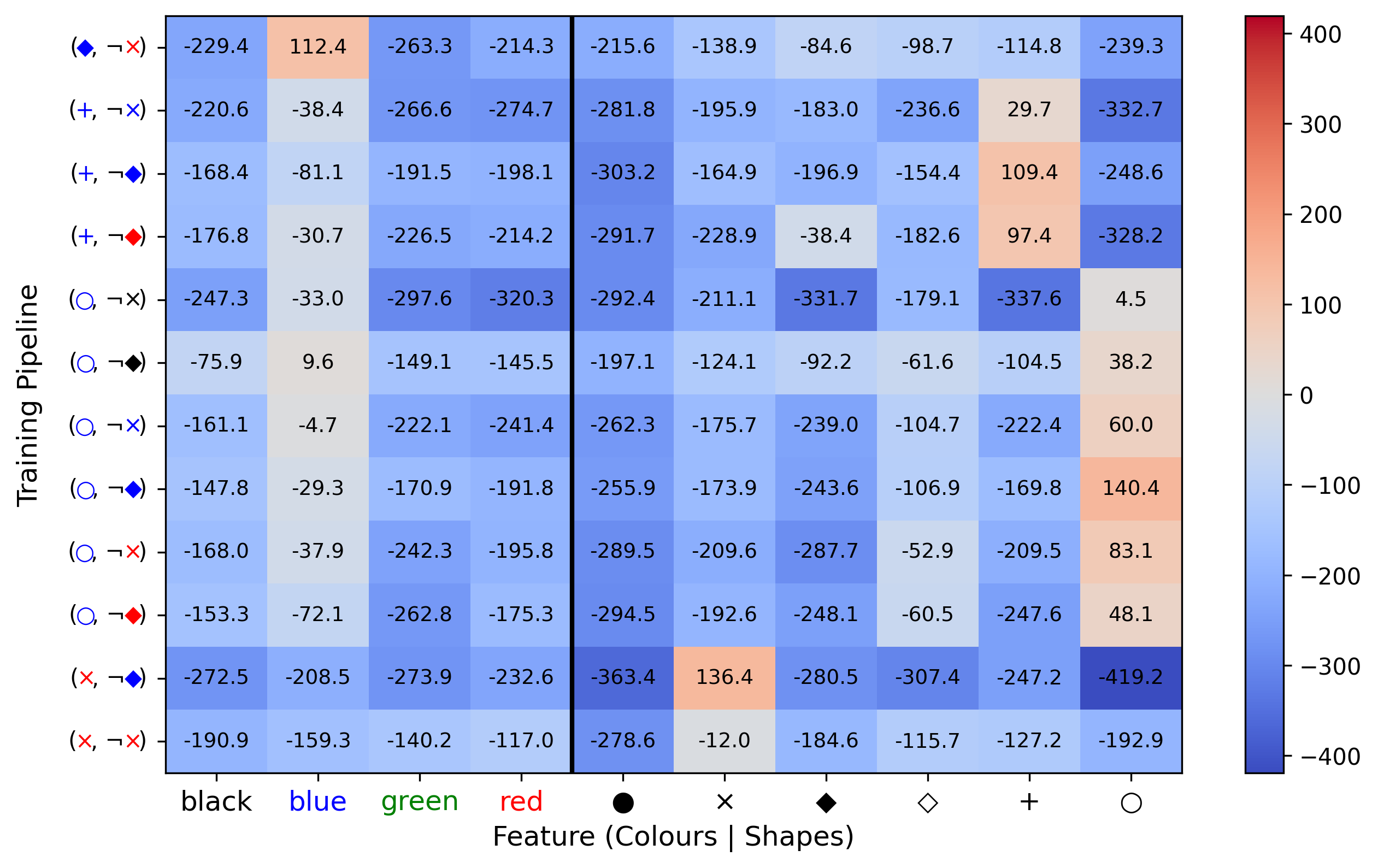}
        \caption{}
        \label{fig:single_stage_distractors_4}
    \end{subfigure}
    \caption{Agent values for single-stage training with distractors (2/3).}
    \label{fig:single_stage_distractors_3_4}
\end{figure}

\begin{figure}[h]
    \centering
    \begin{subfigure}{0.49\linewidth}
        \includegraphics[width=\linewidth]{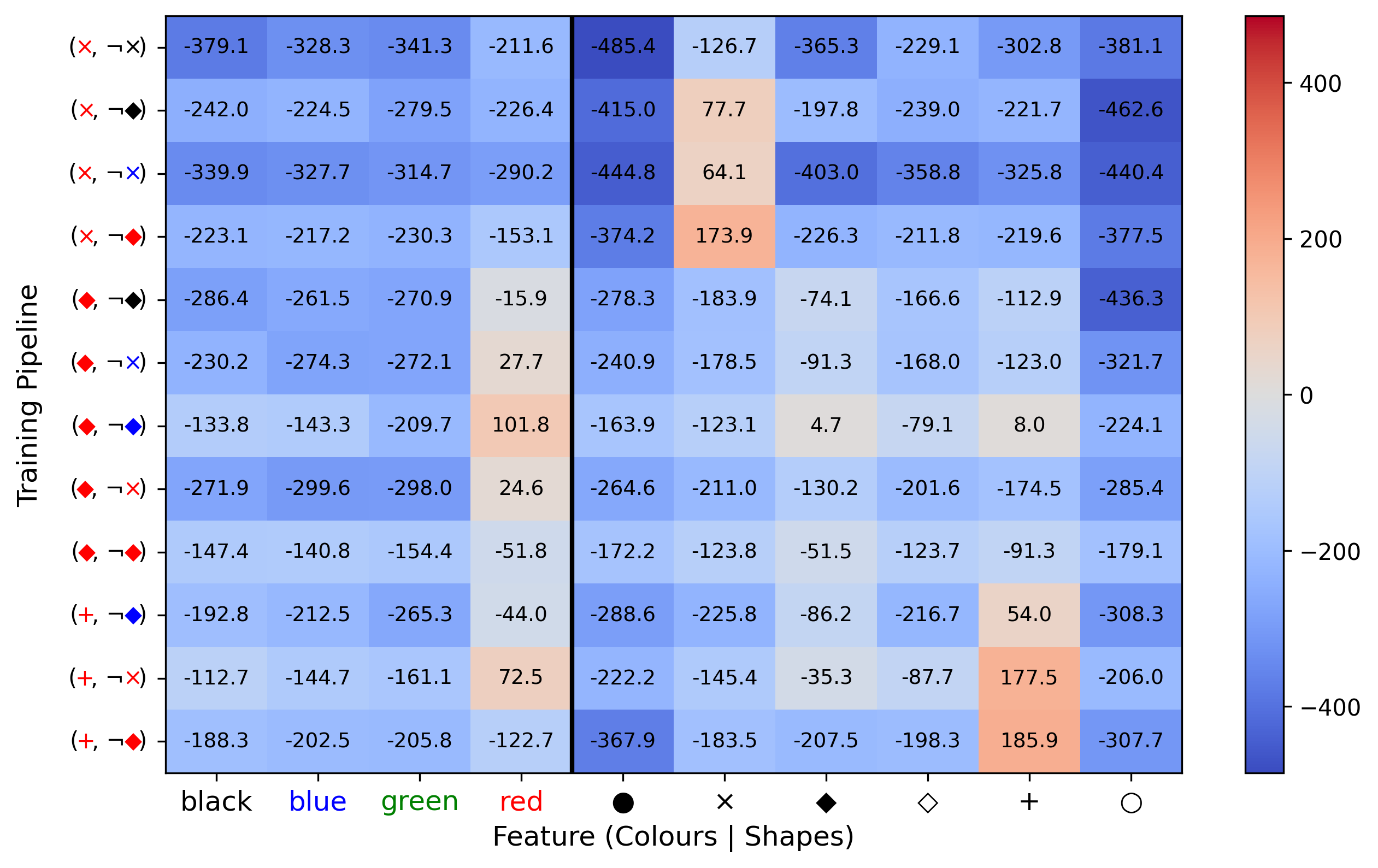}
        \caption{}
        \label{fig:single_stage_distractors_5}
    \end{subfigure}
    \begin{subfigure}{0.49\linewidth}
        \includegraphics[width=\linewidth]{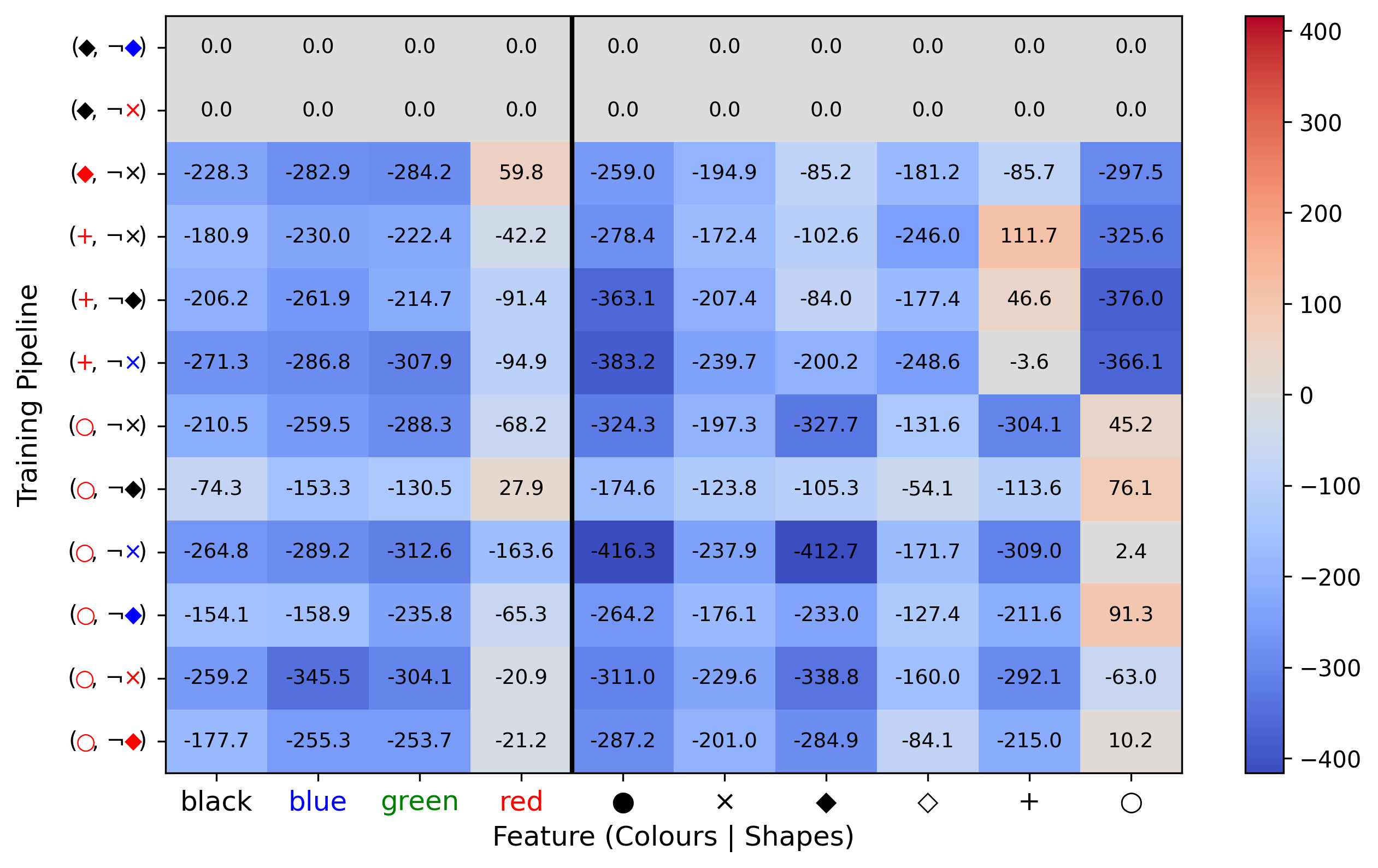}
        \caption{}
        \label{fig:single_stage_distractors_6}
    \end{subfigure}
    \caption{Agent values for single-stage training with distractors (3/3).}
    \label{fig:single_stage_distractors_5_6}
\end{figure}

\begin{figure}[h]
    \centering
    \begin{subfigure}{0.49\linewidth}
        \includegraphics[width=\linewidth]{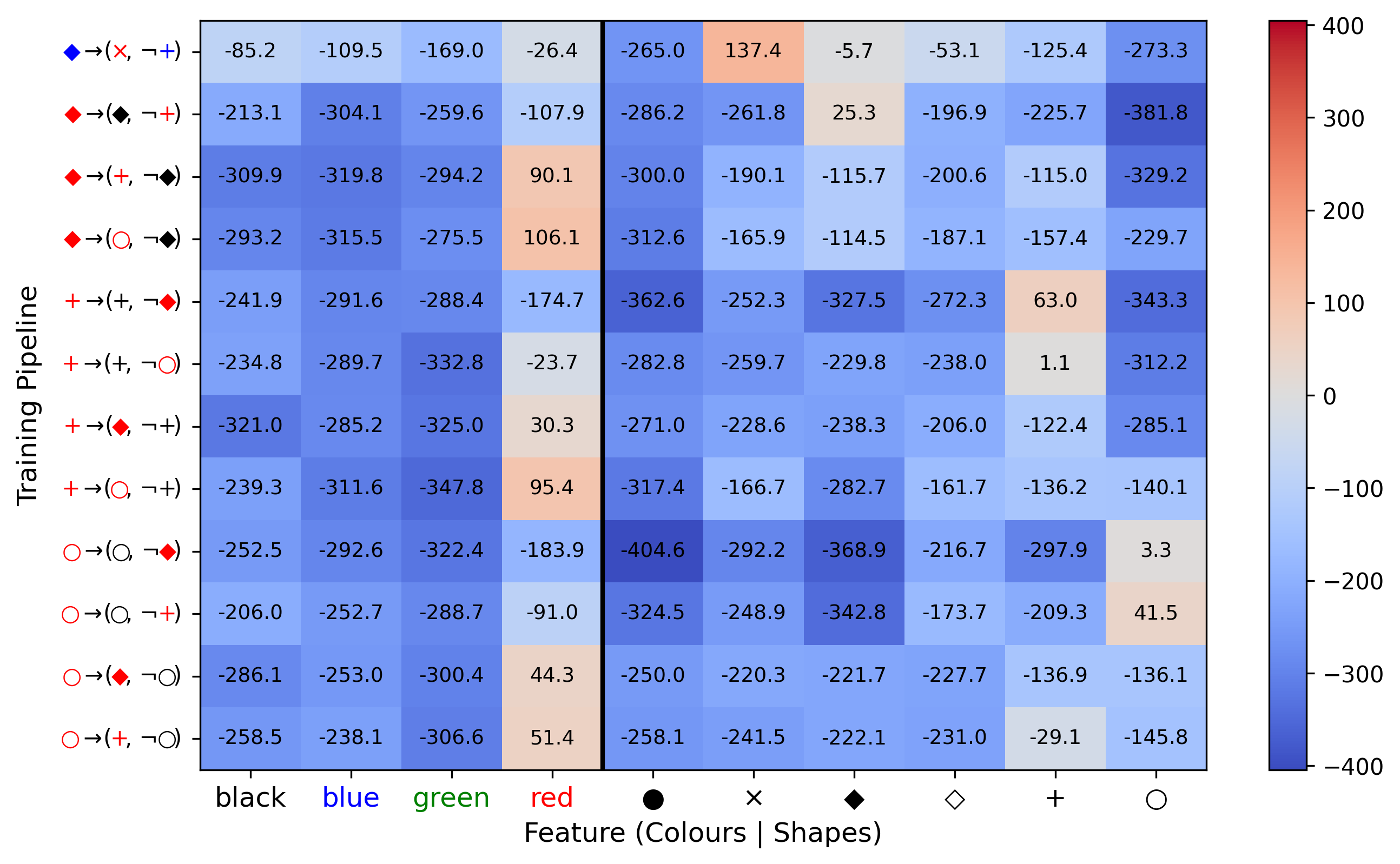}
        \caption{}
        \label{fig:finetune_distractors_1}
    \end{subfigure}
    \begin{subfigure}{0.49\linewidth}
        \includegraphics[width=\linewidth]{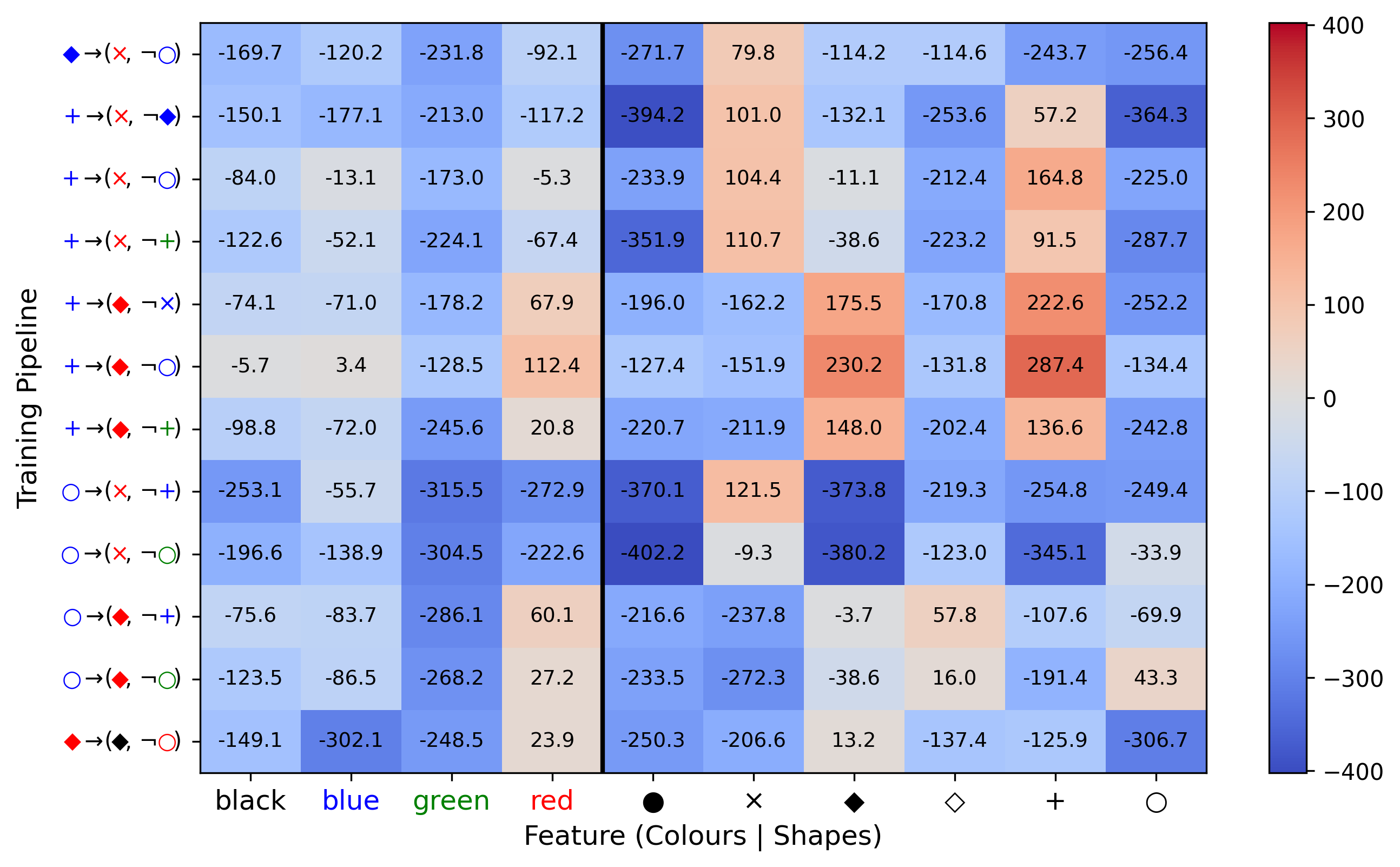}
        \caption{}
        \label{fig:finetune_distractors_2}
    \end{subfigure}
    \caption{Agent values after two-stage training with distractors (1/5).}
    \label{fig:finetune_distractors_1_2}
\end{figure}

\begin{figure}[h]
    \centering
    \begin{subfigure}{0.49\linewidth}
        \includegraphics[width=\linewidth]{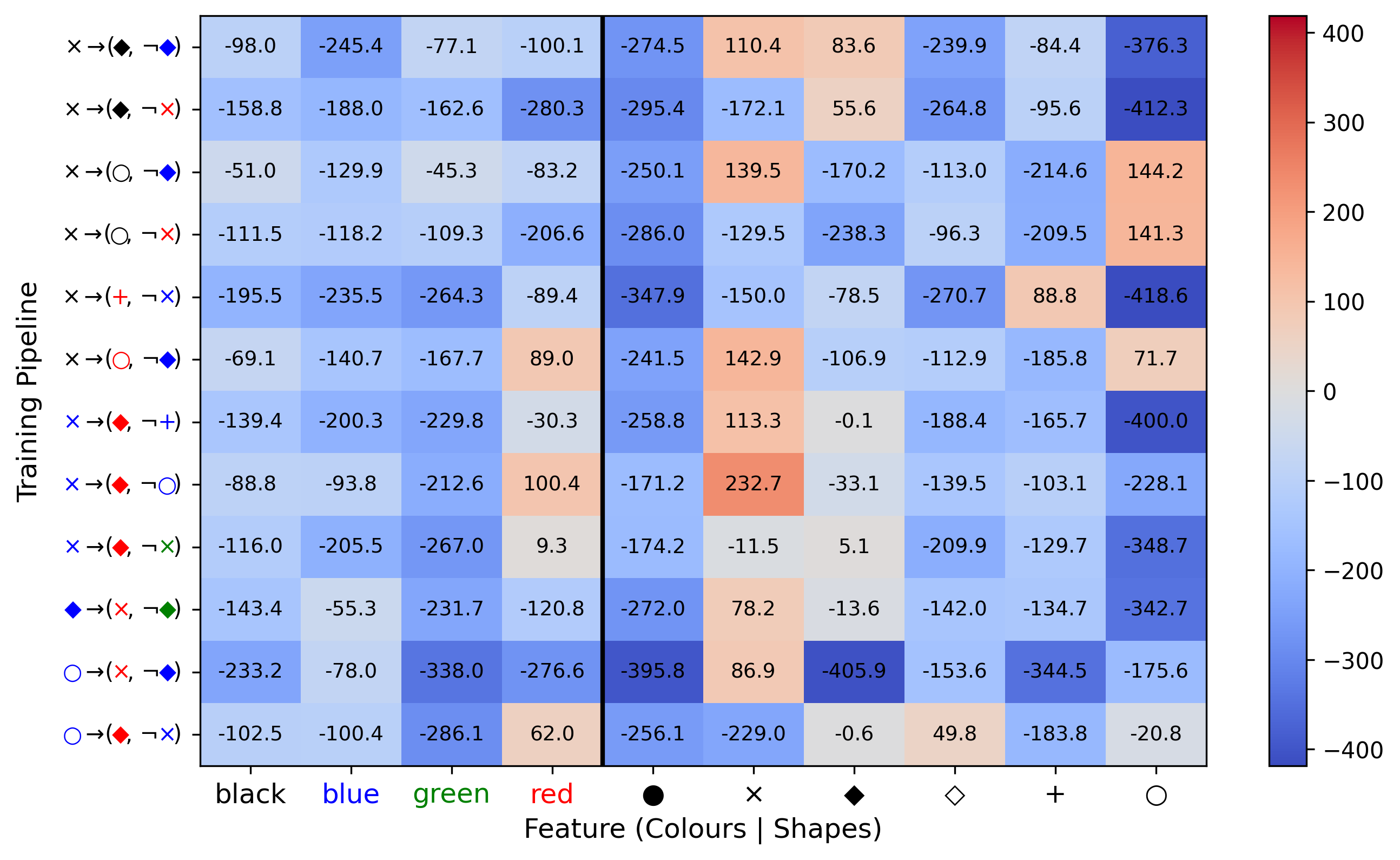}
        \caption{}
        \label{fig:finetune_distractors_3}
    \end{subfigure}
    \begin{subfigure}{0.49\linewidth}
        \includegraphics[width=\linewidth]{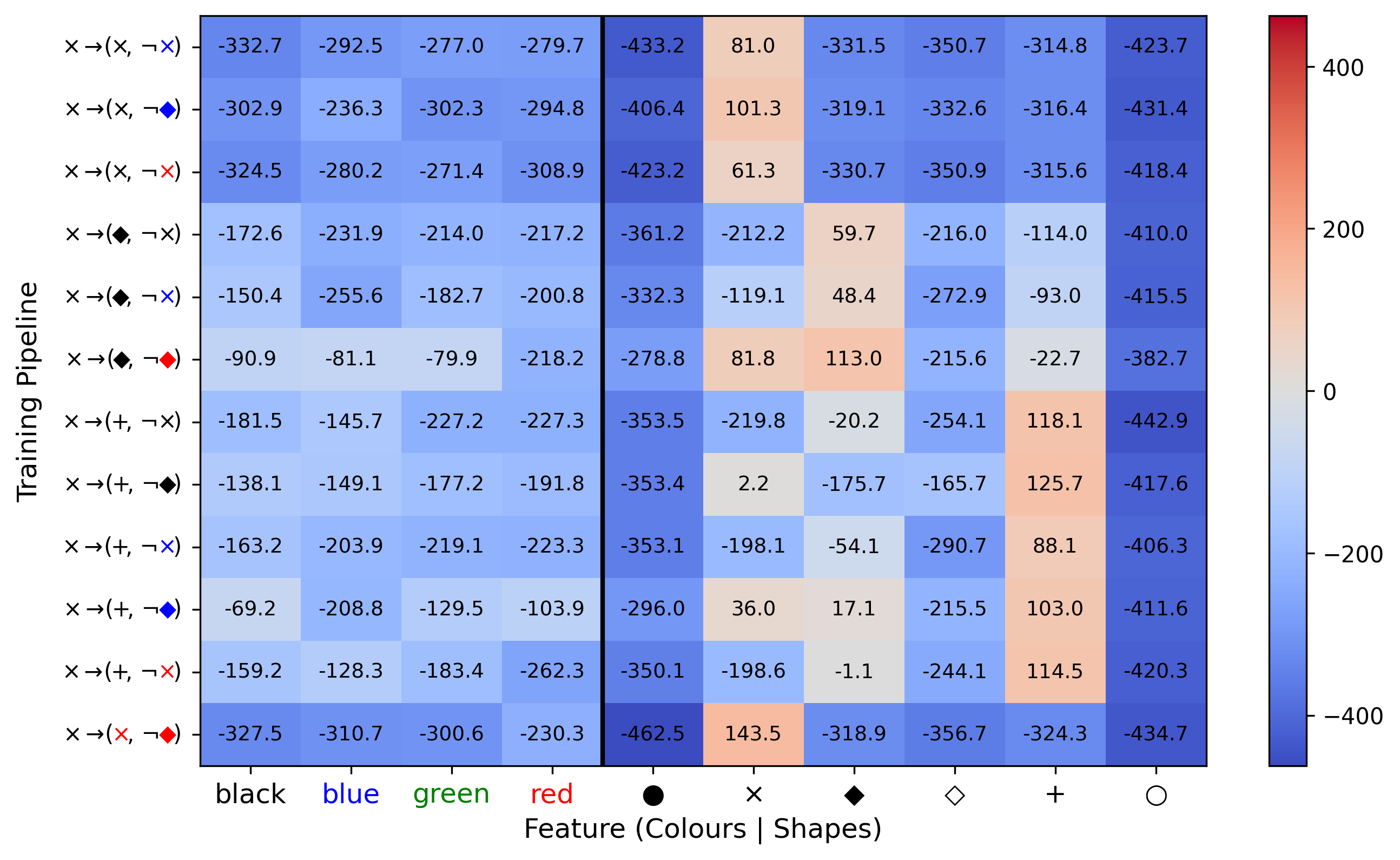}
        \caption{}
        \label{fig:finetune_distractors_4}
    \end{subfigure}
    \caption{Agent values after two-stage training with distractors (2/5).}
    \label{fig:finetune_distractors_3_4}
\end{figure}

\begin{figure}[h]
    \centering
    \begin{subfigure}{0.49\linewidth}
        \includegraphics[width=\linewidth]{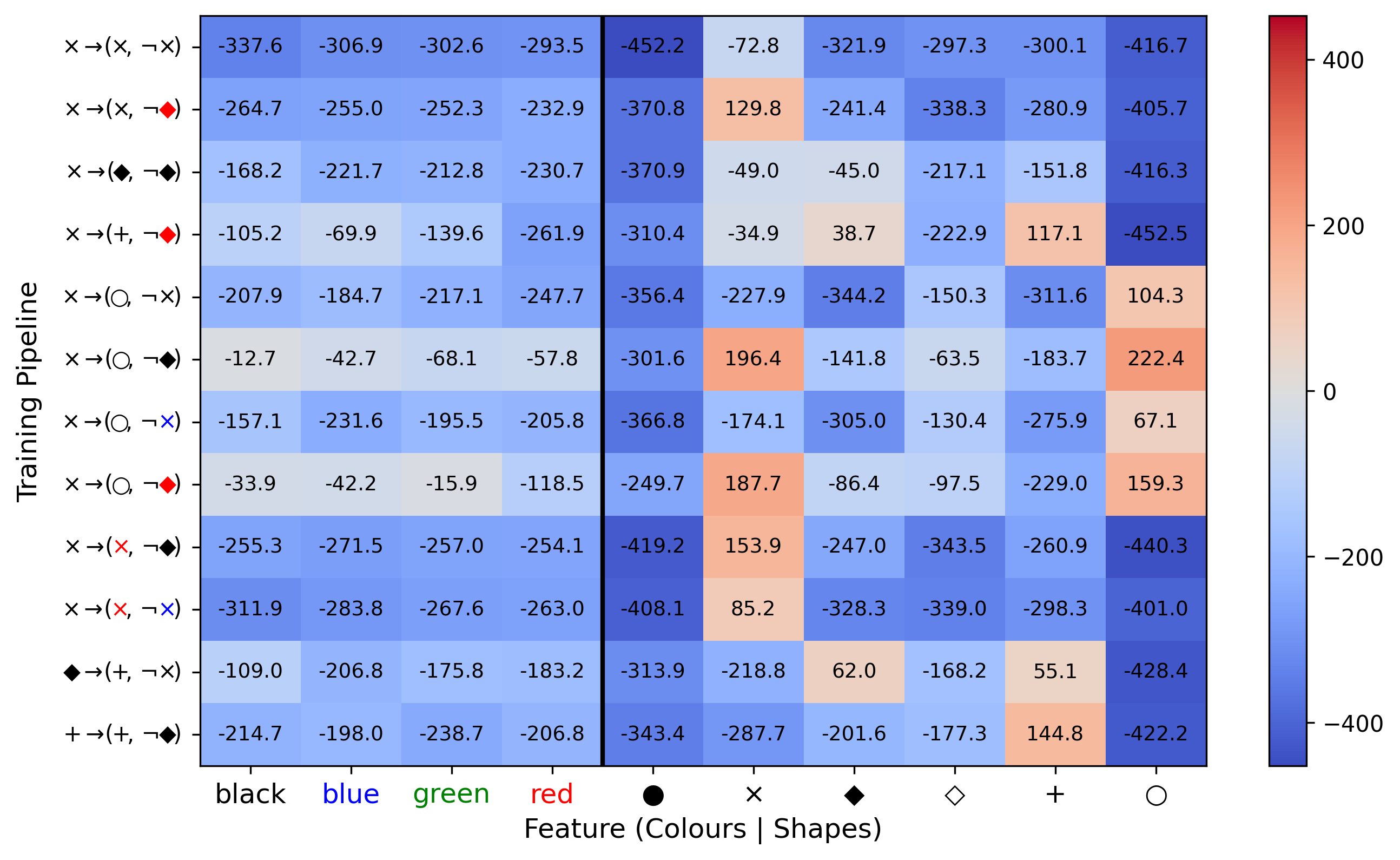}
        \caption{}
        \label{fig:finetune_distractors_5}
    \end{subfigure}
    \begin{subfigure}{0.49\linewidth}
        \includegraphics[width=\linewidth]{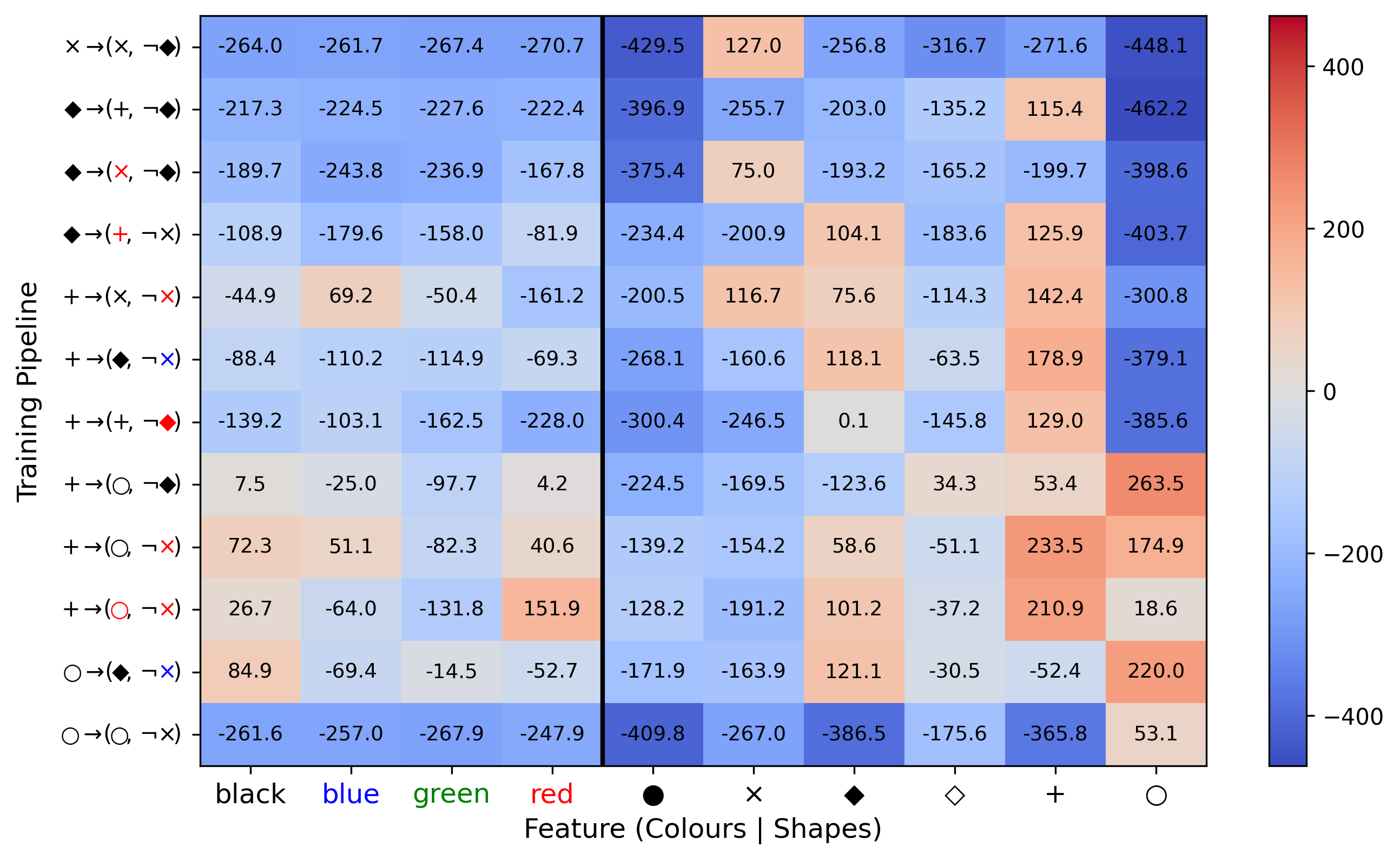}
        \caption{}
        \label{fig:finetune_distractors_6}
    \end{subfigure}
    \caption{Agent values after two-stage training with distractors (3/5).}
    \label{fig:finetune_distractors_5_6}
\end{figure}

\begin{figure}[h]
    \centering
    \begin{subfigure}{0.49\linewidth}
        \includegraphics[width=\linewidth]{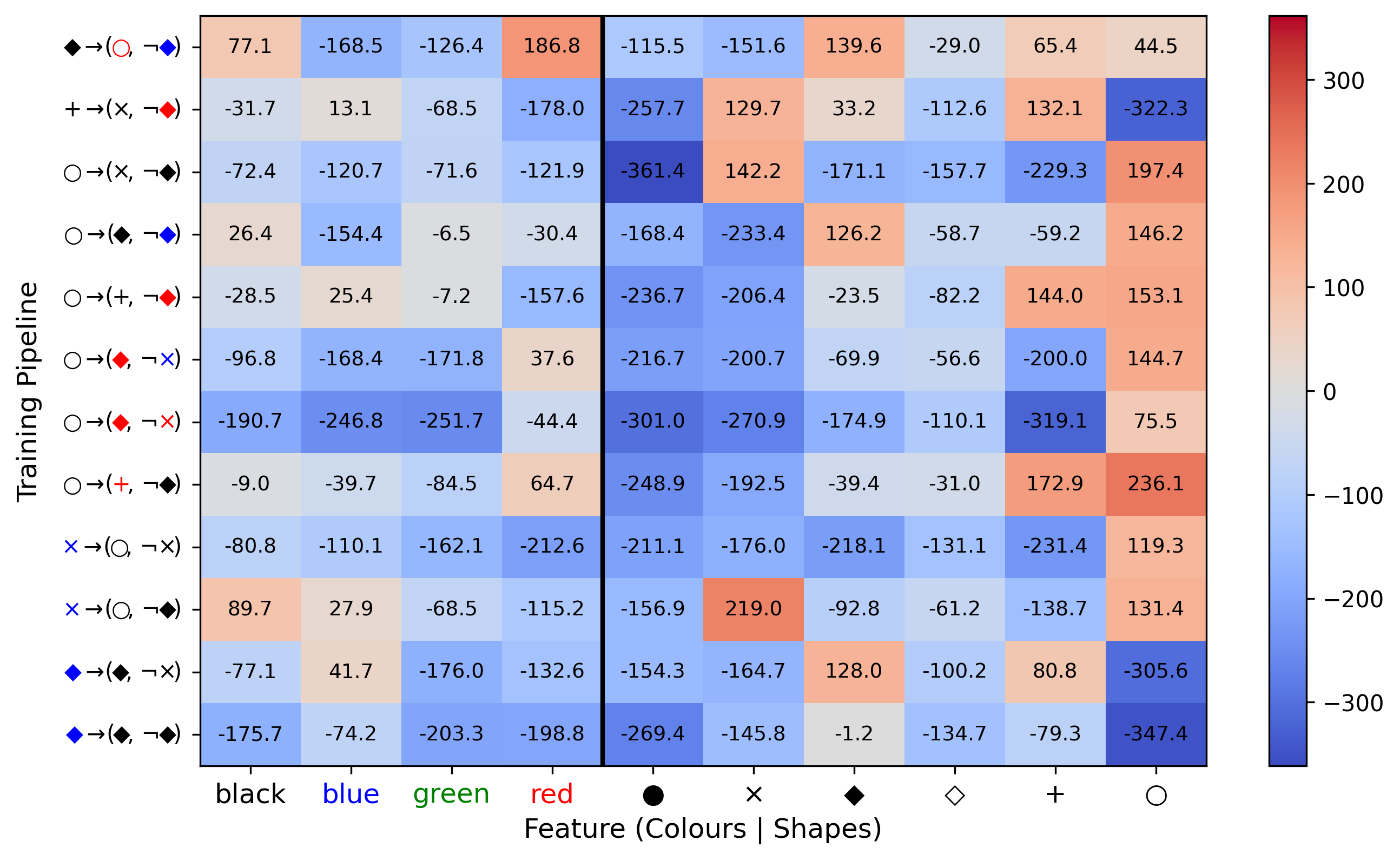}
        \caption{}
        \label{fig:finetune_distractors_7}
    \end{subfigure}
    \begin{subfigure}{0.49\linewidth}
        \includegraphics[width=\linewidth]{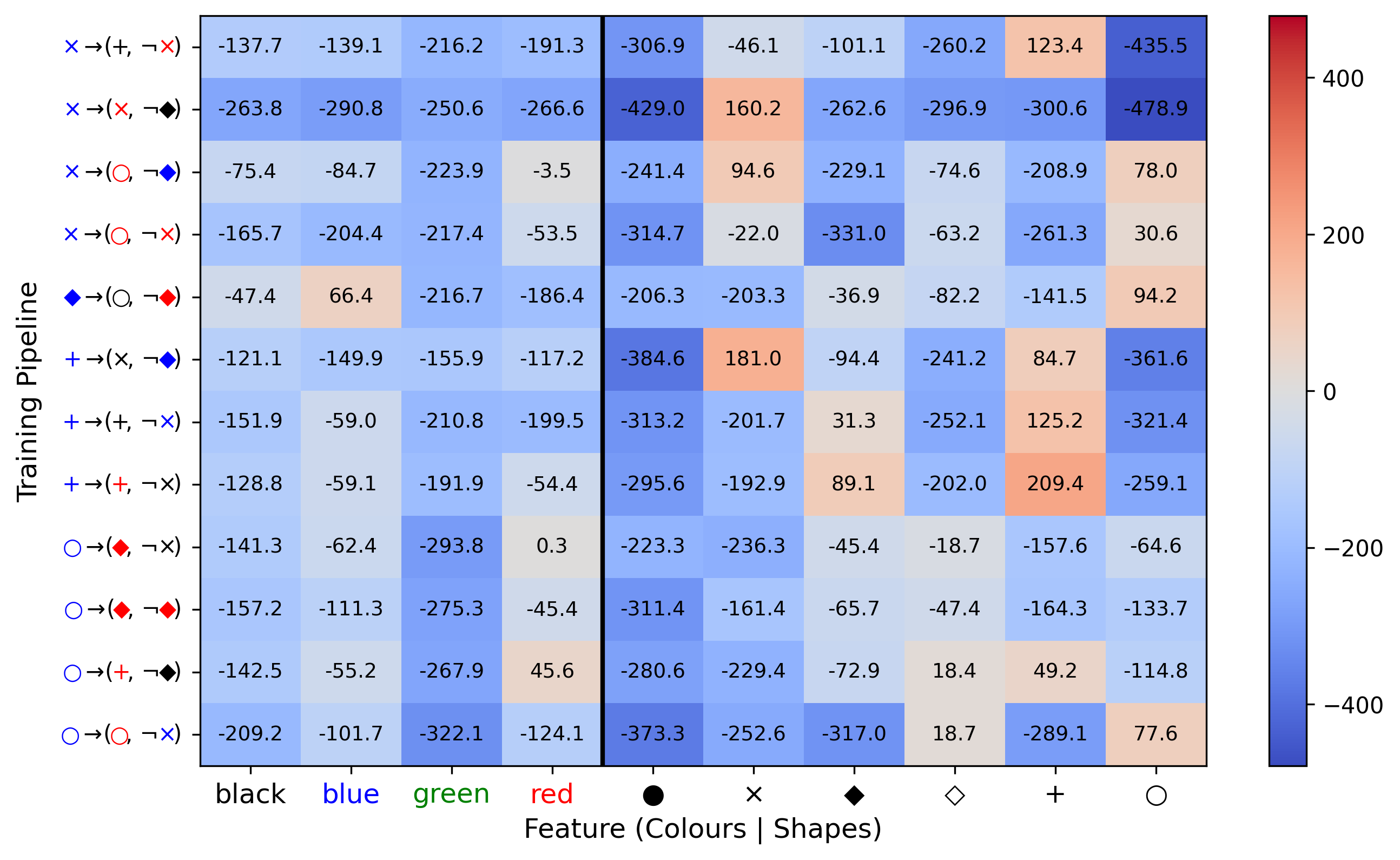}
        \caption{}
        \label{fig:finetune_distractors_8}
    \end{subfigure}
    \caption{Agent values after two-stage training with distractors (4/5).}
    \label{fig:finetune_distractors_7_8}
\end{figure}

\begin{figure}[h]
    \centering
    \begin{subfigure}{0.49\linewidth}
        \includegraphics[width=\linewidth]{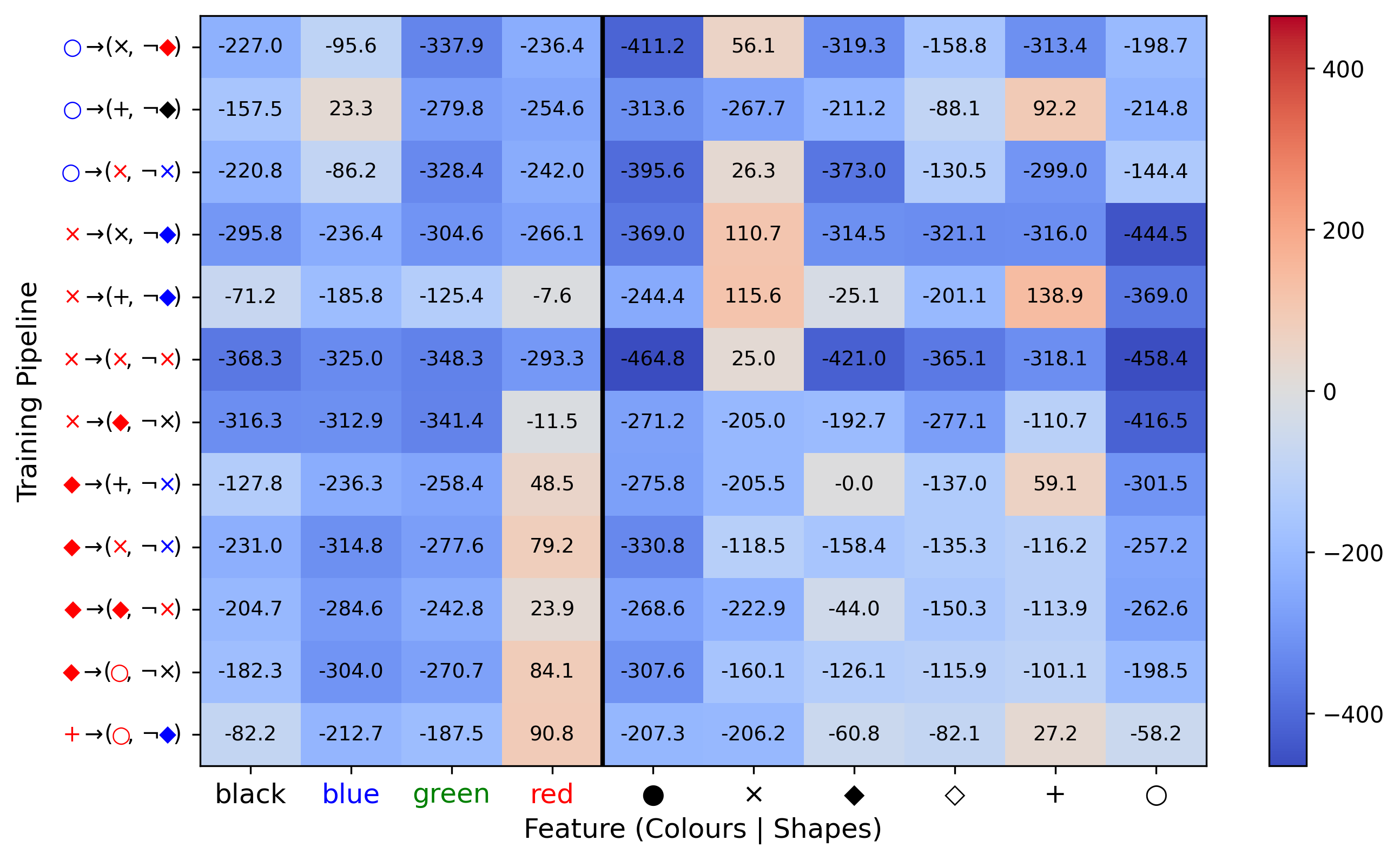}
        \caption{}
        \label{fig:finetune_distractors_9}
    \end{subfigure}
    \begin{subfigure}{0.49\linewidth}
        \includegraphics[width=\linewidth]{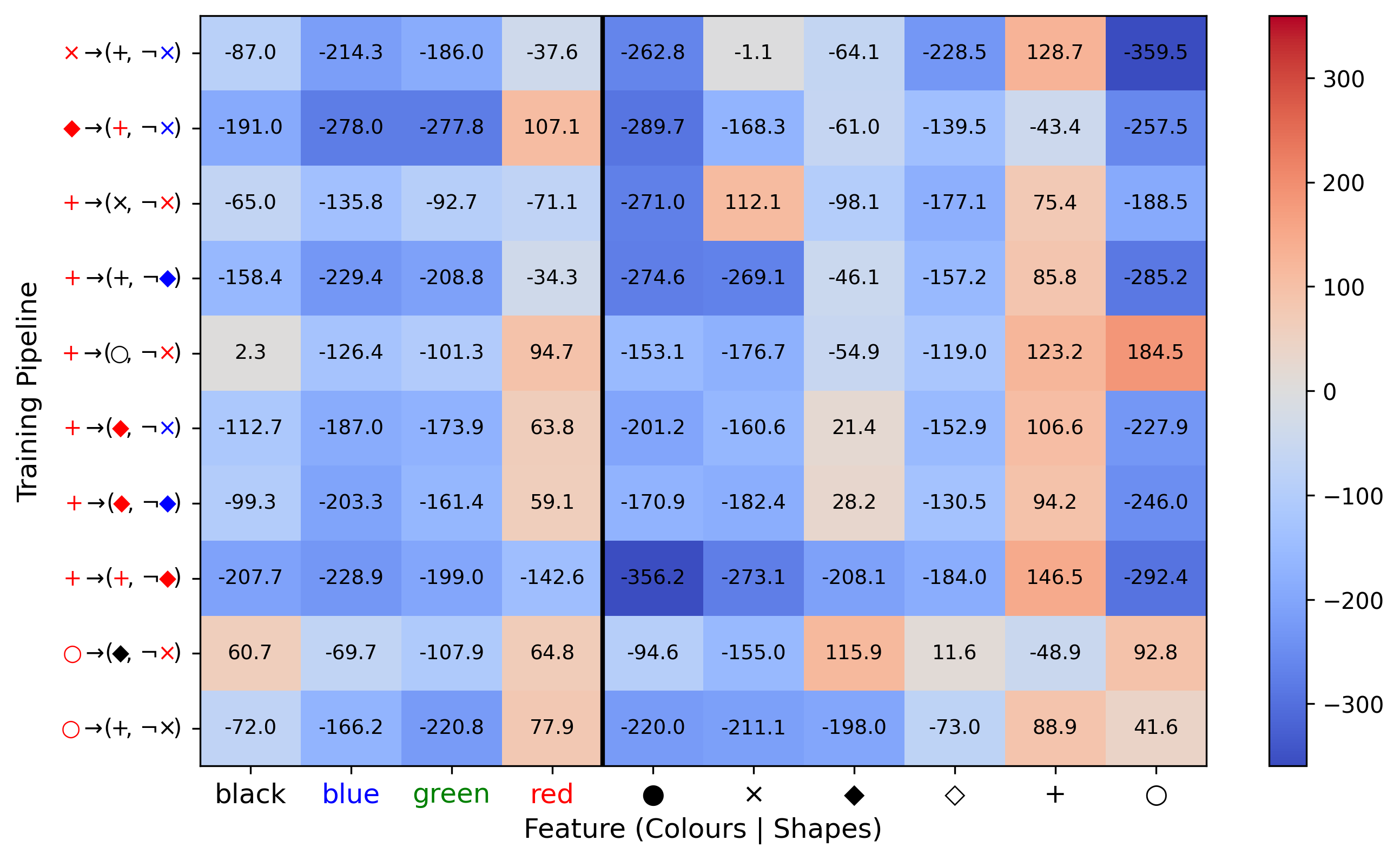}
        \caption{}
        \label{fig:finetune_distractors_10}
    \end{subfigure}
    \caption{Agent values after two-stage training with distractors (5/5).}
    \label{fig:finetune_distractors_9_10}
\end{figure}

\FloatBarrier

\end{document}